\pdfoutput=1

\documentclass[11pt]{article}

\usepackage{acl}

\usepackage{times}
\usepackage{latexsym}

\usepackage[T1]{fontenc}

\usepackage[utf8]{inputenc}

\usepackage{microtype}

\usepackage{xspace}
\usepackage{booktabs}
\usepackage{graphicx}
\usepackage{enumitem}
\usepackage{xcolor,colortbl}
\usepackage{tikz-dependency}
\usepackage{url}
\usepackage{amssymb}
\usepackage{amsmath} %
\usepackage{bbm}  %
\usepackage{float}
\usepackage{todonotes}

\usepackage{titlefoot} %

\usepackage{caption}
\usepackage{subcaption}
\usepackage{tcolorbox}

\usepackage{tikz} %

\usepackage{rotating} %
\usepackage{graphics} %
\usepackage{multirow} %
\usepackage{bibentry} %
\usepackage{cleveref}  %
\usepackage{enumitem}
\setlist{nosep}

\nobibliography*

\setlist[description]{leftmargin=\parindent,labelindent=0pt}

\newcommand{\red}[1]{\textcolor{black}{#1}}
\newcommand{\blue}[1]{\textcolor{black}{#1}}

\newcommand{\enquote}[1]{``#1''}

\newcommand{\sref}[1]{Sec.~\ref{#1}}
\newcommand{\aref}[1]{App.~\ref{#1}}

\newcommand{\tref}[1]{Table~\ref{#1}}
\newcommand{\trefplural}[1]{Tables~\ref{#1}}
\newcommand{\fref}[1]{Fig.~\ref{#1}}
\newcommand{\frefplural}[1]{Fig.~\ref{#1}}

\definecolor{anne}{rgb}{0.635,0.998,0.722}
\definecolor{sophie}{rgb}{0.998,0.722,0.635}
\definecolor{alex}{rgb}{0.722,0.635,0.998}

\definecolor{UniBlue}{rgb}{0.635,0.998,0.722}

\newcommand{\fscore}{F\textsubscript{1}\xspace}

\newcommand{\macsexplanation}{macX: macro scores computed on all labels. macXs: macro scores computed only on labels that occur in the respective evaluation set.}
\newcommand{\exactmatch}{Exact Match\xspace}

\newcommand{\randomrunsdescription}{Averages and standard deviations of 5 runs with different random seeds.}
\newcommand{\lossdescription}[1]{BCE: binary cross-entropy loss; WBCE: weighted binary cross-entropy loss using weights as described in \sref{sec:models}; FL: focal loss#1; OS: oversampling percentage.}
\newcommand{\experimentalsetting}{\randomrunsdescription{} \lossdescription{}}
\newcommand{\rcvsubsampled}{The subsampled test set contains 5{,}000 instances and closely matches the label distribution of the original test set (see \aref{sec:appendix_datasets}).
}

\newcommand{\hidecdaggerinfo}{$\dagger$: best out of 50 training epochs (no early stopping), matching \citet{kim-etal-2024-hierarchy}'s experimental setting. $\dagger \dagger$: best out of 100 training epochs (no early stopping).}

\newcommand{\frequentlabels}{$\geq 1000$ instances in training.}
\newcommand{\mediumlabels}{100 to 999 instances in training.}
\newcommand{\rarelabels}{< 100 instances in training.}
\newcommand{\frequencygroupstatistics}[2]{#1 labels, with $\varnothing$ #2 instances in evaluation.}

\newcommand{\rcvtestfrequentlabels}{\frequencygroupstatistics{17}{100456.9}}
\newcommand{\rcvtestmediumlabels}{\frequencygroupstatistics{50}{15356.9}}
\newcommand{\rcvtestrarelabels}{\frequencygroupstatistics{36}{1598.8}}

\newcommand{\rcvtestsubsampledfrequentlabels}{\frequencygroupstatistics{17}{699.7}}
\newcommand{\rcvtestsubsampledmediumlabels}{\frequencygroupstatistics{50}{109.8}}
\newcommand{\rcvtestsubsampledrarelabels}{\frequencygroupstatistics{36}{10.6}}

\newcommand{\rcvdevfrequentlabels}{\frequencygroupstatistics{14}{333.6}}
\newcommand{\rcvdevmediumlabels}{\frequencygroupstatistics{52}{48.6}}
\newcommand{\rcvdevrarelabels}{\frequencygroupstatistics{37}{4}}

\newcommand{\mimictestfrequentlabels}{\frequencygroupstatistics{144}{206.2}}
\newcommand{\mimictestmediumlabels}{\frequencygroupstatistics{877}{24.9}}
\newcommand{\mimictestrarelabels}{\frequencygroupstatistics{7909}{1.2}}

\newcommand{\mimicdevrarelabels}{\frequencygroupstatistics{7935}{0.5}}

\newcounter{example}

\usepackage{array}
\newcolumntype{L}[1]{>{\raggedright\let\newline\\\arraybackslash\hspace{0pt}}m{#1}}
\newcolumntype{C}[1]{>{\centering\let\newline\\\arraybackslash\hspace{0pt}}m{#1}}
\newcolumntype{R}[1]{>{\raggedleft\let\newline\\\arraybackslash\hspace{0pt}}m{#1}}

\newcommand{\ourScheme}{adaptive\textsubscript{ML}\xspace}
\newcommand{\ourSchemeCapitalized}{Adaptive\textsubscript{ML}\xspace}

\DeclareMathOperator*{\argmax}{argmax} %
\DeclareMathOperator*{\ece}{ECE} %

\usepackage{enumitem}

\title{How to Estimate Whether You Have Found Several Needles in a Haystack:
Measuring Calibration in Multi-Label Text Classification}

\author{Sophie Henning$^{1,2,3\dagger}$ \hspace{5mm} Georg Hofmann$^{4}$ \hspace{5mm} Alexander Schulte$^{5}$ \\ \textbf{Alexander Fraser}$^{1,2,6}$ \hspace{5mm} \textbf{Annemarie Friedrich}$^{4}$ \\
 $^1$TU Munich
 $^2$Munich Center for Machine Learning (MCML)
 $^3$LMU Munich\\
 $^4$University of Augsburg
 $^5$Robert Bosch GmbH
 $^6$Munich Data Science Institute\\
  \texttt{sophie.henning|alexander.fraser@tum.de} \\
  \texttt{georg.hofmann|annemarie.friedrich@informatik.uni-augsburg.de}\\
}

\begin{document}
\maketitle
\begin{abstract}
A key factor in deciding whether to trust an automatic prediction is its \textit{confidence score}, which should be \textit{calibrated} to match the actual probability of the prediction being correct.
Most confidence calibration metrics target binary or multi‑class tasks, while multi‑label calibration remains largely underexplored.
Multi-label classification tasks, such as assigning medical codes to clinical notes or determining news topics, are usually dominated by a large number of \textit{negatives}, i.e., labels that do \textit{not} apply. 
We show that existing binning schemes to compute label-wise expected calibration error either underestimate the error, simply reflect label frequency, \blue{or suffer from many bins with very few instances}.
To achieve trustworthy label-wise calibration errors,
we propose
a new binning scheme that gives equal weight to positive and negative label assignments. 
Our empirical study demonstrates that in contrast to existing binning schemes, our new scheme results in meaningful estimates of calibration error in hierarchical and in extreme multi-label classification.
 We also show that calibrating confidence scores of large language models for multi-label predictions is an open challenge.
 Our detailed analysis lays the foundation for further research %
 by providing a solid evaluation metric for measuring calibration in multi-label classification.
\end{abstract}

\section{Introduction}
Many text classification tasks, such as medical coding, involve assigning multiple labels to each instance.
To enable humans to decide when (not) to trust machine predictions, prediction probabilities must be calibrated so that confidence scores reflect  the empirical probability of the respective prediction being correct.\unmarkedfntext{$^\dagger$Work done while at Bosch Center for Artificial Intelligence, Renningen, Germany.}

Existing calibration metrics largely address binary and multi-class classification \citep{menezes2023calibration_survey} and typically rely on binning predictions by their confidence. %
Prior work on calibration in multi-label classification \citep{DBLP:journals/tip/ChenWPQYLL24, DBLP:conf/cvpr/ChengV24, DBLP:journals/corr/abs-2411-04276} %
computes calibration errors on the union of all binary per-label 
predictions
using standard calibration metrics for binary classification.
Importantly, this requires that binning is performed only once for the joint set of all label confidences.
However, this \textbf{joint binning can conceal %
miscalibration of individual labels} (see \sref{sec:multilabel_metrics}).
This poses problems if a user trusts the confidence estimate of a specific label, e.g., a  medical code, because the classifier seems well-calibrated \textit{across} labels, but is actually poorly calibrated on that specific label. %
\textbf{We therefore propose to separately bin each label in multi-label classification problems.}

\begin{figure*}
  \begin{center}
    \begin{subfigure}[t]{0.32\textwidth}
      \centering
      \includegraphics[width=\linewidth]{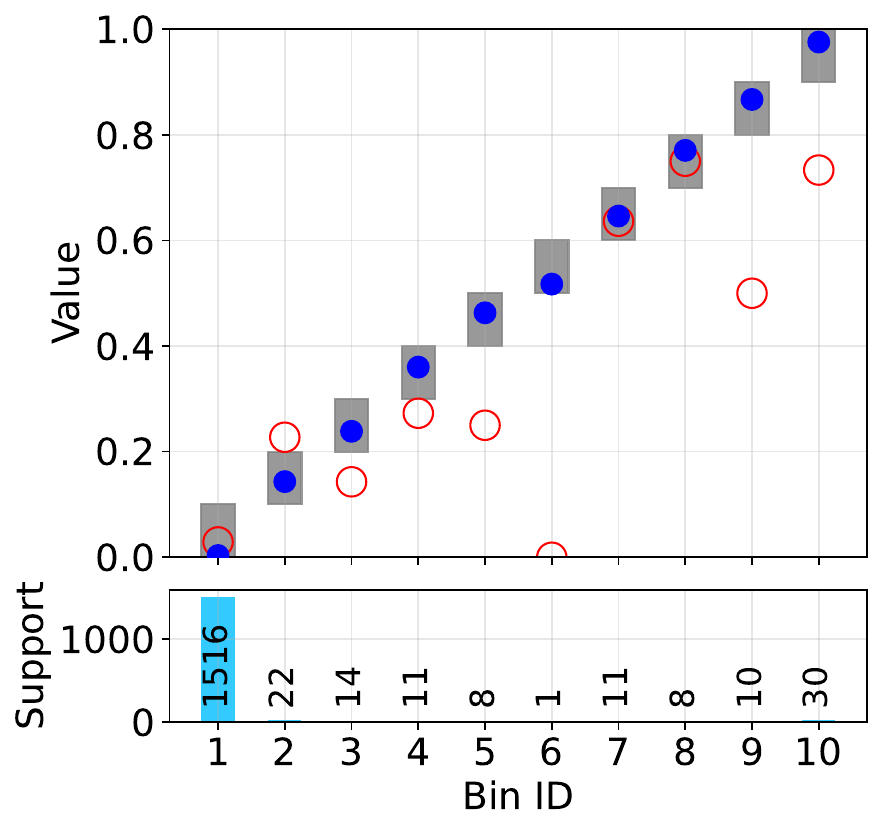}
      \caption{Fixed-width \citep{DBLP:conf/aaai/NaeiniCH15}}
      \label{fig:binning_fixed_width}
    \end{subfigure}
    \hfill
    \begin{subfigure}[t]{0.32\textwidth}
      \centering
      \includegraphics[width=\linewidth]{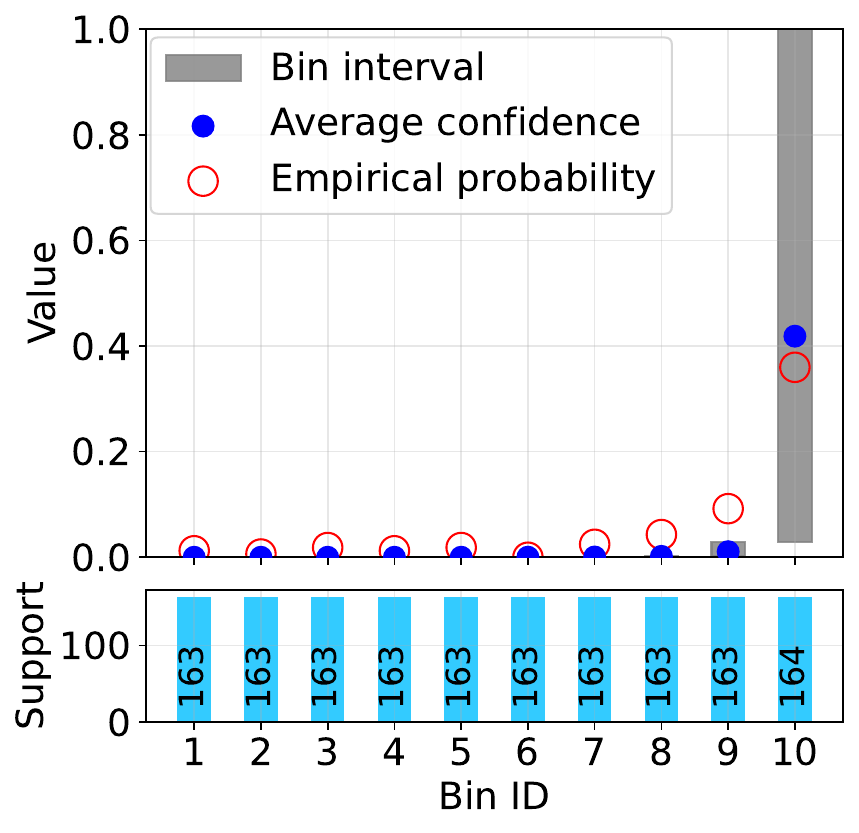}
      \caption{Adaptive \citep{DBLP:conf/cvpr/NixonDZJT19}}
      \label{fig:binning_adaptive}
    \end{subfigure}
    \hfill
    \begin{subfigure}[t]{0.32\textwidth}
      \centering
      \includegraphics[width=\linewidth]{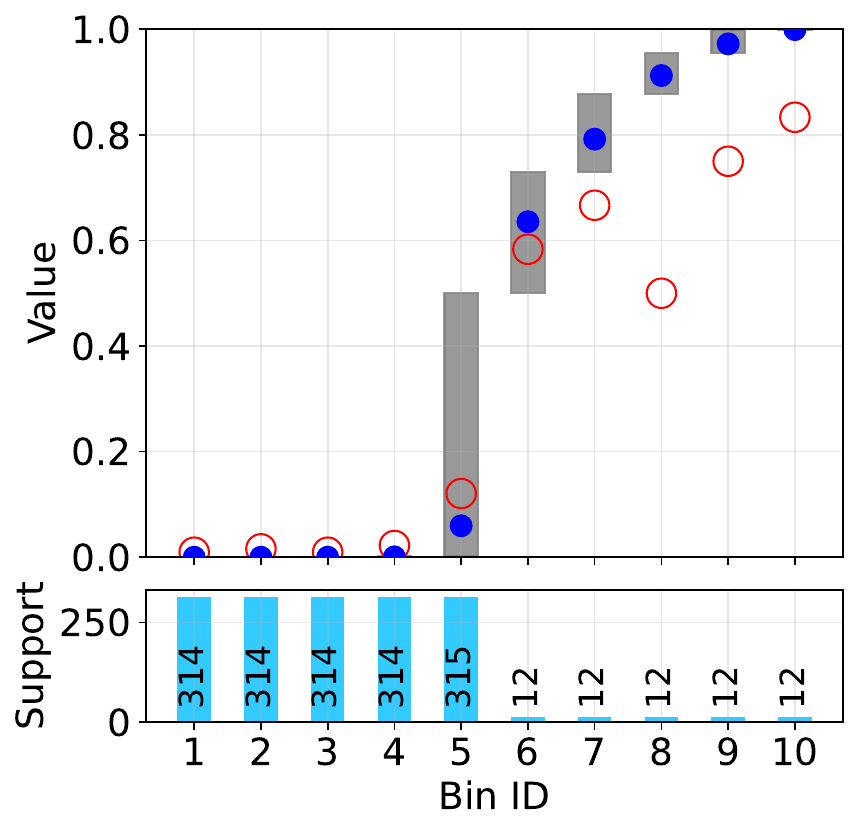}
      \caption{Ours: \ourSchemeCapitalized}
      \label{fig:binning_adaptive_05}
    \end{subfigure}
  \end{center}
  \caption{\textbf{Comparison of binning schemes} \blue{for an individual label} in multi-label classification. \textit{Average confidence} is the mean predicted probability for the label within a bin, and \textit{empirical probability} is the fraction of instances in that bin that actually have the label.\protect\footnotemark  Fixed-width binning results in several bins with only very few examples (see \enquote{Support}, i.e., the bin size), whereas adaptive binning leads to most bins containing predictions close to 0 and a very wide bin for the remaining predictions. We propose the new binning scheme \ourScheme, which strikes a balance between the two extremes, giving equal weight to the cases where the classifier assigns or does not assign the label.
  The example illustrates the confidence scores of the predictions of a fine-tuned BiomedBERT classifier for the label \enquote{33.24} in MIMIC-III-v.1.4.}
  \label{fig:binning_schemes_mimic}
\end{figure*}

A standard binning scheme for estimating calibration errors is fixed-width binning, which splits the probability interval $[0, 1]$ into $n$ evenly spaced intervals (see grey boxes in  \fref{fig:binning_fixed_width}).
However, for individual labels in multi-label classification, the number of \textit{negative instances}, i.e., where the label does not apply, typically exceeds that of \textit{positive instances} by far, and thus models usually predict that a label does not apply (\textit{negative prediction}) much more often than they predict that a label does apply (\textit{positive prediction}).
Hence, \textbf{fixed-width binning suffers from a large number of bins with very few instances} (see \enquote{Support} in \fref{fig:binning_fixed_width} and \sref{sec:experiments}), making the resulting calibration error estimate unreliable.
Adaptive binning \citep{DBLP:conf/cvpr/NixonDZJT19} constructs %
bins of approximately the same number of instances (see \fref{fig:binning_adaptive}).
With many more negative than positive predictions, \textbf{adaptive binning puts all positive predictions into a single bin} (see \fref{fig:binning_adaptive}, where most bins have an average confidence close to 0, and the last bin covers a very wide prediction interval).
Thus, the resulting calibration error (the average of the differences between the average confidence and empirical probability, i.e., the fraction of positive instances in each bin) only marginally considers positive predictions.

We hence propose a \textbf{new binning scheme for multi-label (ML) classification (\ourScheme)} .
Our core insight is that a reliable and practically relevant binning scheme for multi-label prediction must put equal weight on positive and negative predictions.
We hence apply adaptive binning separately to positive and negative predictions,
creating only as many bins as the smaller subset (usually positive predictions) can support while ensuring a minimum bin size (see \fref{fig:binning_adaptive_05}, where the first five bins contain only negative predictions and the last five bins contain only positive ones). %
Based on \ourScheme, we propose the multi-label expected calibration error (\textbf{$\text{ECE}_{\text{ML}}$}), an unweighted average of the calibration errors of the individual bins (\sref{sec:multilabel_metrics}).
Empirically, \ourScheme binning scheme yields well-populated bins (\sref{sec:experiments}) %
and thus a meaningful calibration metric for the multi-label case.

\footnotetext{Calibration error is often visualized with reliability diagrams \citep[e.g.,][]{menezes2023calibration_survey}, showing the average prediction confidence %
on the linearly scaled x-axis and the empirical probability (fraction of positives) on the y-axis. 
A linearly scaled x-axis makes sense with fixed-width binning schemes.
However, with adaptive schemes, bin width can vary drastically.
This is why we use bin IDs on the x-axis instead and visualize the interval covered by the respective bin using grey boxes aligned with the y-xis.
}

We validate our binning scheme and $\text{ECE}_{\text{ML}}$ in an empirical study on two heavily imbalanced large-scale multi-label text classification benchmarks: RCV1-v2 \citep{lewis2004rcv1}, which \blue{provides a hierarchical label set, and MIMIC-III-v1.4 \citep{johnson2016mimicv1.4}, which has a very large label set.}
We evaluate  %
BERT-style models \citep{devlin-etal-2019-bert} and specialized models for the two datasets \citep{kim-etal-2024-hierarchy, huang-etal-2022-plm} in conjunction with methods to tackle class imbalance, as well as LLM-based classifiers in zero-shot and retrieval-augmented generation settings  \blue{\citep[RAG, similar to][]{milios-etal-2023-context}}, which may be a viable alternative to fine-tuned models if large-scale training data is not available.

In sum, our contributions are:

\begin{enumerate}[label=(\arabic*), leftmargin=*, nosep]
    \item We perform an in-depth analysis of pitfalls in directly applying single-label calibration metrics to the multi-label setting.
    \item As a remedy, we propose \ourScheme, a robust new binning scheme for assessing per-label classification in multi-label classification. %
    \item Our empirical study on two large %
    benchmarks shows that fine-tuned discriminative classifiers with focal loss achieve the best performance \textit{and} calibration.
    \item \blue{We find that fine-tuned classifiers outperform zero-shot LLM-based predictors. Although RAG considerably improves both performance and calibration of LLMs, they still fall short of fine-tuned classifiers and have a considerably higher inference time.}
\end{enumerate}

We argue that more research is necessary on the practically relevant setup of using LLMs for multi-label classification. Our newly proposed and validated binning scheme offers a solid foundation for future research in this direction.\footnote{We make our code available under a permissive license at \url{https://github.com/sophiehenning/multilabel-classification-calibration}.}

\section{Background and Related Work}
In this section, we review single-label calibration metrics, %
and give a brief overview on prior work on calibration and class imbalance in NLP. %

\subsection{Single-label Calibration Metrics}
For multi-class classification with $C$ classes, different notions of calibration have been proposed %
\citep{posocco2021estimating_eces}. %
\textit{Confidence calibration} only considers the top-1 prediction:

\begin{gather} 
\forall p \in \left[ 0, 1 \right]:
P(Y=\hat{Y} \mid \hat{p} = p )~=~p \label{eqn:confidence-calibration}
\end{gather}

$(X,Y)$ is a random variable from which inputs and labels are drawn, $f:X\rightarrow [0,1]^C$ is the function learnt by the model, $\hat{Y} = \argmax_{c\in \left[ 1..C \right]} \left( f\left( X \right)_c \right)$ is the top-1 predicted label and $\hat{p} = \max_{c\in \left[ 1..C \right]} \left( f\left( X \right)_c \right)$ is the \textit{confidence}, i.e., the probability of the top-1 prediction.

The stricter \textit{class-wise calibration} \citep{zadrozny2002class-wise_calibration} %
requires all per-class probabilities %
to be calibrated in a one-vs-rest fashion:
\begin{gather} 
    \forall c \in \left[ 1..C \right] \; \forall p \in \left[ 0, 1 \right]: \notag\\ P\left(Y=c \mid  f\left( X \right)_c = p  \right)~=~p\ \label{eqn:class-wise-calibration}
\end{gather}  %
A popular way to operationalize \cref{eqn:confidence-calibration,eqn:class-wise-calibration} is to discretize the probability interval $[0,1]$ into $M$ bins.
For confidence calibration, this leads to the Expected Calibration Error (\textbf{ECE}) \citep{DBLP:conf/aaai/NaeiniCH15, guo2017calibration}:
\begin{gather} 
    \ece = \sum_{m=1}^M \frac{|B_m|}{n} \left| \text{acc}(B_m) - \text{conf}(B_m)\right| \label{eqn:ece}
\end{gather}
where $n$ is the total number of instances, and $|B_m|$, $\text{acc}(B_m)$, and $\text{conf}(B_m)$ are the \textit{support}, the accuracy, and the averaged confidence of the $m$-th bin, respectively. %
Class-wise ECE (\textbf{CWECE}) %
performs the binning separately for each class:
\begin{gather} \label{eqn:cwece}
    \text{CWECE} = \frac{1}{C} \sum_{c=1}^C \text{ECE}_c
\end{gather}
$\text{ECE}_c$ is the ECE for label $c$ (see below) and $\text{share}_c(B_m)$ is the percentage of instances in the bin $m$ that carry label $c$ and $ \bar{p}_c(B_m)$ is the average of the respective prediction probabilities for $c$.
\begin{gather}
\text{ECE}_c = \sum_{m=1}^M \frac{|B_m|}{n} \left| \text{share}_c(B_m) - \bar{p}_c(B_m)\right|
\end{gather}

\blue{Our metric for estimating confidence in multi-label classification tasks is derived from these metrics.}

\subsection{Calibration and Class Imbalance in NLP}%
\textbf{Calibration in NLP.} 
Most prior work on classifier calibration in NLP focuses on single-label tasks.
Here, BERT-style models have been shown to be poorly calibrated, often suffering from overconfidence \citep{desai-durrett-2020-calibration, kong-etal-2020-calibrated, guo-etal-2021-overview, kim-etal-2023-bag}.
\citet{sachdeva-etal-2024-catfood} improve calibration of such models on extractive question-answering by counterfactually augmenting their training data.
\blue{Calibration of generative LLMs is an open research topic \citep[see, e.g.,][]{geng-etal-2024-survey, DBLP:journals/jcst/ZhangW26}.
Similar to discriminative models, confidence estimates can be derived from logits, but generative models can also be prompted to verbalize their confidence.
More advanced calibration techniques rely on the grouping of similar prompts and completions \citep[e.g.,][]{DBLP:conf/icml/DetommasoLF024}.
}

\noindent \textbf{Classifier Calibration on Imbalanced Data.}
\citet{kranzlein-etal-2021-making-heads} investigate single-label calibration with rare classes, proposing to evaluate calibration separately for classes with similar frequency and to also perform post-hoc recalibration based on these groups.
Focal loss has been successfully applied to a variety of imbalanced tasks in both computer vision and NLP \citep[e.g.,][]{MukhotiKSGTD20,wang-etal-2022-calibrating, GhoshSG22, LiuRGDA23, YilmazKKDK23}.
\citet{henning-etal-2023-survey} give an overview on class imbalance methods in NLP.

We take inspiration from this literature to combine several promising sampling methods and loss functions when fine-tuning encoder models for multi-label classification.

\section{Issues when Computing Calibration for Multi-Label Classification}
\label{sec:relwork_multilabel}

In multi-label classification, $Y$ is a random variable in $\left\{0,1\right\}^C$, with $Y_c = 1$ indicating that class $c$ applies (we call this assignment of 0 or 1 to a dimension in the label vector a \textit{label assignment}).
One may model the problem at the level of (i) individual labels, yielding a series of $C$ binary classification problems, or (ii) the full label vector, yielding a prediction problem over subsets of labels.

\blue{Historically, classical machine learning methods using features derived from word counts \citep[see, e.g.,][]{DBLP:journals/information/KowsariMHMBB19} were applied for multi-label text classification. 
Computationally cheap methods like decision trees enable building higher-order classifiers (considering relations between labels) \citep[see, e.g.,][]{DBLP:journals/tcyb/ZhangLYL22}.
Neural classifiers based on Transformer encoders like BERT \citep{devlin-etal-2019-bert} outperform count-based approaches \citep{galke2022we}, typically using a single sigmoid head for each label instead of higher-order classifiers due to their computational expense.
With the rise of powerful models to generate text, multi-label classification can also be approached using generative models (see \sref{sec:models}).}

The different modeling choices (i) and (ii) naturally lead to different notions of calibration. 
Under (ii), a predicted set is correct if and only if it exactly matches the ground truth set of applying labels \citep{li2019multilabel}, %
without distinguishing near-correct from highly deviant predictions.
For models that output per-label probabilities, %
a standard way to compute prediction set probability is to compute it as the product of the per-label prediction probabilities %
for inclusion or exclusion \citep[e.g.,][]{li2019thesis}.
In large label spaces, prediction set probabilities computed this way become infinitesimally small, making calibration evaluation very hard (see \aref{sec:appendix_multilabel_metrics}).

\blue{If a multi-label problem with $C$ labels is modeled at the level of individual labels (i), considering the calibration of the $C$ individual label assignments per instance is a natural choice.}
Put more formally, this requires multi-label classifiers to be calibrated on the level of per-label decisions:
\begin{gather} 
    \forall p \in \left[ 0, 1 \right] \; \forall c \in \left[ 1..C \right] :\notag\\ P\left(Y_c = 1 \mid  f\left( X \right)_c = p  \right) = p\ \label{eqn:multi-label-calibration}
\end{gather}
Prior work on calibration in multi-label classification in computer vision \citep{DBLP:journals/tip/ChenWPQYLL24, DBLP:conf/cvpr/ChengV24} has approximated \Cref{eqn:multi-label-calibration} by %
computing calibration errors on the union of all binary per-label classifications.
This means that binning is performed only once for all $n \cdot C$ label assignments in a multi-label classification task with $n$ instances and $C$ labels.
While %
intuitive at first glance, we argue that this procedure often leads to undesired effects of miscalibrated labels concealing the miscalibration of each other, e.g., an overconfident label concealing the underconfidence of another one, since binning happens before computing the error (see \fref{fig:binning_overconfidence_concealing_underconfidence}) in \aref{sec:appendix_multilabel_metrics}).
In practice, this means that a user might put unjustified trust in the prediction of a specific label as the classifier seems well-calibrated \textit{across} labels, but is actually poorly calibrated on that specific label.

\blue{\citet{DBLP:journals/corr/abs-2411-04276} discuss computing per-label calibration errors, but observe that this may be distorted by the large amount of easily recognizable non-applying labels (\textit{easy negatives}) and propose to instead evaluate the calibration of the $k$ most confident labels per instance.} 
For computing an error on these labels, they perform adaptive binning only once, again potentially inducing undesired effects of one miscalibration concealing the other.
Moreover, it is unclear how to choose $k$, as the standard deviation of the  number of assigned labels per instance can be rather substantial (e.g., 8.14 on MIMIC-III-v1.4, see \tref{tab:datasets} in \aref{sec:appendix_datasets}). 
With a too low $k$, we ignore too many relevant labels, and with a high $k$, this score converges towards binning all $n \cdot C$ label assignments at once.

\section{$\text{ECE}_{\text{ML}}$: A New Calibration Metric for Multi-label Classification}
\label{sec:multilabel_metrics}
\blue{In this section, we derive our proposed method to evaluate the calibration of individual labels in multi-label calibration.
We opt for computing the calibration error of each individual binary classification separately for the reasons discussed in \sref{sec:relwork_multilabel},} enabling displaying label predictions with their confidence score and per-label calibration error.

\blue{We propose to address the issue of many easy negative labels} %
on the level of the binning scheme. %
ECE as defined in \Cref{eqn:ece} relies on a fixed binning scheme, %
in which bins may contain strongly varying numbers of predictions (see, e.g., \fref{fig:binning_fixed_width}).
This can lead to biased estimators \citep{DBLP:conf/aistats/RoelofsCSM22}. 
However, applying adaptive binning \citep{DBLP:conf/cvpr/NixonDZJT19}, which ensures roughly equal bin sizes,  to the binary subtasks of a multi-label classification problem,
leads to the problem that most of the bins will only contain predictions close to 0. The remaining bin(s) will be very wide, conflating predictions with high confidence and those with low confidences (see, e.g., \fref{fig:binning_adaptive}). 
Put differently, standard adaptive ECE implicitly allocates resolution according to prediction density, which becomes pathological in multi-label settings.
One way to tackle this issue is to ignore all predictions below some threshold $\epsilon$ \citep{DBLP:conf/cvpr/NixonDZJT19}, but there is no obvious way to choose $\epsilon$, whose choice will greatly influence the resulting score.
Moreover, the method does not detect miscalibration below the threshold, i.e., if a classifier assigns probabilities below $\epsilon$ when the label actually applies.

\noindent \textbf{\ourSchemeCapitalized Binning.}
Instead of thresholding, we propose to separately bin positive and negative predictions in an adaptive scheme, respectively.
We assume that each label receives a confidence score separately and that a prediction is positive if the corresponding score $\geq 0.5$. 
If a model achieves higher overall accuracy by predicting a label at a threshold different from 0.5, its confidences are not well-calibrated, and they can be scaled to the 0.5 threshold before applying \ourScheme as we exemplify in \sref{sec:models}.
However, if a threshold different from 0.5 is chosen for other reasons (e.g., safety concerns), we advise to not re-scale the original confidences, as the thresholds were not optimized during training. 
It then depends on the use case if separate binning should be performed at the 0.5 threshold (assessing general model calibration) or at the chosen threshold (answering how well the model is calibrated for positive and negative predictions, respectively).
Different models should always be compared at the same threshold.

Binning positive and negative predictions separately enables the calibration error to cover all predictions while \red{giving equal weight to the two scenarios that are of interest to the user when deciding whether to trust a classifier: The label-wise calibration error should reflect (a) how good confidence scores are for the label if the label is predicted, and (b) how good confidence scores are if the label is not predicted.}
If the overall number of bins is specified to be $b$, we reserve $\frac{b}{2}$ bins for the two intervals, respectively. 
We additionally constrain our binning procedure with a minimum bin size $b_{min}$.  
If there are not enough positive or negative predictions for $\frac{b}{2}$ bins of at least $b_{min}$ size, we scale down the number of bins to ensure that each bin contains at least $b_{min}$ predictions.
In the special case of having fewer positive or negative predictions than $2*b_{min}$, we %
use a single bin for positive and negative predictions, respectively.\footnote{\blue{See \aref{sec:appendix_multilabel_metrics} for a discussion of this design choice.}}
If a label is never predicted by the classifier, we bin the negative predictions adaptively into $b$ bins.
In our experiments, we use $b_{min}=5$ and $b=10$.\footnote{We perform a sensitivity analysis for these hyperparameters in \sref{sec:experiments}.}
We call the resulting scheme \textbf{\ourScheme}. %
\fref{fig:binning_schemes_mimic} exemplifies the results of the different binning schemes on a single label. 

\noindent \textbf{Assessing Overall System Calibration.} To assess overall system calibration, we report an average of the per-label calibration errors, indicating by how many percentage points labels are miscalibrated on average.
Our proposed metric \textbf{$\text{ECE}_{\text{ML}}$} indexes bins $m$ additionally by label $c$:
\begin{gather} 
    \text{ECE}_{\text{ML}} = \frac{1}{C} \sum_{c=1}^C \frac{1}{M_c} \sum_{m=1}^{M_c}  \left| \text{acc}(B_{cm}) - \bar{p}(B_{cm})\right|   \label{eqn:ece_multi-label}
\end{gather}
Here, $M_c$ is the number of bins for the class $c$ (usually, $M_c = b$, see above for special cases).
We compute acc as the percentage of instances carrying label $c$ in the bin $B_{cm}$, and $\bar{p}(B_{cm})$ is the average of the prediction probabilities of label $c$.

\section{Experimental Setting}
\subsection{Datasets}
We conduct an empirical study to validate our new metric on RCV1-v2 \citep[][henceforth \textbf{RCV1}]{lewis2004rcv1} and MIMIC-III-v1.4 \citep[][henceforth \textbf{MIMIC-III}]{johnson2016mimicv1.4}, two large complex multi-label text classification benchmarks, which we describe in further detail in this section.
We compute micro- and macro-averaged \fscore, with macro \fscore being our key metric due to the imbalanced nature of the two datasets.
\blue{Similar to \citet{DBLP:conf/cvpr/0002MZWGY19}, we additionally compute performance and calibration metrics separately for highly frequent, medium-frequency, and rare labels (with $n \geq 1000$, $100 \leq n < 1000$, and $n<100$ positive instances, respectively\footnote{\citet{DBLP:journals/corr/abs-2402-12819} found Transformer encoder-based models to outperform generative models on binary classification tasks with 100-1000 training examples.}).}
For dataset statistics including subset statistics, see \aref{sec:appendix_datasets}.

\noindent \textbf{RCV1.} RCV1 is a %
hierarchical news topic classification dataset with 103 unique labels and roughly 800,000 instances.
We use the official RCV1 test set (ca. 780,000 instances) and randomly split the official training set into a train, tune, and a dev set \citep{van-der-goot-2021-need}.
For our exploratory experiment with generative classifiers, due to their much higher runtime, we create a  subsampled version of the RCV1 test set containing 5,000 instances (for sampling details, see \aref{sec:appendix_datasets}).
We compute hierarchical \fscore scores \citep{kiritchenko2005functional}.

\noindent \textbf{MIMIC-III.} MIMIC-III is a %
medical coding dataset, i.e., the task is to assign codes (labels) to free-text medical notes with 8,930 unique labels and roughly 53,000 instances.
Following \citet{mullenbach-etal-2018-explainable}, we predict both diagnosis and procedure codes and use their data split into training, dev, and test set.
We further randomly split the training set into a train and tune set and use \citet{kim-etal-2022-anemic}'s code preprocessing, but no text preprocessing.
While the codes are hierarchical, the task of human medical coders is to always assign the most specific code.\footnote{See Section B of the \href{https://stacks.cdc.gov/view/cdc/140851}{ICD-9-CM official guidelines for coding and reporting}.}
We thus use the standard \fscore score here.
We follow \citet{DBLP:conf/sigir/EdinJHBMRM23} in computing macro-averaged scores only on labels that occur in the respective evaluation set, and report macro-averages on all labels in \aref{sec:appendix_experiments_results}.

\subsection{Models}
\label{sec:models}

Using our newly proposed method for evaluating calibration in multi-label classification, we compare a wide range of neural text classification methods.
Here, we describe the various setups we use for our models, including loss- and sampling-based methods for mitigating class imbalance effects, %
and obtaining confidence scores from LLMs.

\noindent \textbf{Classifiers.}
We use BERT-based classifiers \citep{devlin-etal-2019-bert} and two SOTA model architectures specifically designed for RCV1 and MIMIC-III, respectively.
For RCV1, in addition to BERT, we evaluate HiDEC \citep{Im_Kim_Oh_Jo_Kim_2023}.
HiDEC uses an encoder-decoder setup to generate a sequence of labels representing a path through the label hierarchy.
As our baseline, we train HiDEC with the HBM loss \citep{kim-etal-2024-hierarchy}, a hierarchy-aware loss function specifically designed for HiDEC-like models that also optimizes label-specific prediction thresholds.
To enable comparison with confidence values from other models, %
we piecewise linearly re-scale the label-specific prediction confidences (\textbf{HiDEC-s}, see \aref{sec:appendix_modeling_hidec}).

On MIMIC-III, we first compare BERT, BiomedBERT \citep{gu2022biomedbert}, and ModernBERT \citep{warner2024modernbert}.
For BERT and BiomedBERT, we need to truncate the long medical notes due to their limited context windows, whereas ModernBERT can fit 8,192 tokens. 
Yet, the domain-specific BiomedBERT model is the best model on the dev set  (see \tref{tab:mimic_dev_performance}).
Hence, we use this model as our underlying model for PLM-ICD  \citep{huang-etal-2022-plm}, a model based on chunking and label attention.

\noindent \textbf{Addressing Class Imbalance.}
Our starting point are BERT-based baselines trained with binary cross-entropy loss (\textbf{BCE}). %
To tackle class imbalance, we test variants of up-weighting positive instances compared to negative ones: uniformly up-weighting them \citep[\textbf{WBCEU}, ][]{rathnayaka2019gated}, up-weighting \textit{rare} positive instances proportionally to class frequency (\textbf{WBCEM}), and up-weighting \textit{all} positive instances proportionally to class frequency (\textbf{WBCEP}).
We also evaluate Multilabel Random Oversampling \citep[\textbf{ROS,}][]{charte2015mlros}, which %
assigns more importance to less frequent labels by oversampling them, and focal loss \citep[\textbf{FL,}][]{LinGGHD17}, which down-weights instances for which the model %
is already confidently correct (see \aref{sec:appendix_modeling_imbalance} for loss formulae).

\noindent \textbf{Multi-Label Classification with LLMs.}
Instead of treating multi-label classification as a standard classification problem with discriminative models, it can also be approached using generative models, ranging from fine-tuning methods \citep[e.g.,][]{jung-etal-2023-cluster} to few-shot prompting
\citep[e.g.,][]{DBLP:journals/corr/abs-2401-12178, milios-etal-2023-context, zhu-zamani-2024-icxml}. %
Inspired by these approaches, we design an LLM-based baseline system for multi-label prediction. 
Our system utilizes Llama-3.3-70B-Instruct \citep[][\textbf{Llama-3.3}]{DBLP:journals/corr/abs-2407-21783} or Qwen2.5-72B-Instruct \citep[][\textbf{Qwen-2.5}]{DBLP:journals/corr/abs-2412-15115}, which for each prediction is prompted with a set of few-shot demonstrations from the training set, which are retrieved from a vector database \citep[FAISS,][]{DBLP:journals/corr/abs-2401-08281} if the cosine similarity between their embeddings and the embedding of the test instance, computed using all-mpnet-base-v2\footnote{\url{https://huggingface.co/sentence-transformers/all-mpnet-base-v2}} \citep[]{reimers-gurevych-2019-sentence}, exceeds a tuned similarity threshold. 
We compare this retrieval-enhanced system with zero-shot prompting of the underlying models.
For more details, see \aref{sec:appendix_modeling_llms}.
 
Invalid predicted labels (not contained in the label space) are discarded.
To estimate confidence, we compare (i) extracting the probability of the first generated token corresponding to each label to determine the model's confidence (\textbf{Token Prob.}) and (ii) asking the LLM to estimate its confidence in individual labels (\textbf{Verbalized}, see \aref{sec:appendix_modeling_llms}).

\begin{figure*}[htb]
  \begin{center}
    \begin{subfigure}[t]{0.9\textwidth}
      \centering
      \includegraphics[width=\linewidth]{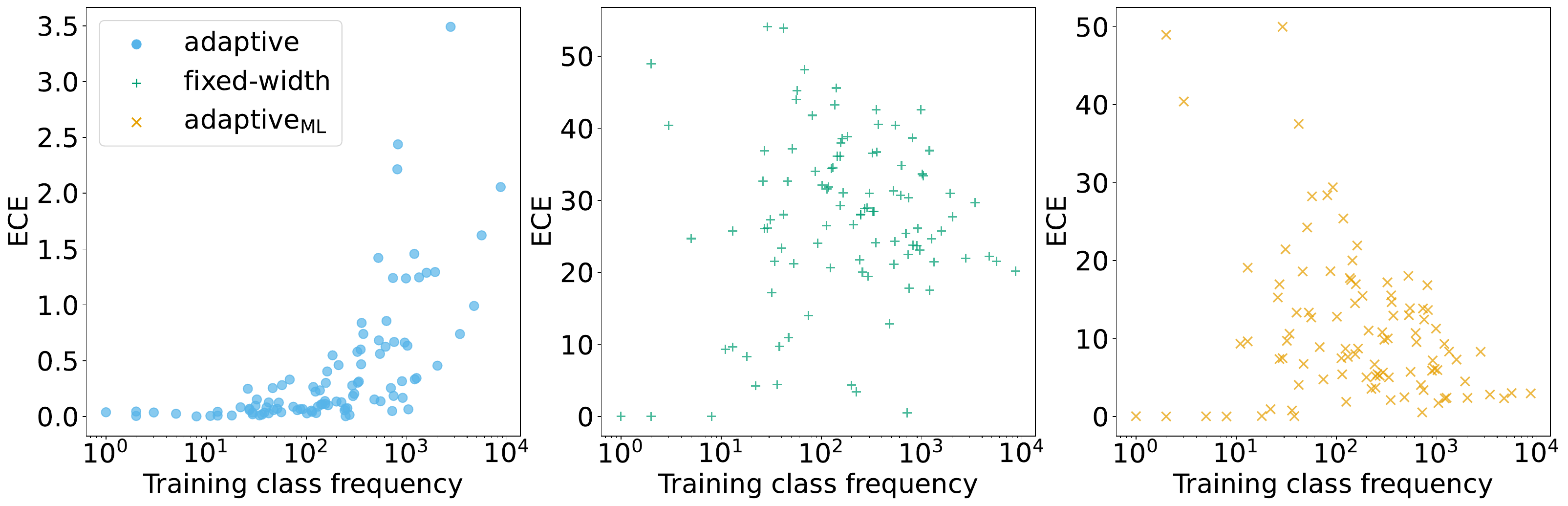}
      \caption{RCV1 dev} %
      \label{fig:ece_by_scheme_rcv_dev}
    \end{subfigure}
    
    \begin{subfigure}[t]{0.9\textwidth}
      \centering
      \includegraphics[width=\linewidth]{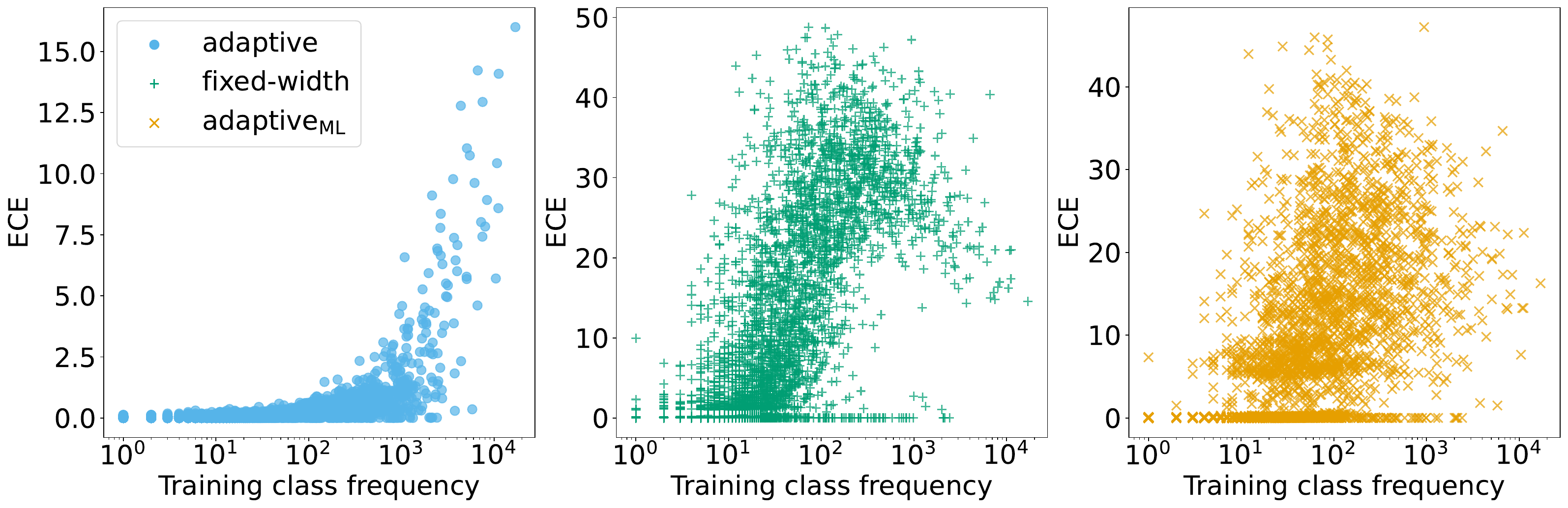}
    \caption{MIMIC-III dev}    %
      \label{fig:ece_by_scheme_mimic_dev}
    \end{subfigure}
  \end{center}
  \caption{\textbf{Expected Calibration Error (ECE) as a function of training class frequency under different binning schemes} (adaptive, fixed-width, and \ourScheme). Results are shown for (a) RCV1 dev using BERT-base and (b) MIMIC-III dev using BiomedBERT-base, both trained with BCE loss and no oversampling, averaged over 5 random seeds. \blue{Under adaptive binning, ECE increases with class frequency on both datasets, whereas under \ourScheme binning, ECE exhibits opposing trends with respect to class frequency on the two datasets.} For ECEs by training frequency on the test sets and for performance by training frequency, see \frefplural{fig:ece_by_scheme_test} and \ref{fig:performance_frequency}, respectively. }
  \label{fig:ece_by_scheme}
\end{figure*}

\subsection{Setup}

On both datasets, we run the experiments with five different random seeds, respectively, and report averages and standard deviations, with the exception of the LLM-based explorative experiments, where we greedily decode the outputs of a single run.
During fine-tuning, we train for 50 epochs.
If not otherwise indicated, we stop early with a patience of ten epochs based on the tune/dev set performance in development/evaluation runs.
When evaluating on the test set, we train on the combination of train and tune and select the best epoch on the dev set.
On RCV1, we tune hyperparameters for the BERT BCE baseline (see \aref{sec:appendix_experiments_setup}).
For HiDEC-s HBM, we use \citet{kim-etal-2024-hierarchy}'s hyperparameters, and on MIMIC-III, we use  \citet{kim-etal-2022-anemic}'s hyperparameters. 
On both datasets, we tune the oversampling rate and $\gamma$ for ML-ROS and FL, respectively (see \trefplural{tab:rcv1_tuning_results_performance} and \ref{tab:mimic_dev_performance} in \aref{sec:appendix_experiments_setup}).
\blue{For an overview of model parameters, computational infrastructure and budget%
, see \aref{sec:appendix_experiments_setup}}.

\section{Experimental Results}
\label{sec:experiments}
In this section, we describe our experimental results for estimating calibration for multi-label classification using our proposed evaluation metric.

\begin{table*}[htb]
    \centering
    \footnotesize
    \setlength{\tabcolsep}{3pt}
\begin{tabular}{l|c|l|rrrr|rrr}
\toprule
& & & \multicolumn{4}{c|}{\textbf{Share (\%) of bins of size}} &\multicolumn{3}{c}{\textbf{Bin size statistics}} \\
\textbf{Dataset} & \textbf{Ideal bin size} & \textbf{Binning scheme} & <= 10 & <= 100 & <= 1000 & <10 000 & Average & Std. dev.  & Median\\
\midrule

\multirow{4}{*}{\vspace{0.4cm}MIMIC-III dev} & \multirow{4}{*}{\vspace{0.4cm}163.1} &  fixed-width & 40.4 & 43.9 & 44.0 & 100 & 914.0 & 805.7 & 1627.2\\[-0.15cm]
&  &  & \tiny {$\pm 0.8$} & \tiny {$\pm 0.7$} & \tiny {$\pm 0.7$} & \tiny {$\pm 0.0$} & \tiny {$\pm 11.7$} & \tiny {$\pm 1.4$} & \tiny {$\pm 1.3$}\\
 & & \ourScheme & \textbf{2.2} & \textbf{2.5} & \textbf{98.8} & 100 & \textbf{180.1} & \textbf{164.7} & \textbf{163.0}\\[-0.15cm]
&  &  & \tiny {$\pm 0.1$} & \tiny {$\pm 0.1$} & \tiny {$\pm 0.1$} & \tiny {$\pm 0.0$} & \tiny {$\pm 0.8$} & \tiny {$\pm 3.7$} & \tiny {$\pm 0.0$}\\[-0.1cm]

\midrule
\multirow{4}{*}{\vspace{0.4cm}RCV1 dev} & \multirow{4}{*}{\vspace{0.4cm}231.5} & fixed-width & 73.0 & 82.1 & 84.6 & 100 & 352.3 & 801.9 & 3.0\\[-0.15cm]
&  &  & \tiny {$\pm 0.5$} & \tiny {$\pm 0.4$} & \tiny {$\pm 0.5$} & \tiny {$\pm 0.0$} & \tiny {$\pm 11.0$} & \tiny {$\pm 11.0$} & \tiny {$\pm 0.0$}\\
& & \ourScheme & \textbf{22.2} & \textbf{42.2} & \textbf{94.3} & 100 & \textbf{321.4}  & \textbf{453.3} & \textbf{231.0}\\[-0.15cm]
&  &  & \tiny {$\pm 0.9$} & \tiny {$\pm 0.6$} & \tiny {$\pm 0.1$} & \tiny {$\pm 0.0$} & \tiny {$\pm 3.7$} & \tiny {$\pm 11.6$} & \tiny {$\pm 0.0$}\\[-0.1cm]

   \bottomrule
    \end{tabular}
    \vspace{-1em}
    \caption{\textbf{Distribution of bin sizes of fixed-width and \ourScheme binning schemes.} \blue{Ideally, bins should be of roughly the same size. The bins produced by \ourScheme are much closer to this ideal size than fixed-width, resulting in more reliable calibration error estimates. Results are shown for MIMIC-III using BiomedBERT-base and RCV1 using BERT-base, both trained with BCE loss and no oversampling, averaged over 5 random seeds. Empty bins, which fixed-width, but not \ourScheme can create, have been excluded from this analysis, since they do not contribute to the calibration error estimation. For test set results, see \trefplural{tab:bin_sizes_mimic_test} and \ref{tab:bin_sizes_rcv_test}.}}
    \label{tab:bin_sizes}
\end{table*}

\noindent \textbf{Binning Scheme Validation.}
We first apply our new binning scheme to validate it in comparison with fixed-width and adaptive binning.
\blue{
\fref{fig:ece_by_scheme} compares ECEs computed based on the adaptive, fixed-width, and \ourScheme binning schemes on RCV1 and MIMIC-III.}
\red{On both datasets, adaptive binning leads to ECE scores that strongly correlate with class frequency: the more frequent a class is, the less calibrated it seems according to adaptive binning, albeit models typically learn to perform better on more frequent classes. %
Adaptive binning results in low ECEs for non-frequent classes because for these labels, typically only a single bin contains positive predictions (see \fref{fig:binning_adaptive}).}

\ourSchemeCapitalized is designed to suffer
neither from the artifact that majorities of negative predictions make ECE diminish nor from unreliable bin sizes.
On RCV1, ECE using \ourScheme is not correlated to label frequency.
In contrast to RCV1, MIMIC has many labels with $10^1$ to $10^2$ training instances that are easy to predict or simply never predicted (see \fref{fig:performance_frequency_mimic_dev} in \aref{sec:appendix_experiments_results}), achieving near-zero calibration errors also under \ourScheme.%

\blue{Fixed-width binning behaves similar to \ourScheme in terms of correlation with label frequency, but comes with large shares of bins of very small size, resulting in less reliable error estimates (see \tref{tab:bin_sizes}).}

\begin{table*}[htb]
    \centering
    \footnotesize
    \setlength{\tabcolsep}{3.6pt}
\begin{tabular}{lll|rr|rrrr}
\toprule
model & loss & OS & $\uparrow$ micro\fscore & $\uparrow$ macro\fscore & $ \downarrow \text{ECE}_{\text{ML}}$ & $ \downarrow \text{ECE}_{\text{ML}}^{\text{high}}$ & $\downarrow \text{ECE}_{\text{ML}}^{\text{med}}$ & $\downarrow \text{ECE}_{\text{ML}}^{\text{rare}}$ \\
    \midrule
BERT & BCE & 0 & 86.2 & 67.8 & 7.8 & 5.0 & 8.4 & 8.3\\[-0.15cm]
& & & \tiny {$\pm 0.1$} & \tiny {$\pm 0.8$} & \tiny {$\pm 0.7$} & \tiny {$\pm 0.4$} & \tiny {$\pm 0.5$} & \tiny {$\pm 1.2$}\\
BERT & WBCEU & 0 & 85.9 & 68.7 & 9.0 & 5.5 & 9.8 & 9.6\\[-0.15cm]
& & & \tiny {$\pm 0.2$} & \tiny {$\pm 0.4$} & \tiny {$\pm 0.6$} & \tiny {$\pm 0.3$} & \tiny {$\pm 0.2$} & \tiny {$\pm 1.7$}\\
BERT & WBCEP & 0 & 84.0 & 67.2 & 14.8 & 6.3 & 13.8 & 20.4\\[-0.15cm]
& & & \tiny {$\pm 0.6$} & \tiny {$\pm 0.9$} & \tiny {$\pm 0.7$} & \tiny {$\pm 0.3$} & \tiny {$\pm 0.7$} & \tiny {$\pm 0.8$}\\
BERT & BCE & 10 & 86.2 & 68.7 & 9.4 & 5.1 & 9.2 & 11.8\\[-0.15cm]
& & & \tiny {$\pm 0.3$} & \tiny {$\pm 0.4$} & \tiny {$\pm 0.9$} & \tiny {$\pm 0.6$} & \tiny {$\pm 1.1$} & \tiny {$\pm 1.2$}\\
BERT & FL ($\gamma=2$) & 0 & \textbf{86.7} & 68.7 & \textbf{5.4} & \textbf{3.4} & \textbf{5.3} & \textbf{6.3}\\[-0.15cm]
& & & \tiny {$\pm 0.0$} & \tiny {$\pm 0.6$} & \tiny {$\pm 0.6$} & \tiny {$\pm 0.7$} & \tiny {$\pm 0.7$} & \tiny {$\pm 0.8$}\\

HiDEC-s & HBM & 0 & 86.4 & \textbf{68.8} & 9.4 & 5.2 & 9.3 & 11.5\\[-0.15cm]
& & & \tiny {$\pm 0.2$} & \tiny {$\pm 0.5$} & \tiny {$\pm 0.4$} & \tiny {$\pm 0.4$} & \tiny {$\pm 0.3$} & \tiny {$\pm 0.8$}\\

HiDEC-s & FL ($\gamma=1$) & 0 & 86.5 & 68.5 & 7.6 & 3.9 & 7.8 & 9.0\\[-0.15cm]
& & & \tiny {$\pm 0.1$} & \tiny {$\pm 0.4$} & \tiny {$\pm 0.6$} & \tiny {$\pm 0.6$} & \tiny {$\pm 0.4$} & \tiny {$\pm 1.0$}\\

    \bottomrule
    \end{tabular}
    \caption{\textbf{Performance and calibration on \blue{entire} RCV1 test set}. Frequent labels: \frequentlabels{} \rcvtestfrequentlabels{}
    Labels of medium frequency: \mediumlabels{}    \rcvtestmediumlabels{}
    Rare labels: \rarelabels{} \rcvtestrarelabels{}
    \experimentalsetting{}
    macro\fscore: hierarchical macro \fscore score.
    }
    \label{tab:rcv1_performance_and_calibration}
\end{table*}

\noindent \textbf{Performance and Calibration When Using Class Imbalance Methods.}
\blue{\tref{tab:rcv1_performance_and_calibration} reports results of standard class imbalance methods on RCV1. While oversampling and loss reweighting improve performance, they also worsen calibration. Focal loss improves both performance and calibration.
We observe similar patterns on MIMIC (see \tref{tab:mimic_performance_and_calibration} in \aref{sec:appendix_experiments}), with PLM-ICD showing a considerably higher performance than BERT, and better calibration on high- and medium-frequency labels.}
\noindent \textbf{Sensitivity of Rankings to Binning Hyperparameters.}
We analyze the sensitivity of \ourScheme to the minimum bin size $b_{min}$ and the number of bins $b$ by computing $ \text{ECE}_{\text{ML}} $ and its frequency-group variants from \tref{tab:rcv1_performance_and_calibration} for all potential combinations of setting $b_{min}$ and $b$ to one of three values (5, 10, 20), respectively.
Each of the nine combinations results in the same ranking of models as displayed in \tref{tab:rcv1_performance_and_calibration}, indicating the robustness of \ourScheme.

\begin{figure}[htb]
  \begin{subfigure}{\linewidth}
  \centering
  \includegraphics[width=\linewidth]{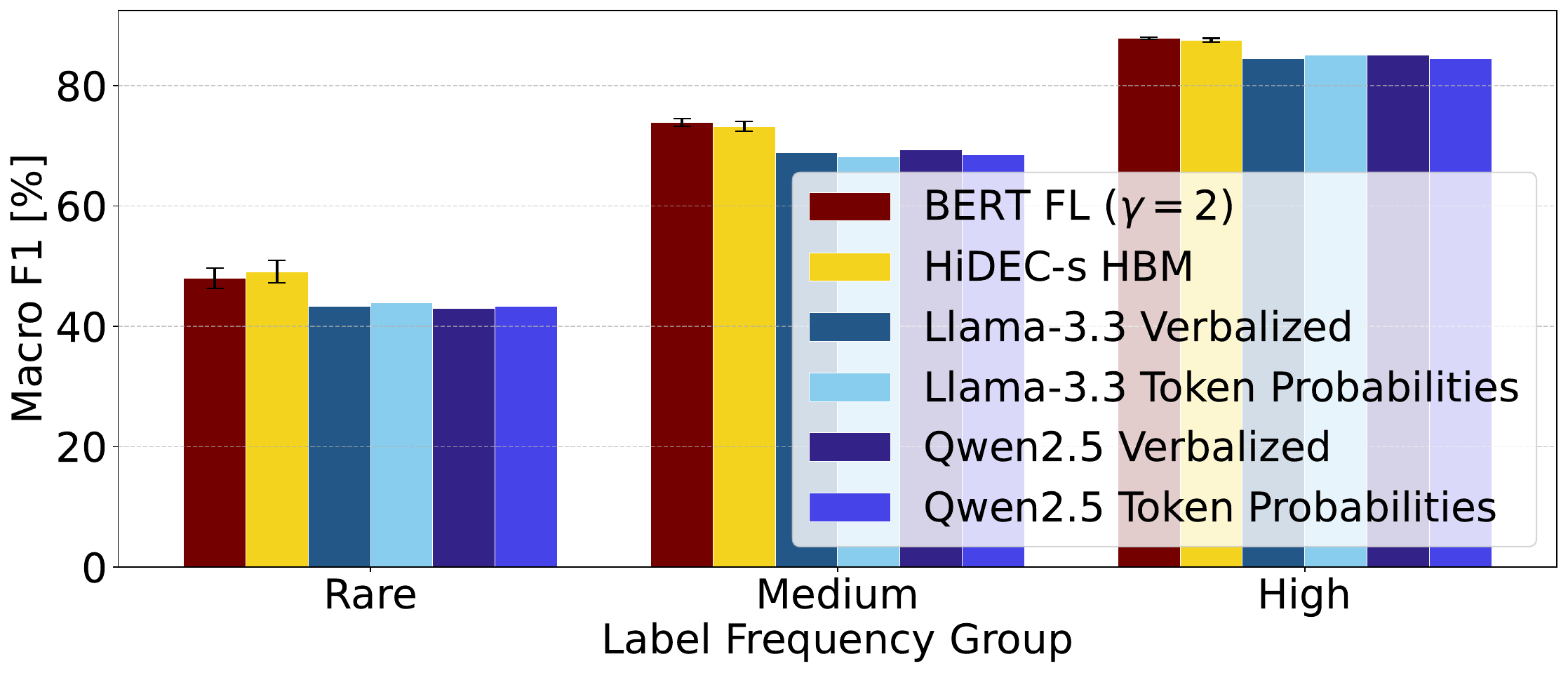}
    \caption{$\uparrow$ Macro \fscore}
  \label{fig:rcv1_subsampled_frequency_groups_performance}
    \end{subfigure}
\hfill
    \begin{subfigure}{\linewidth}
      \centering
      \includegraphics[width=\linewidth]{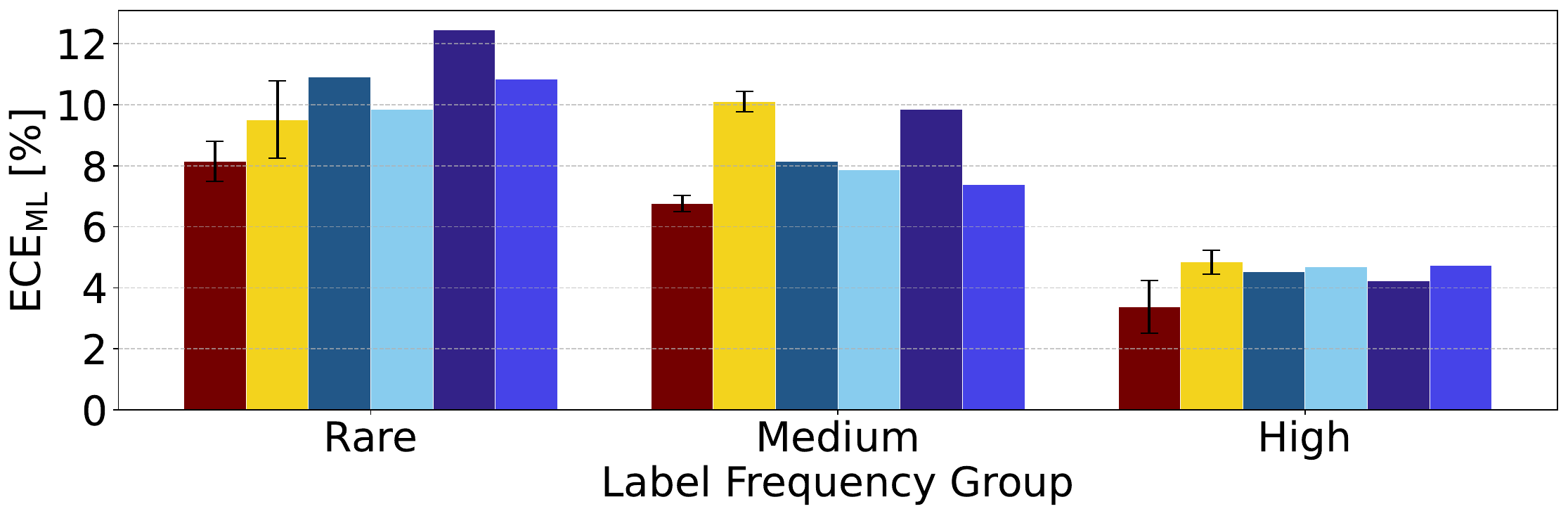}
      \caption{$\downarrow \text{ECE}_{\text{ML}}$}
      \label{fig:rcv1_subsampled_frequency_groups_ece}
    \end{subfigure}   
  \caption{\blue{\textbf{Performance and calibration of fine-tuned classifiers} (best model of its type, e.g., best BERT-based model) \textbf{compared to LLM  classifiers with RAG on subsampled RCV1 testset}.} For experimental details, see \tref{tab:rcv1_subsampled_performance_and_calibration}. Fine-tuned models outperform LLM-based approaches across all frequency groups. BERT-based models are better calibrated than the other models.}
  \label{fig:rcv1_subsampled_frequency_groups_performance_and_calibration}
\end{figure}

\noindent \textbf{Performance and Calibration of LLM-Based Classifiers.}
In a multi-label setting with many labels, retrieving instances with exactly the same labels assigned as the test instance usually cannot be expected.
Nevertheless, LLM-based classifiers heavily benefit from retrieved examples: their performance typically increases by around 20 points in macro \fscore compared to zero-shot prompting (see \tref{tab:rcv1_subsampled_performance_and_calibration} in \aref{sec:appendix_experiments_results}).
The observed benefit of seeing examples from the input distribution and the label space is in line with prior work showing that few-shot prompting can work even with wrong or random labels \citep{min-etal-2022-rethinking,yoo-etal-2022-ground,DBLP:journals/corr/abs-2303-03846}.

\fref{fig:rcv1_subsampled_frequency_groups_performance_and_calibration} provides a frequency-group analysis of the best LLM-based and fine-tuned multi-label classifiers.
Fine-tuned models consistently outperform LLM-based ones in both performance and calibration.
For example, BERT with focal loss achieves lower ECE on frequent and medium-frequency labels.
Fine-tuned models are also far more efficient, running up to 10,000 times faster than RAG-based methods (see \tref{tab:runtimes_rcv1} in \aref{sec:appendix_experiments_results}). %
Thus, when sufficient training data is available, fine-tuned classifiers are the superior choice. 
LLM-based classifiers, however, remain valuable in low-data settings, outperforming random and majority baselines.

\section{Conclusion}
\label{sec:conclusion}
In this work, we have addressed the limitations of existing calibration metrics when applied to multi-label text classification, particularly under severe class imbalance. 
We have introduced a new calibration metric and an adaptive binning scheme that better capture per-label miscalibration and provide meaningful estimates.
Our empirical study on large-scale, imbalanced benchmarks demonstrates that fine-tuned classifiers trained with focal loss offer the most reliable balance of performance and calibration, while LLM-based classifiers remain less well-calibrated despite showing promising results with RAG.
Overall, our contributions provide a foundation for more robust evaluation and development of calibrated models in multi-label text classification, highlighting both effective current strategies and open challenges for future research.

\section*{Limitations}
\citet{baan-etal-2022-stop} argue that ECE and its variants are not meaningful if humans inherently disagree on label assignments for a given single-label task, because even the oracle classifier which perfectly models the human disagreement distribution will be miscalibrated according to these metrics.
They support this claim with a case study on ChaosNLI \citep{nie-etal-2020-learn}, a natural language inference (NLI) dataset consisting of instances with weak human annotator agreement.
Their proposed alternative approach requires a reliable estimate of the human judgement distribution, which can be difficult to obtain in practice and is, to the best of our knowledge, not available for the datasets in our study.
However, we agree that when using $\text{ECE}_{\text{ML}}$ to assess calibration, it should always be kept in mind that it measures calibration with respect to the gold standard.

For obtaining confidence estimates from LLMs, we only compare the currently most common methods including verbalizations and logit-based scores.
Using our proposed methods, future work can also evaluate additional methods for retrieving confidence scores from LLMs \cite{fadeeva-etal-2023-lm}.

\section*{Acknowledgements}
We thank Amaan Ansari, Yaqi Zhang, and Mohamed H. Gad-Elrab for their contributions to preliminary experiments related to this paper.
We also thank Stefan Grünewald for detailed feedback on the paper manuscript and Fabian Kunze for helpful discussions on the paper.

This work was co-funded by the European Union (ERC, EPICAL, 101141712). Views and opinions expressed are however those of the author(s) only and do not necessarily reflect those of the European Union or the European Research Council. Neither the European Union nor the granting authority can be held responsible for them.
\clearpage

\bibliography{anthology_0,anthology_1,custom}

\appendix

\clearpage
\section{Appendix}
\label{sec:appendix}
\subsection{Calibration Metrics for Multi-Label Calibration}
\label{sec:appendix_multilabel_metrics}
\paragraph{Prediction Set Calibration.}
If multi-label classification is modeled as multi-class classification over label sets, a predicted set counts as correct if and only if it exactly matches the ground truth set of applying labels \citep{li2019multilabel}, thus not differentiating between almost correct solutions and those where the predicted label set notably deviates from the ground truth set.
If the prediction set probability is computed as the product of the per-label probabilities of assigning the label (if the specific label is contained in the prediction set) respectively the probability of not assigning it (if the label is not contained in the prediction set) \citep{li2019thesis}, %
the resulting prediction set probabilities become infinitesimally small with large label spaces.
For example, the MIMIC-III dataset consists of 8930 labels, and even an unrealistically good classifier that always assigns a probability of $0.99$ to the correct per-label decision would only assign a probability of $0.99^{8930} \approx 1.05 \times 10^{-39}$ to the ground-truth prediction set for a given instance.
Even the top-1 predictions will hence be associated with extremely low probabilities, making it hard to interpret them or even evaluate if they are calibrated.

\begin{figure*}[htb]
  \begin{center}
    \begin{subfigure}[t]{0.32\textwidth}
      \centering
      \includegraphics[width=\linewidth]{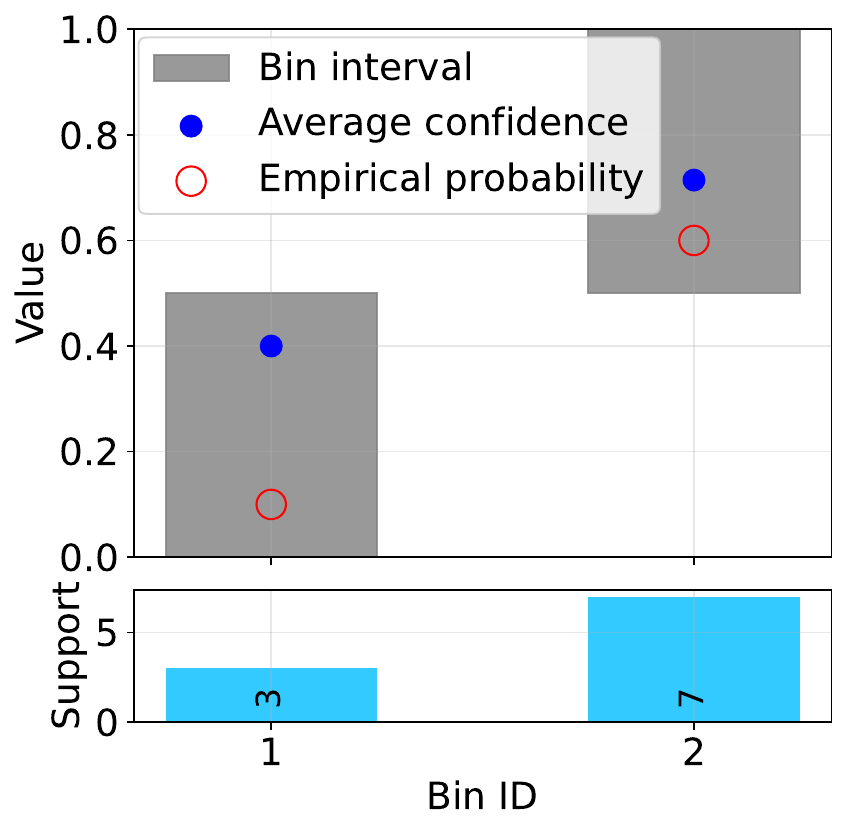}
      \caption{Overconfident label}
      \label{fig:binning_overconfident_label}
    \end{subfigure}
    \hfill
    \begin{subfigure}[t]{0.32\textwidth}
      \centering
      \includegraphics[width=\linewidth]{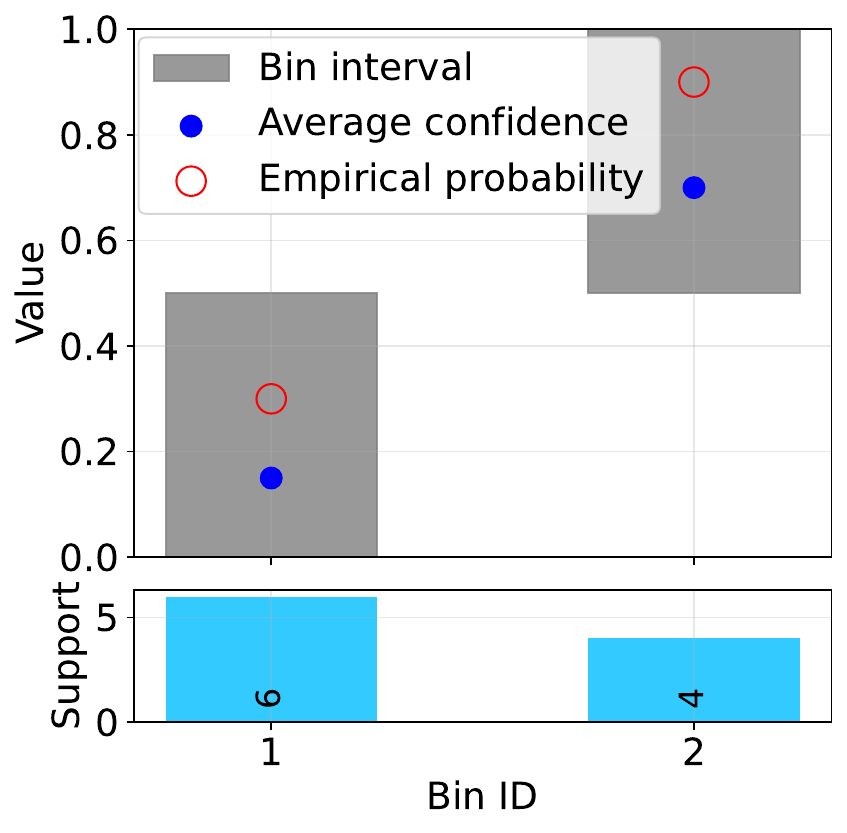}
      \caption{Underconfident label}
      \label{fig:binning_underconfident_label}
    \end{subfigure}
    \hfill
    \begin{subfigure}[t]{0.32\textwidth}
      \centering
      \includegraphics[width=\linewidth]{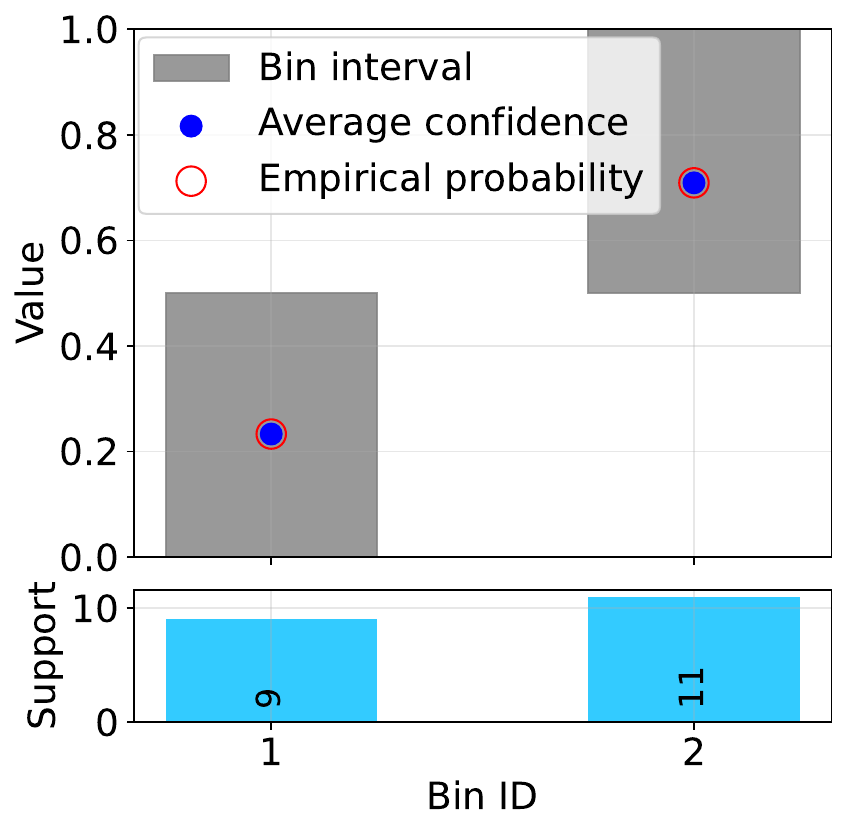}
      \caption{Joint binning}
      \label{fig:binning_over_and_underconfident_merged}
    \end{subfigure}
  \end{center}
  \caption{\textbf{Joint binning of all labels can conceal that the classifier is overconfident on some labels and underconfident on others.} Simple example (not based on real data) to illustrate the problem.}
  \label{fig:binning_overconfidence_concealing_underconfidence}
\end{figure*}

\begin{figure*}[htb]
  \begin{center}
    \begin{subfigure}[t]{0.32\textwidth}
      \centering
      \includegraphics[width=\linewidth]{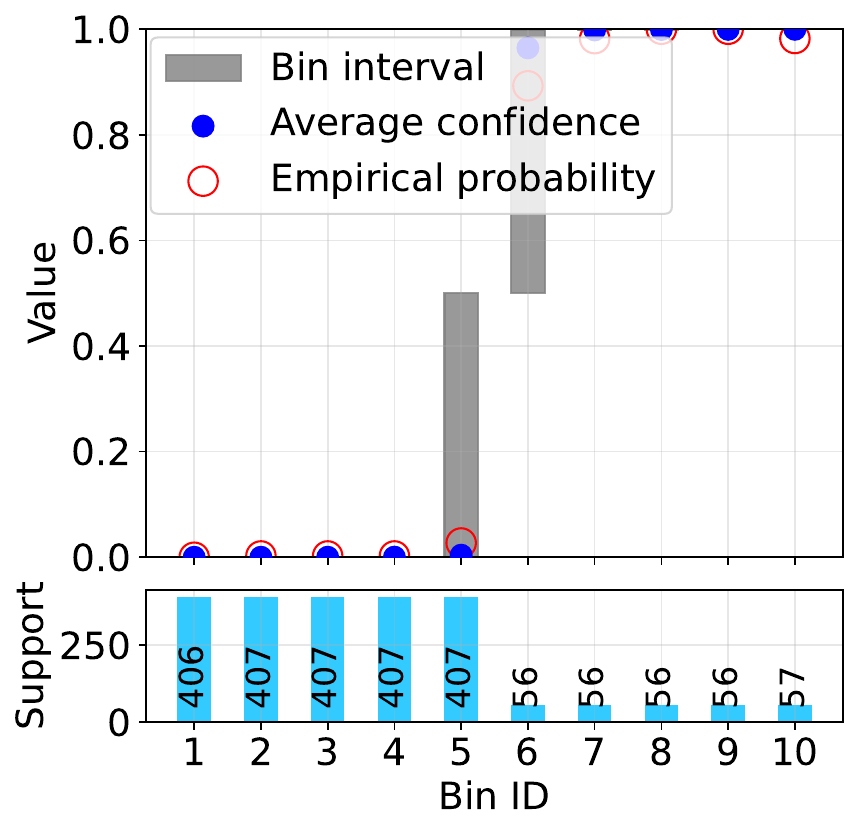}
      \caption{High-frequency label \enquote{M14}}
      \label{fig:binning_frequent_label}
    \end{subfigure}
    \hfill
    \begin{subfigure}[t]{0.32\textwidth}
      \centering
      \includegraphics[width=\linewidth]{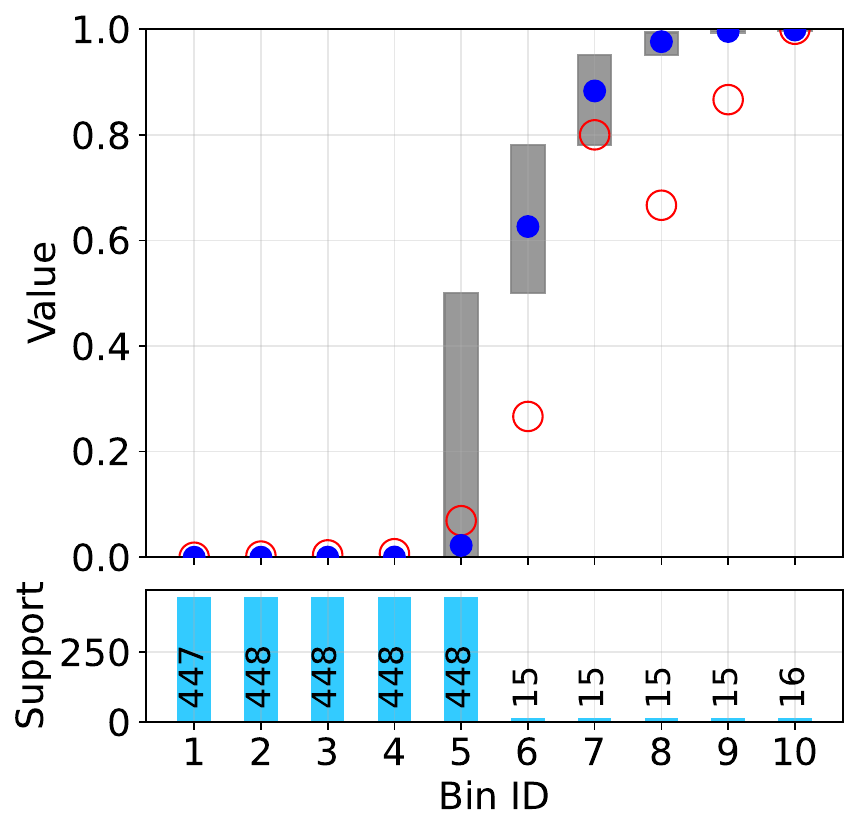}
      \caption{Medium-frequency label \enquote{C21}}
      \label{fig:binning_medium_label}
    \end{subfigure}
    \hfill
    \begin{subfigure}[t]{0.32\textwidth}
      \centering
      \includegraphics[width=\linewidth]{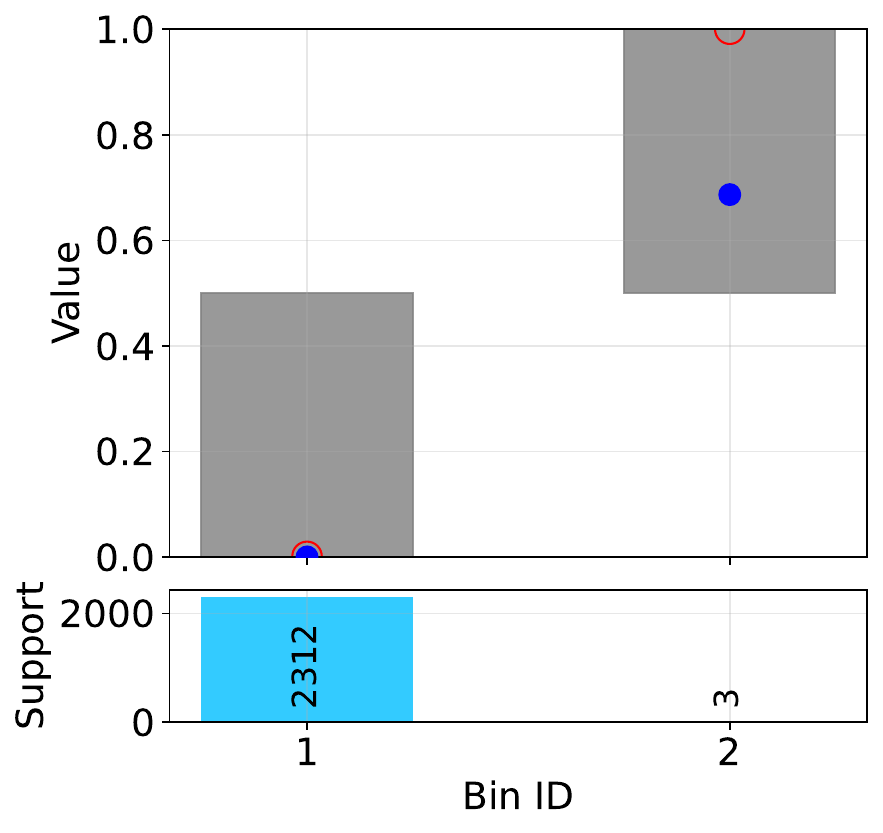}
      \caption{Rare label \enquote{C32}}
      \label{fig:binning_rare_label}
    \end{subfigure}
  \end{center}
  \caption{\textbf{\ourSchemeCapitalized binning for randomly chosen labels with different frequencies}, based on a single development run of BERT-base with BCE loss and no oversampling on RCV1. The method yields sufficient bin sizes for frequent and medium-frequency labels, but rare labels often have only one small bin of positive predictions. \# instances with the respective label in training/evaluation dataset: 2025/287 (M14), 628/92 (C21), 32/7 (C32).}
  \label{fig:binning_frequency_groups}
\end{figure*}

\paragraph{Design of \ourSchemeCapitalized.}
\blue{To robustly handle scenarios with significant prediction imbalance, our scheme is designed to consolidate predictions into single positive and negative bins when a label has fewer than $2 \cdot b_{min}$ positive predictions (as illustrated in \fref{fig:binning_frequency_groups}).
The resulting error is typically dominated by the calibration error on the positive predictions, since the negative predictions are usually well-calibrated.
}

\blue{Furthermore, we made a deliberate design choice to evaluate calibration on a per-label basis. This user-centric approach is motivated by the practical needs of end users, who are typically focused on the specific labels relevant to their text document(s)
rather than the entire label space (e.g., all possible medical codes).}

\blue{While we note recent work \citep {pmlr-v258-panda25a} that explores integrating label dependencies into their calibration error estimate using copulas, their approach is validated on a much less complex dataset with 20 labels and each label having more than 100 instances in the training and validation set, respectively.
Moreover, the reliable estimation of copulas in the very high-dimensional and sparse settings typical of multi-label text classification is an open research problem \citep{anatolyev2022copula}.
}

\subsection{Datasets}
\label{sec:appendix_datasets}
\begin{table}[htb]
    \centering
    \begin{tabular}{lrr}
    \toprule \textbf{per instance} & RCV1 & MIMIC-III \\
    \midrule
    \# sentences & 9.95 & 120.69 \\ [-0.15cm]
    & \footnotesize{$\pm9.72$} & \footnotesize{$\pm62.56$} \\
    \# tokens & 282.83 & 2462.56 \\ [-0.15cm]
    & \footnotesize{$\pm296.43$} & \footnotesize{$\pm1261.15$} \\
    \# labels & 3.24 & 15.88 \\[-0.15cm]
    & \footnotesize{$\pm1.40$} & \footnotesize{$\pm8.14$} \\

    \bottomrule
    \end{tabular}
    \caption{Dataset Statistics.%
    }
    \label{tab:datasets}
\end{table}

\begin{table}[htb]
\centering
\footnotesize
\setlength{\tabcolsep}{2pt}
\begin{tabular}{lrrrrr}
\toprule
\textbf{Split} & \textbf{Train} & \textbf{Tune} & \textbf{Dev} & \textbf{Test} & \textbf{Subs.} \\
& & & & & \textbf{Test}\\
\midrule
\# Instances & 18,519 & 2,315 & 2,315 & 781,261 & 5,000 \\
Avg. \# Labels & 3.19 & 3.14 & 3.17 & 3.24 & 3.55 \\
Avg. \# Tokens & 254.87 & 256.20 & 263.87 & 265.15 & 269.98 \\
Avg. \# Sent. & 9.99 & 10.17 & 10.44 & 10.04 & 10.25 \\
\# Unique Labels & 101 & 99 & 96 & 103 & 100\\
\bottomrule
\end{tabular}
\caption{Statistics of each split of the \textsc{RCV1} dataset. \textbf{Subs. Test: }Subsampled version of the test set used in LLM experiments.}
\label{tab:rcv1-split-stats}
\end{table}

\begin{table}[htb]
\centering
\footnotesize
\setlength{\tabcolsep}{4pt}
\begin{tabular}{lrrrrr}
\toprule
\textbf{Split} & \textbf{Train} & \textbf{Tune} & \textbf{Dev} & \textbf{Test} \\
\midrule
\# Instances & 45,336 & 2,387 & 1,631 & 3,372 \\
Avg. \# Labels & 15.69 & 15.43 & 17.41 & 17.99 \\
Avg. \# Tokens & 2261.19 & 2260.85 & 2745.02 & 2763.11 \\
Avg. \# Sent. & 123.71 & 122.33 & 133.72 & 133.79 \\
\# Unique Labels & 8593 & 3461 & 3012 & 4085 \\
\bottomrule
\end{tabular}
\caption{Statistics of each split of the \textsc{MIMIC-III} dataset.}
\label{tab:mimic-split-stats}
\end{table}

\paragraph{Dataset Statistics.}
\tref{tab:datasets} shows general statistics for RCV1 and MIMIC-III.
\trefplural{tab:rcv1-split-stats} and \ref{tab:mimic-split-stats} show statistics for the subsets of RCV1 and MIMIC-III, respectively, that we use in our experiments as described in \sref{sec:experiments}. 
Dataset statistics were computed based on sentencization and tokenization performed with spaCy's \texttt{en\_core\_web\_sm-3.7.1} model \citep{Honnibal_spaCy_Industrial-strength_Natural_2020}.

\paragraph{Details on Subsampled RCV1 Test Dataset.}
For testing in runtime-intensive scenarios, we create a subsampled version of the RCV1 testset containing 5,000 instances.
Drawing a representative subsample is nontrivial in the multi-label setting, as each instance can be associated with multiple labels. 
Our goal was to preserve the original label distribution as closely as possible.
To achieve this, we designed a sampling algorithm that iteratively selects instances in a way that minimizes the difference in label frequencies between the subsample and the original test set. 
Specifically, during each sampling iteration, we identify the label with the highest relative deficit (i.e., the largest underrepresentation compared to the full distribution) and randomly draw an instance that contains this label. 
This process is repeated until the desired number of instances is reached. We ran the sampling procedure with 11 different random seeds and selected the subsample with the lowest Jensen–Shannon (JS) divergence from the original label distribution. 
The final subsample has a JS divergence of 0.0344 from the original test set.

\subsection{Models}
\label{sec:appendix_modeling}
\subsubsection{HiDEC and HBM Loss}
\label{sec:appendix_modeling_hidec}
Unlike other recent HTC methods that build hierarchy directly into their model structure, \citet{Im_Kim_Oh_Jo_Kim_2023} treat hierarchical text classification as a sequence generation problem. Their Hierarchy DECoder (HiDEC) model uses an encoder-decoder setup to generate a sequence of labels representing a path through the label hierarchy.
HiDEC learns the relationships between labels along the hierarchy path, from the root to each leaf, using an attention mechanism combined with a special masking technique that focuses on parent-child label dependencies. This helps the model understand the hierarchical structure during training. The model then combines this hierarchical information with the text features to make better predictions.
During inference, HiDEC generates labels in a top-down manner, starting from parent labels and moving down to child labels, using a recursive decoding process that respects the hierarchy.

Hierarchy-aware Biased Bound Margin (HBM) Loss \citep{kim-etal-2024-hierarchy} is a hierarchy-aware loss function specifically designed for unit-based HTC models like HiDEC, addressing two central challenges: label imbalance and static thresholding. HBM Loss integrates learnable bounds, biases, and a margin. The biases and margin mitigate the label imbalance by promoting low-confidence labels and excluding high-confidence labels from a loss, respectively. The learnable bounds address the static thresholding problem. These bounds are optimized for all units within a hierarchy during training and serve as dynamic unit thresholds during inference. 
In the context of the HiDEC model, a unit refers to the set of child labels associated with a given parent label in the label hierarchy. That is, when using HBM Loss with the HiDEC model, during hierarchical decoding, the model expands from parent to child labels in a top-down manner, applying a distinct threshold at each step of the sub-hierarchy expansion.

To enable comparison with confidence values from other models optimized for the standard 0.5 threshold, we rescale the thresholds and probabilities of each child label so that the threshold aligns with a fixed value of 0.5 and the probabilities maintain their relative position to the original threshold. 
We calculate the scaled probabilities using the original probabilities $p$ and thresholds $t$. 

\begin{equation}
p_\text{scaled} = 
\begin{cases}
0.5 + 0.5 \cdot \dfrac{p-t}{1 - t} & \text{if } p \geq t \\
0.5 \cdot \dfrac{p}{t} & \text{if } p < t
\end{cases}
\label{eq:hidec-scaling}
\end{equation}

By using this piecewise linear scaling, the probability values above or equal to the original threshold are scaled into the range [0.5, 1], and values below the original threshold into [0, 0.5), preserving their relative distance from the original threshold and ensuring a fair and consistent evaluation under metrics that assume this standard threshold. 

\subsubsection{Class Imbalance Methods}
\label{sec:appendix_modeling_imbalance}
For the weighted BCE variants, we use the following formulae:
\begin{itemize}
    \item $w^{\text{neg}} = 1$ for all variants
    \item \textbf{WBCEU}: Uniformly up-weighting positive instances \citep{rathnayaka2019gated} with $w^{\text{pos}}=2$
    \item \textbf{WBCEM}: Up-weighting positive instances of rare classes (proportional to class frequency):
    \[
    w^{\text{pos}}_i =\max{\left(\frac{N}{2*n_i},1\right)}
    \]
    where $N$ is the number of instances in the data set and $n_i$ is the number of instances of class $i$
    \item \textbf{WBCEP}: Up-weighting positive instances of all classes with class frequency considered:
    \[
    w^{\text{pos}}_i = \frac{N}{2*n_i} + 1
    \]
\end{itemize}

For a single instance, FL is computed as follows:
$-\sum_{j=1}^C [y_j (1-p_j)^\beta \log p_j + (1-y_j) p_j^\beta \log(1-p_j)]$

\subsubsection{Multi-Label Classification with LLMs}
\label{sec:appendix_modeling_llms}
\begin{figure*}[htb]
    \centering
    \includegraphics[width=1.0\linewidth]{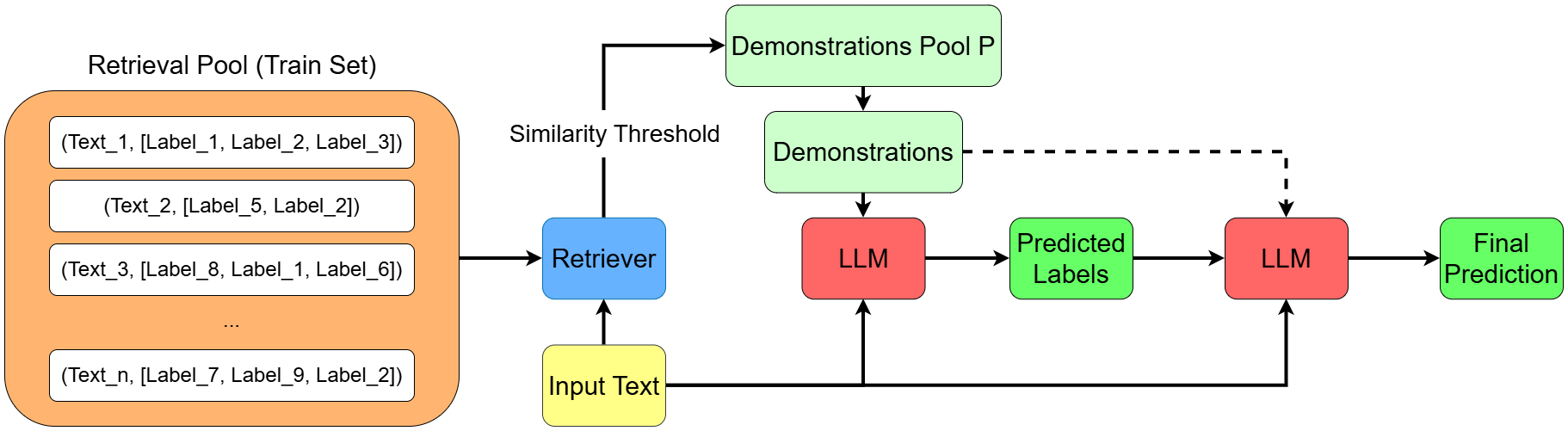}
    \caption{\textbf{LLM-based multi-label text classification system} with dense demonstrations retrieval component and confidence verbalizer. First, we prompt the LLM as shown in \frefplural{fig:prediction-prompt-RCV1} to predict the labels (first red box), which yields the predicted label set. 
    For \textbf{Token Prob.} confidence estimation, we extract the probability of the first generated token corresponding to each of these labels and use it to determine the confidence scores, which are then used as the final prediction.
    For \textbf{Verbalized} confidence estimation, we prompt the LLM to estimate its confidence for each predicted label on a scale from 0 to 1, as shown in \ref{fig:confidence-prompt-RCV1} (second red box). These confidence scores are then used as the final prediction.}
   \label{fig:llm-system}
\end{figure*}

Here, we detail how we perform multi-label classification using LLMs. 
\fref{fig:llm-system} gives an overview of how LLMs are prompted using few-shot demonstrations retrieved from the training set.

\paragraph{Model Selection.}
We use instruction-tuned variants of both the Llama 3.3 \citep{DBLP:journals/corr/abs-2407-21783} and Qwen 2.5 \citep{DBLP:journals/corr/abs-2412-15115} models. These models are available in various sizes; specifically, we employ Llama-3.3-70B-Instruct\footnote{\url{https://huggingface.co/meta-llama/Llama-3.3-70B-Instruct}} and Qwen2.5-72B-Instruct\footnote{\url{https://huggingface.co/Qwen/Qwen2.5-72B-Instruct}}. The Llama-3.3-70B-Instruct model has 70 billion parameters and supports a context length of up to 128,000 tokens. The Qwen2.5-72B-Instruct model consists of 72.7 billion parameters and supports a full context length of 131,072 tokens. For simplicity, we omit the \enquote{-3.3-70B-Instruct} and \enquote{2.5-72B-Instruct} suffixes in the model names throughout the following text.
These large models were chosen because their size and instruction tuning make them particularly capable of following complex prompts. Their long context length allows us to include a large number of in-context examples (few-shot demonstrations) directly within a single prompt, which is important for our underlying tasks of extreme multi-label text classification and sequence labeling. 

\begin{figure}[t]
    \centering
    \begin{tcolorbox}[colback=gray!5, colframe=black!50, width=0.45\textwidth, title=Prompt Format for Label Prediction]
    \footnotesize
\textbf{System:}\\
You are a helpful assistant for multi-label news topic classification.\\
Given a news article, predict the relevant topic labels.\\
Return the answer strictly in this format: ["label1", "label2", ..., "labeln"].\\
Your answer must start with '[' and end with ']'. Each label must be enclosed in double quotes, e.g., "label1". Labels must be comma-separated with no trailing comma.\\[1ex]
\textbf{User:}\\
Text: \textless text of first demonstration\textgreater\\
What topic labels apply?\\[1ex]
\textbf{Assistant:}\\
\textless labels of first demonstration\textgreater\\[1ex]
\texttt{... \textless more few-shot demonstrations\textgreater ...}\\[1ex]
\textbf{User:}\\
Text: \textless text of the current test instance\textgreater\\
What topic labels apply?\\[1ex]
\textbf{Assistant:}\\
\textless predicted labels\textgreater
    \end{tcolorbox}
    \caption{\textbf{Prompt format used for label prediction on RCV1 dataset} including instructions, few-shot demonstrations, and text of the current test instance.}
    \label{fig:prediction-prompt-RCV1}
\end{figure}

\begin{figure}[ht]
    \centering
    \begin{tcolorbox}[colback=gray!5, colframe=black!50, width=0.45\textwidth, title=Prompt Format for verbalized Confidence Prediction]
    \small
\textbf{System:}\\
You are a helpful assistant for multi-label news topic classification.\\
Given a news article, predict the relevant topic labels.\\
Return the answer strictly in this format: ["label1", "label2", ..., "labeln"].\\
Your answer must start with '[' and end with ']'. Each label must be enclosed in double quotes, e.g., "label1". Labels must be comma-separated with no trailing comma.\\[1ex]
\textbf{User:}\\
Text: \textless text of first demonstration\textgreater\\
What topic labels apply?\\[1ex]
\textbf{Assistant:}\\
\textless labels of first demonstration\textgreater\\[1ex]
\texttt{... \textless more few-shot demonstrations\textgreater ...}\\[1ex]
\textbf{User:}\\
Text: \textless text of the current test instance\textgreater\\
What topic labels apply?\\[1ex]
\textbf{Assistant:}\\
\textless predicted labels\textgreater\\[1ex]
\textbf{User:}\\
How confident are you in the label \textless one of the predicted labels\textgreater? Please provide a confidence score between 0 and 1. Only reply with the number, with no other text.\\[1ex]
\textbf{Assistant:}\\
\textless verbalized confidence prediction for specific label\textgreater
    \end{tcolorbox}
    \caption{\textbf{Prompt format used for verbalized confidence prediction on RCV1 dataset} including instructions, few-shot demonstrations used for prior label prediction, the text of the current test instance and the prior predicted labels. This prompt is used separately for each of the prior predicted labels.}
    \label{fig:confidence-prompt-RCV1}
\end{figure}
We prompt the LLMs as shown in \frefplural{fig:prediction-prompt-RCV1} to predict the labels given the selected demonstrations, their corresponding labels, and the text of the current test instance for which the labels are to be predicted. For \textbf{Token Prob.} confidence estimation, we extract the probability of the first generated token corresponding to each label and use it to determine the model's confidence.
For \textbf{Verbalized} confidence estimation, we construct a prompt for each predicted label that includes the selected demonstrations, their corresponding labels, the text of the current test instance, and the previously predicted labels. The LLM is then instructed to estimate its confidence in the predicted label on a scale from 0 to 1, as shown in \fref{fig:confidence-prompt-RCV1}. This prompt is applied separately to each predicted label contained in the label space. Invalid predicted labels (i.e., labels not contained in the label space) are discarded.

\paragraph{Retriever Tuning.}
We tune the retriever on the development set to achieve high recall while maintaining non-trivial precision. This balance is crucial in our use case because we want to provide the LLM with as many relevant candidate labels as possible, ensuring it has sufficient information to make accurate predictions. At the same time, we need to avoid overwhelming the model with too many irrelevant labels, which could reduce overall performance and increase computational cost. This process involves adjusting two key hyperparameters: the similarity threshold $t$ and $top\text{-}k$. The similarity threshold defines the minimum required similarity between a document in the retrieval pool and the input text during inference. Only documents that exceed this threshold are selected into the demonstrations pool $P$ and are considered for the in-context demonstrations. The $top\text{-}k$ parameter determines how many of the most similar documents from the demonstrations pool $P$ are selected to serve as in-context demonstrations.
During this phase, the labels from the retrieved demonstrations are used directly as predictions. 
The retriever is individually tuned for each dataset to optimize performance, resulting in $t = 0.25$ and $top\text{-}k = 30$ for RCV1 and in $t = 0.3$ and $top\text{-}k = 10$ for MIMIC-III.
On RCV1, we get a hierarchical macro recall of 87.8\%, a hierarchical macro precision of 17.4\%, and a hierarchical macro \fscore of 26.9\%.
On MIMIC-III, the respective configuration yields a macro recall of 8.3\%, a macro precision of 1.9\%, and a macro \fscore of 2.9\%, which is substantially lower
than the results obtained on RCV1. The lower performance is primarily due to the retriever struggling with rare labels. 

\subsection{Experiments}
\label{sec:appendix_experiments}
\subsubsection{Setup}
\label{sec:appendix_experiments_setup}
\paragraph{RCV1.}
We tune hyperparameters for BERT by selecting the best model epoch according to tune set performance and choosing the best hyperparameters based on dev set performance (according to macro \fscore). 
We tune batch size, learning rate, and dropout in a grid search, first exploring all combinations on our default random split of the training set and then evaluating the best 5 hyperparameter combinations on four more random splits of the training set, taking the one with best average performance as the final hyperparameter combination (batch size: 16, learning rate: $3e^{-5}$, dropout: 0.25).
We perform hyperparameter tuning once for the baseline model, i.e., BCE loss with no oversampling, and use the same hyperparameters for the other models evaluated.
For HiDEC, we use \citet{kim-etal-2024-hierarchy}'s hyperparameters for HBM. 
For ML-ROS and FL, we tune the oversampling rate and $\gamma$, respectively (\tref{tab:rcv1_tuning_results_performance}).
As an optimizer, we use Adam with default values.
When evaluating on the test set, we train on the combination of train and tune and select the best model epoch according to dev set performance.
\paragraph{MIMIC-III.}
On this dataset, we use  \citet{kim-etal-2022-anemic}'s hyperparameters. For ML-ROS and FL, we tune the oversampling rate and $\gamma$, respectively (\tref{tab:mimic_dev_performance}). 
\paragraph{Calibration on the Dev Set.} For calibration errors on the dev sets, see \trefplural{tab:rcv_dev_performance_and_calibration} and \ref{tab:mimic_dev_performance_and_calibration}.
\blue{\paragraph{Total Computational Effort of Experiments.}
Here, we estimate the total computational effort for the experiments reported in our paper. 
The numbers reported for HiDEC and the LLM-based models refer to 4 NVIDIA H100 GPUs.
Inference with HiDEC and the LLM-based models on RCV1 dev, subsampled test, and full test took roughly 153 hours.
Training all variants of HiDEC in the development and evaluation configuration took at most 125 hours (assuming the worst case training time of 2.5 hours with training for 100 epochs w/o early stopping for each variant).
For the BERT-style models, development runs took 18.6 hours on a single NVIDIA V100 GPU and 63.6 ours on a single NVIDIA A100 GPU, and evaluation runs took roughly 200 hours on a single NVIDIA A100 GPU.
For the models reported on MIMIC-III, development runs took 50.3 hours on a single NVIDIA V100 GPU and 1016.6 hours on a single NVIDIA A100 GPU, and evaluation runs took roughly 119 hours on a single NVIDIA A100 GPU.
A single run with the PLM-ICD models took up to 3 days on the A100, whereas a single run with the BERT-style models took roughly 16 hours.}

\begin{table}[htb]
    \centering
\begin{tabular}{l|r}
\toprule
model & \# parameters\\
    \midrule
BERT & 116M\\
 
HiDEC-s & 128M \\

Llama-3.3 & 70B \\
Qwen2.5 & 72B \\

ModernBERT & 149M\\
BiomedBERT & 116M \\
PLM-ICD & 124M\\

    \bottomrule
    \end{tabular}
    \caption{\textbf{Sizes of models used in experiments}. M: million, B: billion. BERT: bert-based-uncased using CLS token for classification. HiDEC-s: HiDEC implementation by \citet{kim-etal-2024-hierarchy}, using bert-base-uncased as the encoder, with scaled confidences to match 0.5 threshold (not affecting predicted labels). ModernBERT: ModernBERT-base, BiomedBERT: BiomedBERT-base-uncased-abstract, each using the CLS token for classification. PLM-ICD: PLM-ICD \citep{huang-etal-2022-plm} with BiomedBERT-base-uncased-abstract.     
    }
    \label{tab:model_sizes}
\end{table}

\begin{table*}[ht]
    \centering
    \footnotesize
\begin{tabular}{lll|rrr|rrr|r}
\toprule
model & loss & OS (\%) & microP & microR & micro\fscore   & macroP  & macroR & macro\fscore & \exactmatch\\
    \midrule
BERT & BCE & 0 & 87.2 & 88.9 & 88.0 & 69.5 & 69.6 & 68.1 & 66.5\\[-0.15cm]
&  &  & \tiny {$\pm 0.8$} & \tiny {$\pm 0.7$} & \tiny {$\pm 0.1$} & \tiny {$\pm 1.5$} & \tiny {$\pm 1.1$} & \tiny {$\pm 0.7$} & \tiny {$\pm 0.5$}\\
 
BERT & WBCEU & 0 & 85.9 & 89.5 & 87.7 & 68.2 & 70.1 & 67.8 & 66.0\\[-0.15cm]
&  &  & \tiny {$\pm 0.7$} & \tiny {$\pm 0.1$} & \tiny {$\pm 0.3$} & \tiny {$\pm 1.7$} & \tiny {$\pm 1.0$} & \tiny {$\pm 0.9$} & \tiny {$\pm 0.7$}\\

BERT & WBCEM & 0 & 79.5 & 90.9 & 84.7 & 60.1 & 77.5 & 65.5 & 61.3\\[-0.15cm]
&  &  & \tiny {$\pm 6.4$} & \tiny {$\pm 0.6$} & \tiny {$\pm 3.6$} & \tiny {$\pm 4.4$} & \tiny {$\pm 1.2$} & \tiny {$\pm 3.3$} & \tiny {$\pm 5.1$}\\

BERT & WBCEP & 0 & 82.4 & 91.2 & 86.6 & 62.6 & 77.1 & 67.7 & 63.4\\[-0.15cm]
&  &  & \tiny {$\pm 0.9$} & \tiny {$\pm 0.5$} & \tiny {$\pm 0.3$} & \tiny {$\pm 1.2$} & \tiny {$\pm 0.9$} & \tiny {$\pm 0.8$} & \tiny {$\pm 0.9$}\\

BERT & BCE & 10 & 86.8 & 89.3 & 88.0 & 69.5 & 70.1 & 68.3 & 66.3\\[-0.15cm]
&  &  & \tiny {$\pm 0.4$} & \tiny {$\pm 0.6$} & \tiny {$\pm 0.2$} & \tiny {$\pm 1.0$} & \tiny {$\pm 1.1$} & \tiny {$\pm 0.8$} & \tiny {$\pm 0.7$}\\

BERT & BCE & 20 & 87.3 & 88.4 & 87.8 & 69.8 & 69.1 & 68.0 & 65.6\\[-0.15cm]
&  &  & \tiny {$\pm 0.5$} & \tiny {$\pm 0.9$} & \tiny {$\pm 0.2$} & \tiny {$\pm 0.9$} & \tiny {$\pm 2.5$} & \tiny {$\pm 1.3$} & \tiny {$\pm 0.9$}\\
 
BERT & BCE & 30 & 87.1 & 88.5 & 87.8 & 68.9 & 69.2 & 67.6 & 65.8\\[-0.15cm]
&  &  & \tiny {$\pm 0.7$} & \tiny {$\pm 0.5$} & \tiny {$\pm 0.1$} & \tiny {$\pm 0.7$} & \tiny {$\pm 1.4$} & \tiny {$\pm 0.7$} & \tiny {$\pm 0.4$}\\

BERT & FL ($\gamma=1$) & 0 & 86.6 & 89.2 & 87.9 & 68.8 & 69.4 & 67.8 & 66.0\\[-0.15cm]
&  &  & \tiny {$\pm 0.5$} & \tiny {$\pm 0.4$} & \tiny {$\pm 0.3$} & \tiny {$\pm 1.4$} & \tiny {$\pm 0.8$} & \tiny {$\pm 0.8$} & \tiny {$\pm 0.8$}\\

BERT & FL ($\gamma=2$) & 0 & 87.7 & 88.9 & 88.3 & 71.4 & 69.2 & 68.7 & 66.8\\[-0.15cm]
&  &  & \tiny {$\pm 1.1$} & \tiny {$\pm 0.7$} & \tiny {$\pm 0.2$} & \tiny {$\pm 1.4$} & \tiny {$\pm 1.6$} & \tiny {$\pm 1.2$} & \tiny {$\pm 0.2$}\\

BERT & FL ($\gamma=3$) & 0 & 87.4 & 89.1 & 88.2 & 70.7 & 69.6 & 68.5 & 66.8\\[-0.15cm]
&  &  & \tiny {$\pm 1.1$} & \tiny {$\pm 0.6$} & \tiny {$\pm 0.4$} & \tiny {$\pm 1.4$} & \tiny {$\pm 0.8$} & \tiny {$\pm 0.8$} & \tiny {$\pm 0.9$}\\

HiDEC-s & HBM & 0 & 88.0 & 88.9 & 88.4 & 71.3 & 71.0 & 69.8 & 68.3\\[-0.15cm]
& & & \tiny {$\pm 0.3$} & \tiny {$\pm 0.7$} & \tiny {$\pm 0.4$} & \tiny {$\pm 1.2$} & \tiny {$\pm 1.6$} & \tiny {$\pm 1.0$} & \tiny {$\pm 0.5$}\\

$\text{HiDEC-s}^{\dagger} $ & HBM & 0 & 88.0 & 89.0 & 88.5 & 71.6 & 71.3 & 70.0 & 68.4\\[-0.15cm]
& & & \tiny {$\pm 0.5$} & \tiny {$\pm 0.4$} & \tiny {$\pm 0.1$} & \tiny {$\pm 1.5$} & \tiny {$\pm 1.4$} & \tiny {$\pm 1.1$} & \tiny {$\pm 0.4$}\\

$\text{HiDEC-s}^{\dagger \dagger}$ & HBM  & 0 & 88.0 & 88.9 & 88.5 & 71.1 & 70.9 & 69.7 & 68.5\\[-0.15cm]
& & & \tiny {$\pm 0.7$} & \tiny {$\pm 0.5$} & \tiny {$\pm 0.3$} & \tiny {$\pm 1.1$} & \tiny {$\pm 1.0$} & \tiny {$\pm 0.7$} & \tiny {$\pm 0.6$}\\

HiDEC-s & FL ($\gamma=1$) & 0 & 88.0 & 88.9 & 88.4 & 71.4 & 70.0 & 69.5 & 68.4\\[-0.15cm]
&  &  & \tiny {$\pm 0.5$} & \tiny {$\pm 0.4$} & \tiny {$\pm 0.2$} & \tiny {$\pm 0.5$} & \tiny {$\pm 0.9$} & \tiny {$\pm 0.8$} & \tiny {$\pm 0.6$}\\

HiDEC-s  & FL ($\gamma=2$) & 0  & 88.1 & 88.7 & 88.4 & 70.7 & 69.5 & 68.7 & 68.3\\[-0.15cm]
&  &  & \tiny {$\pm 0.5$} & \tiny {$\pm 0.6$} & \tiny {$\pm 0.2$} & \tiny {$\pm 1.2$} & \tiny {$\pm 1.3$} & \tiny {$\pm 1.3$} & \tiny {$\pm 0.6$}\\

HiDEC-s  & FL ($\gamma=3$) & 0 & 88.2 & 88.4 & 88.3 & 70.0 & 69.3 & 68.3 & 67.9\\[-0.15cm]
&  &  & \tiny {$\pm 0.5$} & \tiny {$\pm 0.5$} & \tiny {$\pm 0.3$} & \tiny {$\pm 1.2$} & \tiny {$\pm 1.0$} & \tiny {$\pm 0.7$} & \tiny {$\pm 0.9$}\\

\midrule
model & prompting & confidence & microP & microR & micro\fscore   & macroP  & macroR & macro\fscore & \exactmatch\\
\midrule

Llama-3.3 & zero-shot & Verbalized &  54.1 & 74.4 & 62.7 & 40.7 & 62.0 & 43.1 & 2.3\\
Llama-3.3 & zero-shot & Token Prob. &  52.3 & 72.5 & 60.8 & 39.6 & 60.9 & 41.6 & 2.7\\
Llama-3.3 & RAG & Verbalized & 88.6 & 84.4 & 86.5 & 74.3 & 68.8 & 68.8 & 56.1\\
Llama-3.3 & RAG & Token Prob. &  88.7 & 84.7 & 86.6 & 74.2 & 68.0 & 68.4 & 62.9\\

\hline
Qwen2.5 & zero-shot & Verbalized &  43.3 & 72.7 & 54.3 & 29.8 & 62.4 & 35.3 & 1.2\\
Qwen2.5 & zero-shot & Token Prob. &  48.8 & 66.9 & 56.5 & 31.2 & 54.1 & 35.0 & 1.1\\
Qwen2.5 & RAG & Verbalized &  87.8 & 86.5 & 87.2 & 72.8 & 70.7 & 69.9 & 64.7\\
Qwen2.5 & RAG & Token Prob. &  88.7 & 85.5 & 87.1 & 73.9 & 69.6 & 69.7 & 64.3\\

    \bottomrule
    \end{tabular}
    \caption{\textbf{Performance on RCV1 dev set}. BERT: bert-based-uncased using CLS token for classification. HiDEC-s: HiDEC implementation by \citet{kim-etal-2024-hierarchy}, using bert-base-uncased as the encoder, with scaled confidences to match 0.5 threshold (not affecting predicted labels). 
    \experimentalsetting{} 
    We follow \citet{MukhotiKSGTD20} and tune $\gamma \in \{1,2,3\}$.
    All scores are computed in their hierarchical version.   \hidecdaggerinfo{}
    }
    \label{tab:rcv1_tuning_results_performance}
\end{table*}

\begin{table*}[ht]
    \centering
    \footnotesize
    \setlength{\tabcolsep}{3pt}
\begin{tabular}{lll|rr|rrrr}
\toprule
model & loss & OS & $\uparrow$ micro\fscore & $\uparrow$ macro\fscore & $ \downarrow \text{ECE}_{\text{ML}}$ & $ \downarrow \text{ECE}_{\text{ML}}^{\text{high}}$ & $\downarrow \text{ECE}_{\text{ML}}^{\text{med}}$ & $\downarrow \text{ECE}_{\text{ML}}^{\text{rare}}$ \\
    \midrule
BERT & BCE & 0 & 88.0 & 68.1 & 8.9 & 4.1 & 8.6 & 11.2\\[-0.15cm]
 &  &  &  \tiny {$\pm 0.1$} & \tiny {$\pm 0.7$} & \tiny {$\pm 0.9$} & \tiny {$\pm 0.1$} & \tiny {$\pm 0.6$} & \tiny {$\pm 1.8$}\\
 
BERT & WBCEU & 0 & 87.7 & 67.8 & 11.1 & 4.9 & 9.7 & 15.5\\[-0.15cm]
 &  &  &  \tiny {$\pm 0.3$} & \tiny {$\pm 0.9$} & \tiny {$\pm 0.7$} & \tiny {$\pm 0.3$} & \tiny {$\pm 0.6$} & \tiny {$\pm 1.4$}\\
 
BERT & WBCEM & 0 & 84.7 & 65.5 & 15.7 & 5.2 & 13.1 & 23.2\\[-0.15cm]
 &  &  &  \tiny {$\pm 3.6$} & \tiny {$\pm 3.3$} & \tiny {$\pm 1.4$} & \tiny {$\pm 0.7$} & \tiny {$\pm 1.9$} & \tiny {$\pm 1.2$}\\

BERT & WBCEP & 0 & 86.6 & 67.7 & 14.7 & 4.9 & 12.1 & 22.0\\[-0.15cm]
 &  &  &  \tiny {$\pm 0.3$} & \tiny {$\pm 0.8$} & \tiny {$\pm 0.7$} & \tiny {$\pm 0.3$} & \tiny {$\pm 0.6$} & \tiny {$\pm 1.5$}\\
 
BERT & BCE & 10 & 88.0 & 68.3 & 10.4 & 4.2 & 8.6 & 15.2\\[-0.15cm]
 &  &  &  \tiny {$\pm 0.2$} & \tiny {$\pm 0.8$} & \tiny {$\pm 0.5$} & \tiny {$\pm 0.3$} & \tiny {$\pm 0.5$} & \tiny {$\pm 0.9$}\\
 
BERT & BCE & 20 & 87.8 & 68.0 & 9.4 & 3.5 & 7.5 & 14.3\\[-0.15cm]
 &  &  &  \tiny {$\pm 0.2$} & \tiny {$\pm 1.3$} & \tiny {$\pm 1.0$} & \tiny {$\pm 0.4$} & \tiny {$\pm 1.2$} & \tiny {$\pm 1.2$}\\
 
BERT & BCE & 30 & 87.8 & 67.6 & 10.8 & 4.3 & 8.8 & 16.0\\[-0.15cm]
 &  &  &  \tiny {$\pm 0.1$} & \tiny {$\pm 0.7$} & \tiny {$\pm 0.7$} & \tiny {$\pm 0.4$} & \tiny {$\pm 0.6$} & \tiny {$\pm 1.1$}\\
 
BERT & FL ($\gamma=1$) & 0 & 87.9 & 67.8 & 9.4 & 3.4 & 7.9 & 13.7\\[-0.15cm]
 &  &  &  \tiny {$\pm 0.3$} & \tiny {$\pm 0.8$} & \tiny {$\pm 0.5$} & \tiny {$\pm 0.4$} & \tiny {$\pm 0.3$} & \tiny {$\pm 1.2$}\\
 
BERT & FL ($\gamma=2$) & 0 & 88.3 & 68.7 & 8.8 & 3.8 & 7.0 & 13.2\\[-0.15cm]
 &  &  &  \tiny {$\pm 0.2$} & \tiny {$\pm 1.2$} & \tiny {$\pm 0.8$} & \tiny {$\pm 0.2$} & \tiny {$\pm 0.4$} & \tiny {$\pm 2.0$}\\
 
BERT & FL ($\gamma=3$) & 0 & 88.2 & 68.5 & 9.5 & 5.2 & 7.9 & 13.5\\[-0.15cm]
 &  &  &  \tiny {$\pm 0.4$} & \tiny {$\pm 0.8$} & \tiny {$\pm 0.3$} & \tiny {$\pm 0.5$} & \tiny {$\pm 0.4$} & \tiny {$\pm 0.5$}\\

HiDEC-s & HBM & 0 & 88.4 & 69.8 & 10.4 & 4.5 & 9.0 & 14.6\\[-0.15cm]
 &  &  &  \tiny {$\pm 0.4$} & \tiny {$\pm 1.0$} & \tiny {$\pm 0.7$} & \tiny {$\pm 0.1$} & \tiny {$\pm 0.4$} & \tiny {$\pm 1.4$}\\
 
$\text{HiDEC-s}^{\dagger}$ & HBM & 0 & 88.5 & 70.0 & 10.1 & 4.4 & 9.4 & 13.4\\[-0.15cm]
&  &  &  \tiny {$\pm 0.1$} & \tiny {$\pm 1.1$} & \tiny {$\pm 0.4$} & \tiny {$\pm 0.3$} & \tiny {$\pm 0.5$} & \tiny {$\pm 1.0$}\\

$\text{HiDEC-s}^{\dagger\dagger}$ & HBM & 0 & 88.5 & 69.7 & 10.9 & 4.4 & 9.3 & 15.5\\[-0.15cm]
 &  &  &  \tiny {$\pm 0.3$} & \tiny {$\pm 0.7$} & \tiny {$\pm 0.8$} & \tiny {$\pm 0.3$} & \tiny {$\pm 0.5$} & \tiny {$\pm 1.9$}\\
 
HiDEC-s & FL ($\gamma=1$) & 0 & 88.4 & 69.5 & 9.3 & 3.7 & 8.2 & 13.0\\[-0.15cm]
 &  &  &  \tiny {$\pm 0.2$} & \tiny {$\pm 0.8$} & \tiny {$\pm 1.0$} & \tiny {$\pm 0.3$} & \tiny {$\pm 0.8$} & \tiny {$\pm 2.1$}\\
HiDEC-s & FL ($\gamma=2$) & 0 & 88.4 & 68.7 & 9.3 & 3.2 & 7.8 & 13.8\\[-0.15cm]
 &  &  &  \tiny {$\pm 0.2$} & \tiny {$\pm 1.3$} & \tiny {$\pm 0.7$} & \tiny {$\pm 0.5$} & \tiny {$\pm 0.3$} & \tiny {$\pm 1.5$}\\
HiDEC-s & FL ($\gamma=3$) & 0 & 88.3 & 68.3 & 8.6 & 3.3 & 7.0 & 12.9\\[-0.15cm]
 &  &  &  \tiny {$\pm 0.3$} & \tiny {$\pm 0.7$} & \tiny {$\pm 0.7$} & \tiny {$\pm 0.2$} & \tiny {$\pm 0.5$} & \tiny {$\pm 1.4$}\\

\midrule

model & prompting & confidence & $\uparrow$ micro\fscore & $\uparrow$ macro\fscore & $ \downarrow \text{ECE}_{\text{ML}}$ & $ \downarrow \text{ECE}_{\text{ML}}^{\text{high}}$ & $\downarrow \text{ECE}_{\text{ML}}^{\text{med}}$ & $\downarrow \text{ECE}_{\text{ML}}^{\text{rare}}$ \\
\midrule

Llama-3.3 & zero-shot & Verbalized & 62.7 & 43.1 & 22.5 & 15.0 & 20.7 & 27.8\\
Llama-3.3 & zero-shot & Token Probabilities & 60.8 & 41.6 & 22.5 & 16.0 & 20.2 & 28.2\\

Llama-3.3 & RAG & Verbalized & 86.5 & 68.8 & 7.1 & 4.4 & 6.5 & 9.1\\
Llama-3.3 & RAG & Token Probabilities & 86.6 & 68.4 & 7.6 & 4.4 & 7.3 & 9.2\\

Qwen2.5 & zero-shot & Verbalized & 54.3 & 35.3 & 29.6 & 19.1 & 28.5 & 35.0\\
Qwen2.5 & zero-shot & Token Probabilities & 56.5 & 35.0 & 27.9 & 19.2 & 24.0 & 36.6\\

Qwen2.5 & RAG & Verbalized & 87.2 & 69.9 & 8.7 & 3.4 & 8.4 & 11.1\\
Qwen2.5 & RAG & Token Probabilities & 87.1 & 69.7 & 7.4 & 4.0 & 6.9 & 9.4\\

   \bottomrule
    \end{tabular}
    \vspace{-1em}
    \caption{\textbf{Performance and calibration on RCV1 dev set}. Highly frequent labels: \frequentlabels{} \rcvdevfrequentlabels{} 
    Medium-frequency: \mediumlabels{}    \rcvdevmediumlabels{}
    Rare: \rarelabels{} \rcvdevrarelabels{}
    \randomrunsdescription{} macro\fscore: hierarchical macro \fscore score.
    \hidecdaggerinfo{}
}
    \label{tab:rcv_dev_performance_and_calibration}
\end{table*}

\begin{table*}[ht]
    \centering
    \footnotesize
    \setlength{\tabcolsep}{4pt}
\begin{tabular}{lll|rrr|rrr|rrr|r}
\toprule
 & loss & OS & micP & micR & mic\fscore & macP & macR & mac\fscore & macPs & macRs & mac\fscore{s} & \exactmatch \\
    model  \\
    \midrule

BERT & BCE & 0 & 54.0 & 28.9 & 37.7 & 5.0 & 3.4 & 3.7 & 14.7 & 10.0 & 11.0 & 0.0\\[-0.15cm]
&  &  & \tiny {$\pm 1.3$} & \tiny {$\pm 0.4$} & \tiny {$\pm 0.4$} & \tiny {$\pm 0.2$} & \tiny {$\pm 0.1$} & \tiny {$\pm 0.1$} & \tiny {$\pm 0.5$} & \tiny {$\pm 0.4$} & \tiny {$\pm 0.4$} & \tiny {$\pm 0.0$}\\

ModernBERT & BCE & 0 & 60.6 & 31.3 & 41.3 & 4.8 & 3.2 & 3.5 & 14.2 & 9.4 & 10.5 & 0.0\\[-0.15cm]
&  &  & \tiny {$\pm 1.2$} & \tiny {$\pm 0.8$} & \tiny {$\pm 0.5$} & \tiny {$\pm 0.2$} & \tiny {$\pm 0.2$} & \tiny {$\pm 0.2$} & \tiny {$\pm 0.5$} & \tiny {$\pm 0.5$} & \tiny {$\pm 0.5$} & \tiny {$\pm 0.0$}\\

BiomedBERT & BCE & 0 & 56.5 & 31.9 & 40.8 & 5.6 & 4.0 & 4.3 & 16.7 & 11.8 & 12.8 & 0.1\\[-0.15cm]
& & & \tiny {$\pm 0.9$} & \tiny {$\pm 0.3$} & \tiny {$\pm 0.2$} & \tiny {$\pm 0.1$} & \tiny {$\pm 0.1$} & \tiny {$\pm 0.1$} & \tiny {$\pm 0.4$} & \tiny {$\pm 0.3$} & \tiny {$\pm 0.2$} & \tiny {$\pm 0.0$}\\

BiomedBERT & WBCEU & 0 & 50.3 & 36.1 & 42.0 & 5.5 & 4.7 & 4.7 & 16.4 & 13.9 & 14.0 & 0.0\\[-0.15cm]
 &  &  & \tiny {$\pm 1.3$} & \tiny {$\pm 0.9$} & \tiny {$\pm 0.4$} & \tiny {$\pm 0.1$} & \tiny {$\pm 0.2$} & \tiny {$\pm 0.1$} & \tiny {$\pm 0.2$} & \tiny {$\pm 0.6$} & \tiny {$\pm 0.3$} & \tiny {$\pm 0.0$}\\

BiomedBERT & WBCEM & 0 & 0.1 & 21.2 & 0.3 & 0.0 & 9.7 & 0.1 & 0.1 & 28.7 & 0.2 & 0.0\\[-0.15cm]
 &  &  & \tiny {$\pm 0.1$} & \tiny {$\pm 21.1$} & \tiny {$\pm 0.1$} & \tiny {$\pm 0.0$} & \tiny {$\pm 5.3$} & \tiny {$\pm 0.1$} & \tiny {$\pm 0.1$} & \tiny {$\pm 15.8$} & \tiny {$\pm 0.2$} & \tiny {$\pm 0.0$}\\

BiomedBERT & WBCEP & 0 & 0.2 & 20.3 & 0.4 & 0.0 & 8.6 & 0.1 & 0.1 & 25.4 & 0.2 & 0.0\\[-0.15cm]
&  &  & \tiny {$\pm 0.1$} & \tiny {$\pm 10.4$} & \tiny {$\pm 0.1$} & \tiny {$\pm 0.0$} & \tiny {$\pm 4.2$} & \tiny {$\pm 0.0$} & \tiny {$\pm 0.1$} & \tiny {$\pm 12.5$} & \tiny {$\pm 0.1$} & \tiny {$\pm 0.0$}\\

BiomedBERT & BCE & 10 & 55.5 & 32.1 & 40.7 & 5.6 & 4.1 & 4.4 & 16.7 & 12.2 & 13.1 & 0.0\\[-0.15cm]
&  &  & \tiny {$\pm 0.6$} & \tiny {$\pm 0.5$} & \tiny {$\pm 0.3$} & \tiny {$\pm 0.1$} & \tiny {$\pm 0.1$} & \tiny {$\pm 0.1$} & \tiny {$\pm 0.2$} & \tiny {$\pm 0.2$} & \tiny {$\pm 0.3$} & \tiny {$\pm 0.0$}\\

BiomedBERT & BCE & 20 & 56.0 & 31.7 & 40.5 & 5.6 & 4.0 & 4.3 & 16.6 & 11.9 & 12.9 & 0.0\\[-0.15cm]
 &  &  & \tiny {$\pm 1.0$} & \tiny {$\pm 0.4$} & \tiny {$\pm 0.1$} & \tiny {$\pm 0.2$} & \tiny {$\pm 0.2$} & \tiny {$\pm 0.2$} & \tiny {$\pm 0.5$} & \tiny {$\pm 0.5$} & \tiny {$\pm 0.5$} & \tiny {$\pm 0.1$}\\
 
BiomedBERT & BCE & 30 & 55.1 & 32.1 & 40.5 & 5.5 & 4.0 & 4.3 & 16.3 & 12.0 & 12.8 & 0.1\\[-0.15cm]
 &  &  & \tiny {$\pm 0.7$} & \tiny {$\pm 0.5$} & \tiny {$\pm 0.3$} & \tiny {$\pm 0.1$} & \tiny {$\pm 0.1$} & \tiny {$\pm 0.1$} & \tiny {$\pm 0.2$} & \tiny {$\pm 0.3$} & \tiny {$\pm 0.2$} & \tiny {$\pm 0.1$}\\

BiomedBERT & FL ($\gamma=1$) & 0 & 57.2 & 32.1 & 41.1 & 5.8 & 4.0 & 4.4 & 17.1 & 12.0 & 13.0 & 0.0\\[-0.15cm]
& & & \tiny {$\pm 1.1$} & \tiny {$\pm 0.2$} & \tiny {$\pm 0.3$} & \tiny {$\pm 0.1$} & \tiny {$\pm 0.2$} & \tiny {$\pm 0.2$} & \tiny {$\pm 0.4$} & \tiny {$\pm 0.5$} & \tiny {$\pm 0.5$} & \tiny {$\pm 0.1$}\\

BiomedBERT & FL ($\gamma=2$) & 0 & 57.8 & 31.3 & 40.6 & 5.7 & 4.0 & 4.3 & 16.8 & 11.8 & 12.9 & 0.0\\[-0.15cm]
& & & \tiny {$\pm 1.9$} & \tiny {$\pm 0.6$} & \tiny {$\pm 0.4$} & \tiny {$\pm 0.2$} & \tiny {$\pm 0.2$} & \tiny {$\pm 0.1$} & \tiny {$\pm 0.5$} & \tiny {$\pm 0.5$} & \tiny {$\pm 0.4$} & \tiny {$\pm 0.1$}\\

BiomedBERT & FL ($\gamma=3$) & 0 & 58.4 & 31.0 & 40.5 & 5.6 & 3.8 & 4.2 & 16.7 & 11.4 & 12.5 & 0.0\\[-0.15cm]
& & & \tiny {$\pm 1.6$} & \tiny {$\pm 0.5$} & \tiny {$\pm 0.2$} & \tiny {$\pm 0.1$} & \tiny {$\pm 0.1$} & \tiny {$\pm 0.1$} & \tiny {$\pm 0.3$} & \tiny {$\pm 0.2$} & \tiny {$\pm 0.3$} & \tiny {$\pm 0.0$}\\

PLM-ICD & BCE & 0 & 61.9 & 45.0 & 52.1 & 8.2 & 6.8 & 7.0 & 24.4 & 20.3 & 20.7 & 0.2\\[-0.15cm]
& & & \tiny {$\pm 3.6$} & \tiny {$\pm 2.2$} & \tiny {$\pm 2.7$} & \tiny {$\pm 0.6$} & \tiny {$\pm 0.5$} & \tiny {$\pm 0.5$} & \tiny {$\pm 1.9$} & \tiny {$\pm 1.6$} & \tiny {$\pm 1.6$} & \tiny {$\pm 0.2$}\\

PLM-ICD & FL ($\gamma=1$) & 0 & 60.6 & 46.8 & 52.8 & 9.1 & 7.7 & 7.8 & 26.9 & 22.9 & 23.2 & 0.1\\[-0.15cm]
& & & \tiny {$\pm 2.0$} & \tiny {$\pm 0.4$} & \tiny {$\pm 0.8$} & \tiny {$\pm 0.1$} & \tiny {$\pm 0.1$} & \tiny {$\pm 0.1$} & \tiny {$\pm 0.3$} & \tiny {$\pm 0.3$} & \tiny {$\pm 0.2$} & \tiny {$\pm 0.1$}\\

PLM-ICD & FL ($\gamma=2$) & 0 & 62.1 & 46.7 & 53.3 & 9.1 & 7.7 & 7.8 & 27.1 & 22.7 & 23.2 & 0.2\\[-0.15cm]
&  &  & \tiny {$\pm 1.7$} & \tiny {$\pm 0.8$} & \tiny {$\pm 1.0$} & \tiny {$\pm 0.2$} & \tiny {$\pm 0.2$} & \tiny {$\pm 0.2$} & \tiny {$\pm 0.5$} & \tiny {$\pm 0.6$} & \tiny {$\pm 0.5$} & \tiny {$\pm 0.1$}\\

PLM-ICD & FL ($\gamma=3$) & 0 & 60.5 & 46.7 & 52.7 & 9.0 & 7.7 & 7.8 & 26.7 & 22.8 & 23.2 & 0.1\\[-0.15cm]
& & & \tiny {$\pm 1.4$} & \tiny {$\pm 0.3$} & \tiny {$\pm 0.6$} & \tiny {$\pm 0.2$} & \tiny {$\pm 0.1$} & \tiny {$\pm 0.1$} & \tiny {$\pm 0.5$} & \tiny {$\pm 0.3$} & \tiny {$\pm 0.3$} & \tiny {$\pm 0.0$}\\

    \bottomrule
    \end{tabular}
    \caption{\textbf{Performance on MIMIC-III dev set.} BERT: bert-base-uncased, ModernBERT: ModernBERT-base, BiomedBERT: BiomedBERT-base-uncased-abstract, each using the CLS token for classification. PLM-ICD: PLM-ICD \citep{huang-etal-2022-plm} with BiomedBERT-base-uncased-abstract. \experimentalsetting{}
    We follow \citet{MukhotiKSGTD20} and tune $\gamma \in \{1,2,3\}$.
    \macsexplanation{}
    }
    \label{tab:mimic_dev_performance}
\end{table*}

\begin{table*}[ht]
    \centering
    \footnotesize
    \setlength{\tabcolsep}{3pt}
\begin{tabular}{lll|rr|rrrr}
\toprule
model & loss & OS & $\uparrow$ micro\fscore & $\uparrow$ macro\fscore & $ \downarrow \text{ECE}_{\text{ML}}$ & $ \downarrow \text{ECE}_{\text{ML}}^{\text{high}}$ & $\downarrow \text{ECE}_{\text{ML}}^{\text{med}}$ & $\downarrow \text{ECE}_{\text{ML}}^{\text{rare}}$ \\
    \midrule
BERT & BCE & 0 & 37.7 & 11.0 & 3.2 & 18.9 & 17.2 & 1.4\\[-0.15cm]
& & & \tiny {$\pm 0.4$} & \tiny {$\pm 0.4$} & \tiny {$\pm 0.1$} & \tiny {$\pm 0.4$} & \tiny {$\pm 0.8$} & \tiny {$\pm 0.1$}\\
    
ModernBERT & BCE & 0 & 41.3 & 10.5 & 2.6 & 18.8 & 15.4 & 0.9\\[-0.15cm]
& & & \tiny {$\pm 0.5$} & \tiny {$\pm 0.5$} & \tiny {$\pm 0.1$} & \tiny {$\pm 0.3$} & \tiny {$\pm 0.4$} & \tiny {$\pm 0.1$}\\

BiomedBERT & BCE & 0 & 40.8 & 12.8 & 3.2 & 17.1 & 16.7 & 1.5\\[-0.15cm]
& & & \tiny {$\pm 0.2$} & \tiny {$\pm 0.2$} & \tiny {$\pm 0.2$} & \tiny {$\pm 0.5$} & \tiny {$\pm 0.5$} & \tiny {$\pm 0.1$}\\

BiomedBERT & WBCEU & 0 & 42.0 & 14.0 & 3.8 & 16.5 & 18.1 & 2.0\\[-0.15cm]
& & & \tiny {$\pm 0.4$} & \tiny {$\pm 0.3$} & \tiny {$\pm 0.2$} & \tiny {$\pm 0.6$} & \tiny {$\pm 0.8$} & \tiny {$\pm 0.1$}\\

BiomedBERT & WBCEM & 0 & 0.3 & 0.2 & 42.1 & 38.3 & 45.5 & 41.8\\[-0.15cm]
& & & \tiny {$\pm 0.1$} & \tiny {$\pm 0.2$} & \tiny {$\pm 5.8$} & \tiny {$\pm 3.9$} & \tiny {$\pm 3.7$} & \tiny {$\pm 6.1$}\\

BiomedBERT & WBCEP & 0 & 0.4 & 0.2 & 42.7 & 39.0 & 45.0 & 42.5\\[-0.15cm]
& & & \tiny {$\pm 0.1$} & \tiny {$\pm 0.1$} & \tiny {$\pm 4.4$} & \tiny {$\pm 1.8$} & \tiny {$\pm 2.3$} & \tiny {$\pm 4.6$}\\

BiomedBERT & BCE & 10 & 40.7 & 13.1 & 3.5 & 17.4 & 17.6 & 1.7\\[-0.15cm]
& & & \tiny {$\pm 0.3$} & \tiny {$\pm 0.3$} & \tiny {$\pm 0.2$} & \tiny {$\pm 1.0$} & \tiny {$\pm 0.9$} & \tiny {$\pm 0.1$}\\

BiomedBERT & BCE & 20 & 40.5 & 12.9 & 3.4 & 17.4 & 17.3 & 1.7\\[-0.15cm]
& & & \tiny {$\pm 0.1$} & \tiny {$\pm 0.5$} & \tiny {$\pm 0.2$} & \tiny {$\pm 0.5$} & \tiny {$\pm 0.5$} & \tiny {$\pm 0.1$}\\

BiomedBERT & BCE & 30 & 40.5 & 12.8 & 3.5 & 17.7 & 17.9 & 1.7\\[-0.15cm]
& & & \tiny {$\pm 0.3$} & \tiny {$\pm 0.2$} & \tiny {$\pm 0.2$} & \tiny {$\pm 0.7$} & \tiny {$\pm 0.6$} & \tiny {$\pm 0.1$}\\

BiomedBERT & FL ($\gamma=1$) & 0 & 41.1 & 13.0 & 2.9 & 12.6 & 14.5 & 1.5\\[-0.15cm]
& & & \tiny {$\pm 0.3$} & \tiny {$\pm 0.5$} & \tiny {$\pm 0.2$} & \tiny {$\pm 1.1$} & \tiny {$\pm 1.1$} & \tiny {$\pm 0.1$}\\

BiomedBERT & FL ($\gamma=2$) & 0 & 40.6 & 12.9 & 3.1 & 12.1 & 14.2 & 1.8\\[-0.15cm]
& & & \tiny {$\pm 0.4$} & \tiny {$\pm 0.4$} & \tiny {$\pm 0.1$} & \tiny {$\pm 0.4$} & \tiny {$\pm 0.7$} & \tiny {$\pm 0.1$}\\

BiomedBERT & FL ($\gamma=3$) & 0 & 40.5 & 12.5 & 4.1 & 14.7 & 15.5 & 2.7\\[-0.15cm]
& & & \tiny {$\pm 0.2$} & \tiny {$\pm 0.3$} & \tiny {$\pm 0.6$} & \tiny {$\pm 1.2$} & \tiny {$\pm 0.5$} & \tiny {$\pm 0.6$}\\

PLM-ICD & BCE & 0 & 52.1 & 20.7 & 4.6 & 10.3 & 14.7 & 3.4\\[-0.15cm]
& & & \tiny {$\pm 2.7$} & \tiny {$\pm 1.6$} & \tiny {$\pm 1.0$} & \tiny {$\pm 2.6$} & \tiny {$\pm 3.0$} & \tiny {$\pm 0.7$}\\
PLM-ICD & FL ($\gamma=1$) & 0 & 52.8 & 23.2 & 4.7 & 10.1 & 14.8 & 3.5\\[-0.15cm]
& & & \tiny {$\pm 0.8$} & \tiny {$\pm 0.2$} & \tiny {$\pm 0.4$} & \tiny {$\pm 1.4$} & \tiny {$\pm 1.4$} & \tiny {$\pm 0.3$}\\
PLM-ICD & FL ($\gamma=2$) & 0 & 53.3 & 23.2 & 4.6 & 9.7 & 13.5 & 3.6\\[-0.15cm]
& & & \tiny {$\pm 1.0$} & \tiny {$\pm 0.5$} & \tiny {$\pm 0.2$} & \tiny {$\pm 0.3$} & \tiny {$\pm 0.7$} & \tiny {$\pm 0.1$}\\
PLM-ICD & FL ($\gamma=3$) & 0 & 52.7 & 23.2 & 5.4 & 11.3 & 14.6 & 4.3\\[-0.15cm]
& & & \tiny {$\pm 0.6$} & \tiny {$\pm 0.3$} & \tiny {$\pm 0.2$} & \tiny {$\pm 0.9$} & \tiny {$\pm 0.3$} & \tiny {$\pm 0.2$}\\

   \bottomrule
    \end{tabular}
    \vspace{-1em}
    \caption{\textbf{Performance and calibration on MIMIC-III dev set}. Highly frequent labels: \frequentlabels{} \mimictestfrequentlabels{} 
    Medium-frequency: \mediumlabels{}    \mimictestmediumlabels{}
    Rare: \rarelabels{} \mimicdevrarelabels{}
    \randomrunsdescription{}
    macro\fscore: Macro \fscore on labels with support in the test set.
}
    \label{tab:mimic_dev_performance_and_calibration}
\end{table*}

\paragraph{Model Sizes.} See \tref{tab:model_sizes} for an overview of the number of parameters of the models used in our experiments.

\clearpage
\subsubsection{Results and Discussion}
\label{sec:appendix_experiments_results}
\begin{figure*}[htb]
  \begin{center}
    \begin{subfigure}[t]{\textwidth}
      \centering
      \includegraphics[width=\linewidth]{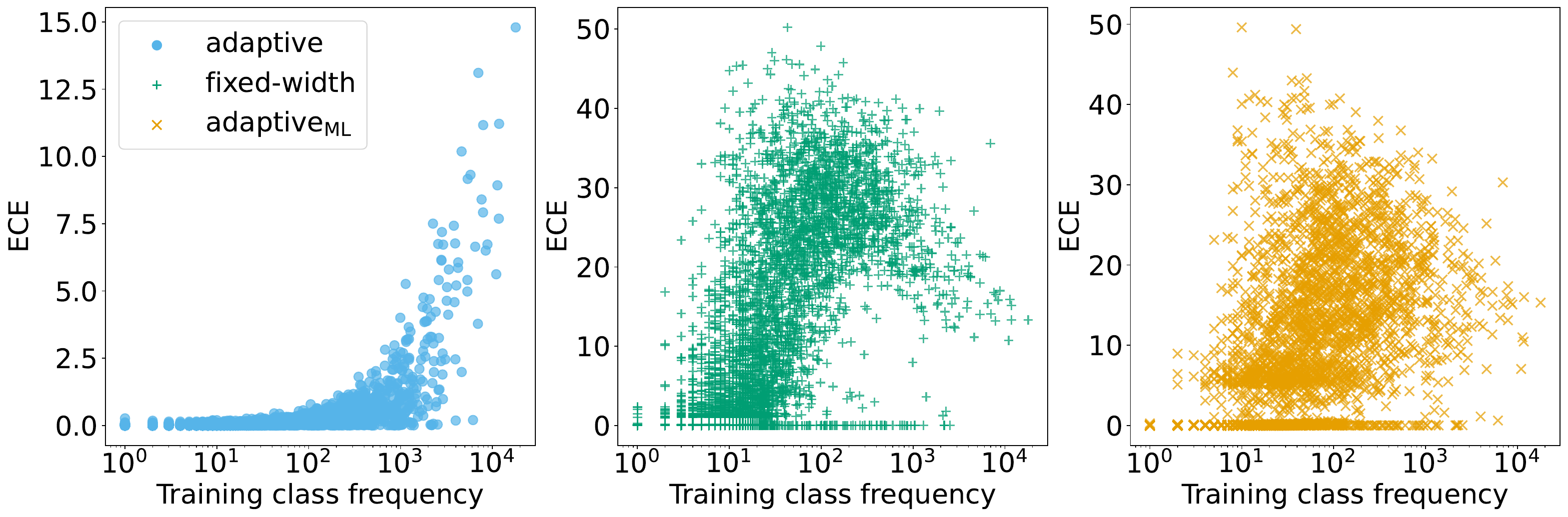}
      \caption{MIMIC-III test} %
      \label{fig:ece_by_scheme_mimic_test}
    \end{subfigure}
    \hfill
    \begin{subfigure}[t]{\textwidth}
      \centering
      \includegraphics[width=\linewidth]{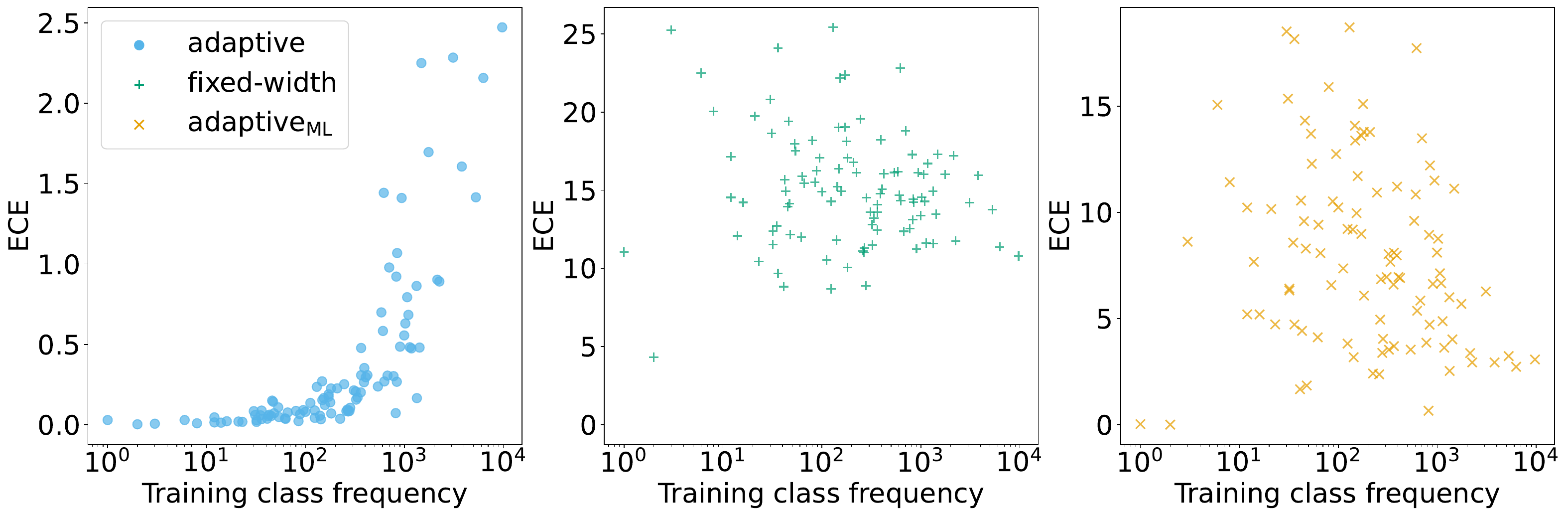}
      \caption{RCV1 test} %
      \label{fig:ece_by_scheme_rcv_test}
    \end{subfigure}
  \end{center}
  \caption{\textbf{Expected Calibration Error (ECE) as a function of training class frequency under different binning schemes} (adaptive, fixed-width, and \ourScheme). Results are shown for (a) MIMIC-III test using BiomedBERT-base and (b) RCV1 test using BERT-base, both trained with BCE loss and no oversampling, averaged over 5 random seeds. }
  \label{fig:ece_by_scheme_test}
\end{figure*}

\begin{figure*}[htb]
  \begin{center}
    \begin{subfigure}[t]{0.24\textwidth}
      \centering
      \includegraphics[width=\linewidth]{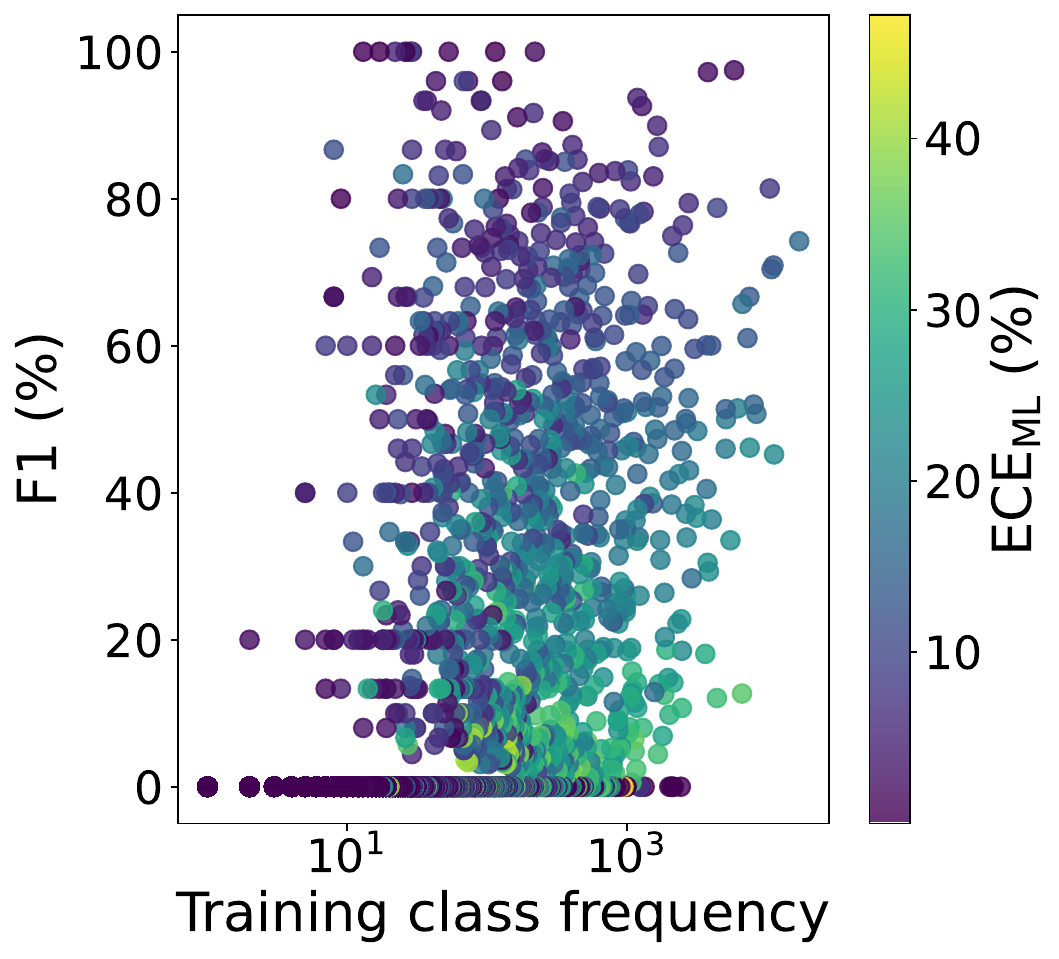}
      \caption{MIMIC-III dev}
      \label{fig:performance_frequency_mimic_dev}
    \end{subfigure}
    \hfill
    \begin{subfigure}[t]{0.24\textwidth}
      \centering
      \includegraphics[width=\linewidth]{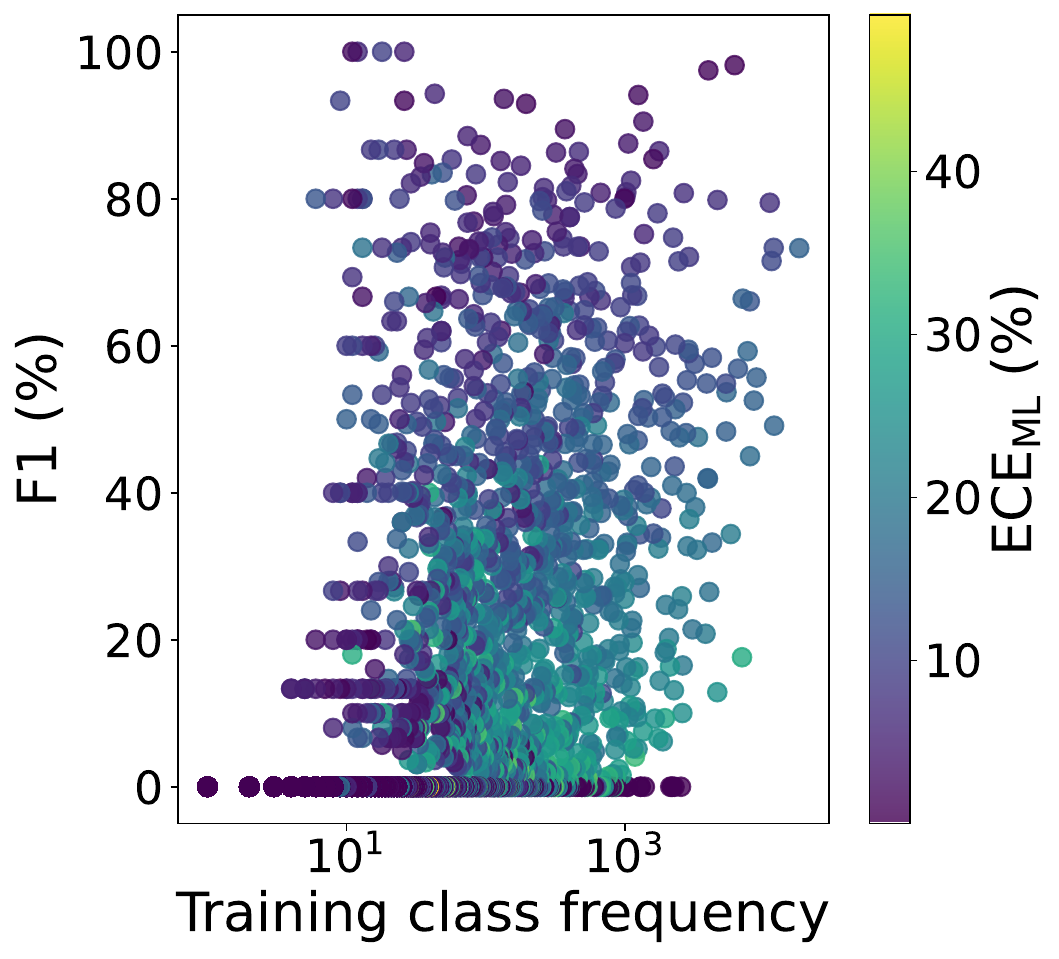}
      \caption{MIMIC-III test}
      \label{fig:performance_frequency_mimic_test}
    \end{subfigure}
    \hfill
    \begin{subfigure}[t]{0.24\textwidth}
      \centering
      \includegraphics[width=\linewidth]{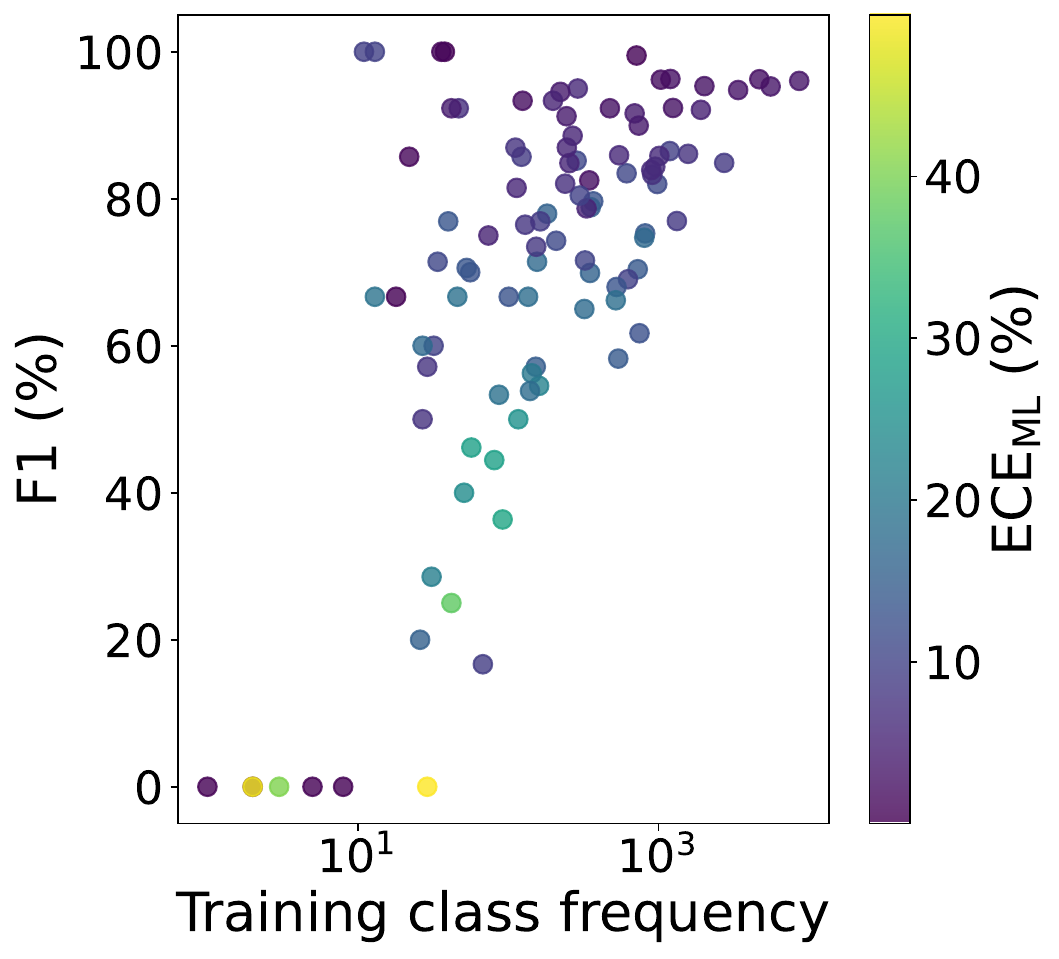}
      \caption{RCV1 dev}
      \label{fig:performance_frequency_rcv_dev}
    \end{subfigure}
    \hfill
    \begin{subfigure}[t]{0.24\textwidth}
      \centering
      \includegraphics[width=\linewidth]{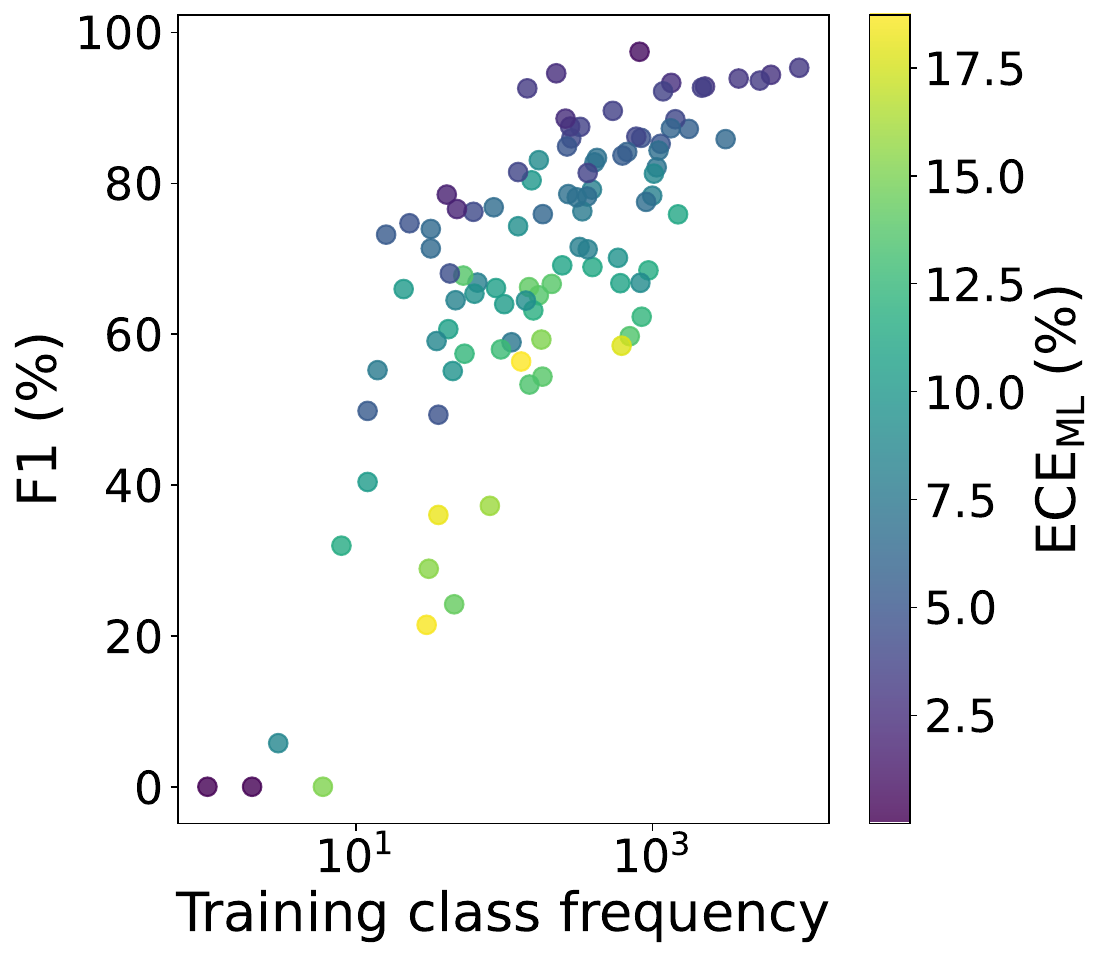}
      \caption{RCV1 test}
      \label{fig:performance_frequency_rcv_test}
    \end{subfigure}
  \end{center}
  \caption{\textbf{\fscore score as a function of training class frequency}. Results are shown for (a, b) MIMIC-III using BiomedBERT-base and (c, d) RCV1 using BERT-base, both trained with BCE loss and no oversampling, averaged over 5 random seeds. $\text{ECE}_{\text{ML}}$ scores: scores computed based on \ourScheme binning.}
  \label{fig:performance_frequency}
\end{figure*}

\textbf{Comparison of Binning Schemes.}
\fref{fig:ece_by_scheme_test} compares the calibration errors based on fixed-width, adaptive, and \ourScheme binning on the test sets of MIMIC-III and RCV1, respectively. 
Additionally, \fref{fig:performance_frequency} shows training frequency vs. \fscore score per label, with each marker indicating the respective $\text{ECE}_{\text{ML}}$ score.
\trefplural{tab:bin_sizes_mimic_test} and \ref{tab:bin_sizes_rcv_test} show the distribution of bin sizes on the test sets of MIMIC-III and RCV1, respectively.

\begin{table*}[h]
    \centering
    \footnotesize
    \setlength{\tabcolsep}{3pt}
\begin{tabular}{l|r|l|rrrr|rrr}
\toprule
& & & \multicolumn{4}{c|}{\textbf{Share (\%) of bins of size}} &\multicolumn{3}{c}{\textbf{Bin size statistics}} \\
\textbf{Dataset} & \textbf{Ideal bin size} & \textbf{Binning scheme} & <= 10 & <= 100 & <= 1000 & <10 000 & Average & Std. dev.  & Median\\
\midrule

\multirow{4}{*}{\vspace{0.4cm}MIMIC-III test} & \multirow{4}{*}{\vspace{0.4cm}337.2} & fixed-width & 46.7 & 53.9 & 54.4 & 100 & 1539.0 & 1671.8 & 18.8\\[-0.15cm]
&  &  & \tiny {$\pm 0.5$} & \tiny {$\pm 0.6$} & \tiny {$\pm 0.6$} & \tiny {$\pm 0.0$} & \tiny {$\pm 20.8$} & \tiny {$\pm 2.3$} & \tiny {$\pm 1.8$}\\
 & & \ourScheme & \textbf{2.9} & \textbf{3.7} & \textbf{98.1} & 100 & \textbf{378.9} & \textbf{371.6} & \textbf{337.0}\\[-0.15cm]
&  &  & \tiny {$\pm 0.1$} & \tiny {$\pm 0.2$} & \tiny {$\pm 0.1$} & \tiny {$\pm 0.0$} & \tiny {$\pm 1.5$} & \tiny {$\pm 6.3$} & \tiny {$\pm 0.0$}\\[-0.1cm]

   \bottomrule
    \end{tabular}
    \vspace{-1em}
    \caption{\textbf{Distribution of bin sizes of fixed-width and \ourScheme binning schemes for MIMIC-III test.} \blue{See \tref{tab:bin_sizes} for more details.}}
    \label{tab:bin_sizes_mimic_test}
\end{table*}

\begin{table*}[h]
    \centering
    \footnotesize
    \setlength{\tabcolsep}{3pt}
\begin{tabular}{l|r|l|rrrrrr|rrr}
\toprule
& & & \multicolumn{6}{c|}{\textbf{Share (\%) of bins of size $b$}} &\multicolumn{3}{c}{\textbf{Bin size statistics}} \\
\textbf{Dataset} & \textbf{Ideal bin size} & \textbf{Binning scheme} & <= 10 & <= 100 & <= 1000 & <= 10 000 & <=100K & < 1M & Average & Std. dev.  & Median\\
\midrule

 \midrule
\multirow{4}{*}{\vspace{0.4cm}RCV1 test} & \multirow{4}{*}{\vspace{0.4cm}78126.1} & fixed-width & 1.3 & 12.0 & 60.0 & 85.4 & 89.0 & 100.0 & 81798.7 & 230853.8 & 661.6\\[-0.15cm]
 & &  & \tiny {$\pm 0.3$} & \tiny {$\pm 2.1$} & \tiny {$\pm 4.2$}  & \tiny {$\pm 0.1$} & \tiny {$\pm 0.1$} & \tiny {$\pm 0.0$} & \tiny {$\pm 614.6$} & \tiny {$\pm 533.6$} & \tiny {$\pm 113.7$}\\
 
& & \ourScheme & \textbf{0.4} & \textbf{2.2} & \textbf{17.6} & \textbf{42.1} & \textbf{52.9} & 100.0 & \textbf{78771.9} & \textbf{75103.8} & \textbf{78126.0}\\[-0.15cm]
&  &  & \tiny {$\pm 0.3$} & \tiny {$\pm 0.5$} & \tiny {$\pm 0.7$} & \tiny {$\pm 0.7$} & \tiny {$\pm 0.3$} & \tiny {$\pm 0.0$} & \tiny {$\pm 581.2$} & \tiny {$\pm 3329.3$} & \tiny {$\pm 0.0$}\\

   \bottomrule
    \end{tabular}
    \vspace{-1em}
    \caption{\textbf{Distribution of bin sizes of fixed-width and \ourScheme binning schemes for RCV1 test.} \blue{Note that with \ourScheme, 10.8\% of bins falls in the range of bin size $10000 < b \leq 100000$, while for fixed-width, only 3.6\% of bins fall in this range. See \tref{tab:bin_sizes} for more details.}}
    \label{tab:bin_sizes_rcv_test}
\end{table*}

\begin{table*}[t]
    \centering
    \footnotesize
    \setlength\tabcolsep{4pt}
    \begin{tabular}{llrrrrrr}
        \toprule
         Model & Prompt & Dev & Dev (per inst.) & Subsampled Test & Subsampled Test (per inst.) & Full Test & Full Test (per inst.) \\
        \midrule
        HiDEC & – & 10 s & $\approx$ 4.3 ms & ~15 s & $\approx$ 3.0 ms & ~30 min & $\approx$ 2.3 ms \\
        Llama-3.3 & Zero-Shot & 3.1 h & $\approx$ 4.8 s & 6.5 h & $\approx$ 4.7 s & - & - \\
        Qwen2.5 & Zero-Shot & 3.9 h & $\approx$ 6.1 s & 7.8 h & $\approx$ 5.6 s & - & - \\
        Llama-3.3 & RAG & 17.5 h & $\approx$ 27.2 s & 40.8 h & $\approx$ 29.4 s & - & - \\
        Qwen2.5 & RAG & 20.0 h & $\approx$ 31.1 s & 43.2 h & $\approx$ 31.1 s & - & - \\
        \bottomrule
    \end{tabular}
    \caption{\textbf{Inference runtimes in total and per instance for each model across the development, subsampled test, and full test splits of RCV1 dataset} measured using MLFlow, on 4 NVIDIA H100 GPUs. The inference runtimes for HiDEC are averaged over 5 runs with different random seeds. Inference runtimes for the BERT-based models are not included in this table, as we ran the experiments with these models on a single NVIDIA A100 GPU; hence, inference times are not comparable.}
    \label{tab:runtimes_rcv1}
\end{table*}

\noindent \textbf{Training and Inference Run Times.}
In this section, we compare the time required for training and inference across different approaches, highlighting trade-offs between LLM-based and supervised fine-tuning methods. An overview of the inference times for the RCV1 dataset splits is provided in \tref{tab:runtimes_rcv1}.

LLMs require substantial inference time, especially in the RAG setting, where the prompt length increases due to the added demonstrations. For the RCV1 dataset, the size of the official test set presents a practical challenge. While the development set contains only 2,315 instances, inference with our RAG approach in the few-shot setting already takes up to 20 hours on 4 NVIDIA H100 GPUs. The full test set, however, comprises 781,261 instances, which would result in an estimated inference time of over 9 months.
This computational bottleneck motivated the creation of a subsampled test set, as described in \aref{sec:appendix_datasets}. Even on this reduced subset, inference with RAG can take over 43 hours. 

In comparison, inference with HiDEC is significantly faster: it takes around 10 seconds on the development set, 15 seconds on the subsampled test set (5,000 instances), and approximately 30 minutes on the full test set. However, unlike the RAG approach, HiDEC requires training. Using HBM Loss, training HiDEC for 100 epochs without early stopping takes roughly 2.5 hours, with only minimal variation depending on whether the training set includes the tuning set.
Even when considering both training and inference, the total runtime for HiDEC remains substantially lower than that of the RAG-based methods. Moreover, training HiDEC for 50 epochs with early stopping yields nearly the same performance as training for 100 epochs without it, making the efficiency gap even more pronounced.

\begin{figure}[htb]
\begin{subfigure}{\linewidth}
  \centering
  \includegraphics[width=\linewidth]{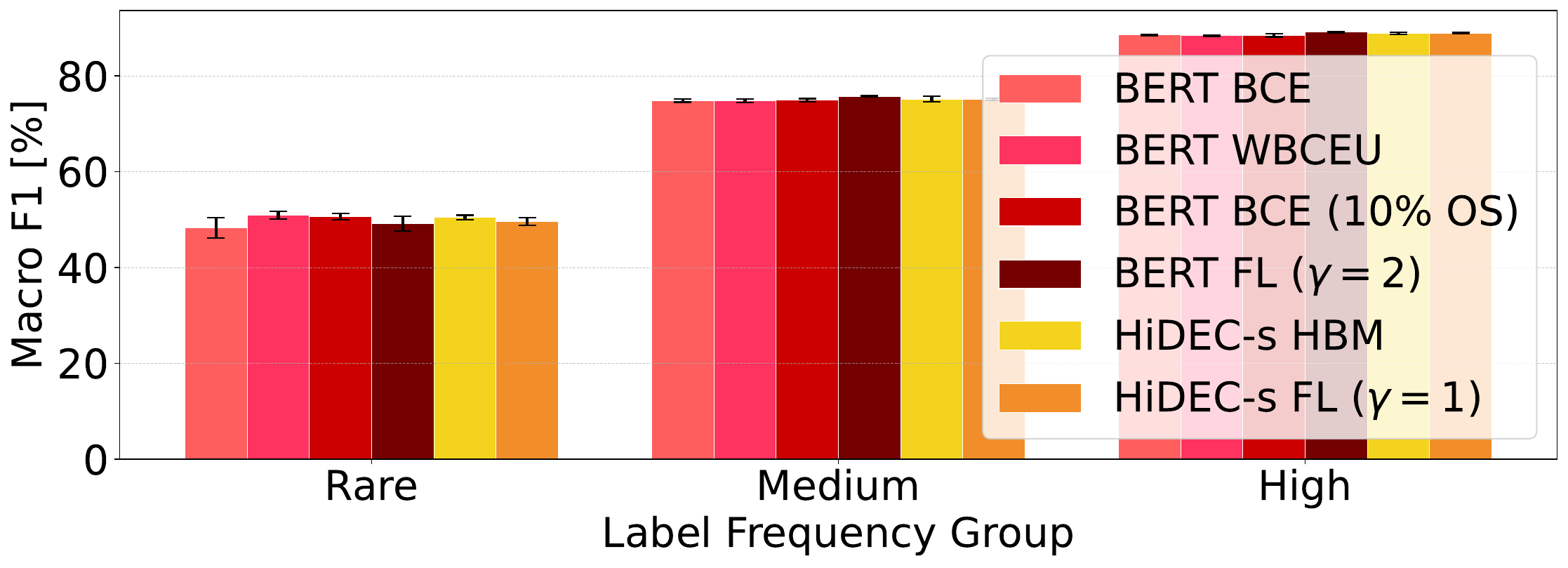} 
  \caption{$\uparrow$ Macro \fscore}
  \label{fig:rcv1_test_frequency_groups_performance}
\end{subfigure}
  \begin{center}
    \begin{subfigure}{\linewidth}
      \centering
      \includegraphics[width=\linewidth]{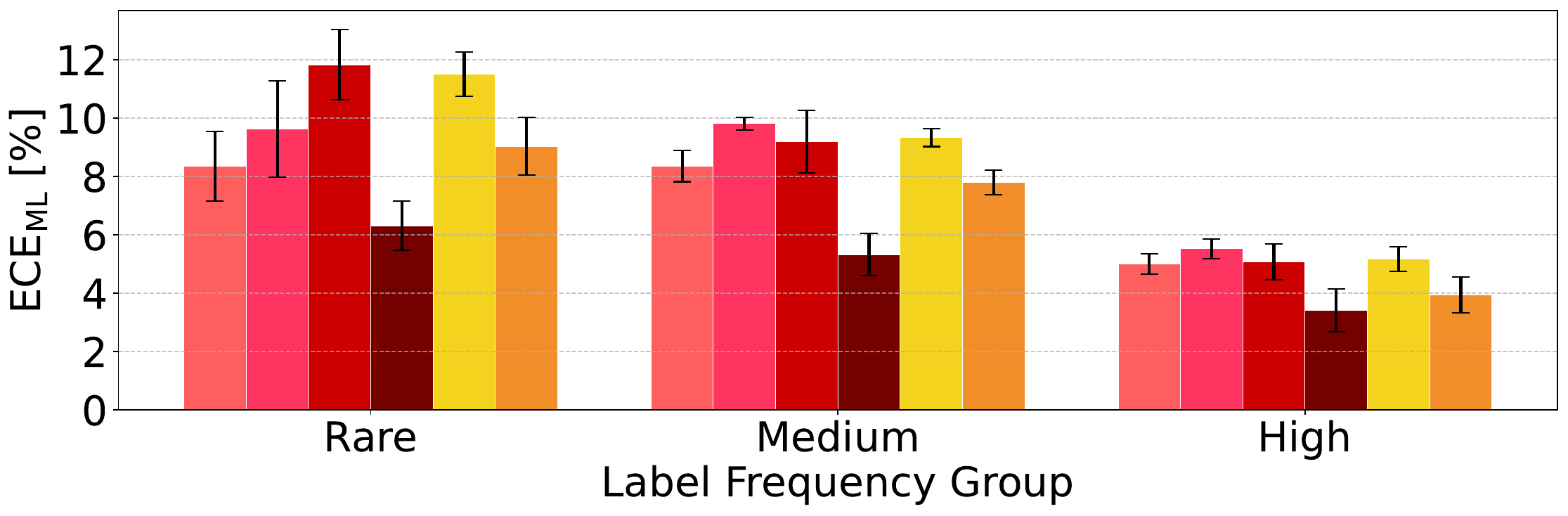} 
      \caption{$\downarrow \text{ECE}_{\text{ML}}$} 
      \label{fig:rcv1_test_frequency_groups_ece}
    \end{subfigure}
    \hfill
\end{center}
  \caption{\textbf{Performance and calibration on RCV1 test across frequency groups.} For experimental details, see \tref{tab:rcv1_performance_and_calibration_full}. %
  }
  \label{fig:rcv1_test_frequency_groups_performance_and_calibration}
\end{figure}

\noindent \textbf{Additional Results on RCV1 Test Set.}
\trefplural{tab:rcv1_performance_and_calibration_full} and \ref{tab:rcv_test_performance} contain additional results for HiDEC (matching \citet{kim-etal-2024-hierarchy}'s experimental setting) and further performance metrics on the full RCV1 test set.
\fref{fig:rcv1_test_frequency_groups_performance_and_calibration} shows performance and calibration across frequency groups on that test set.

\begin{figure}[htb]
    \begin{subfigure}{\linewidth}
  \centering
  \includegraphics[width=\linewidth]{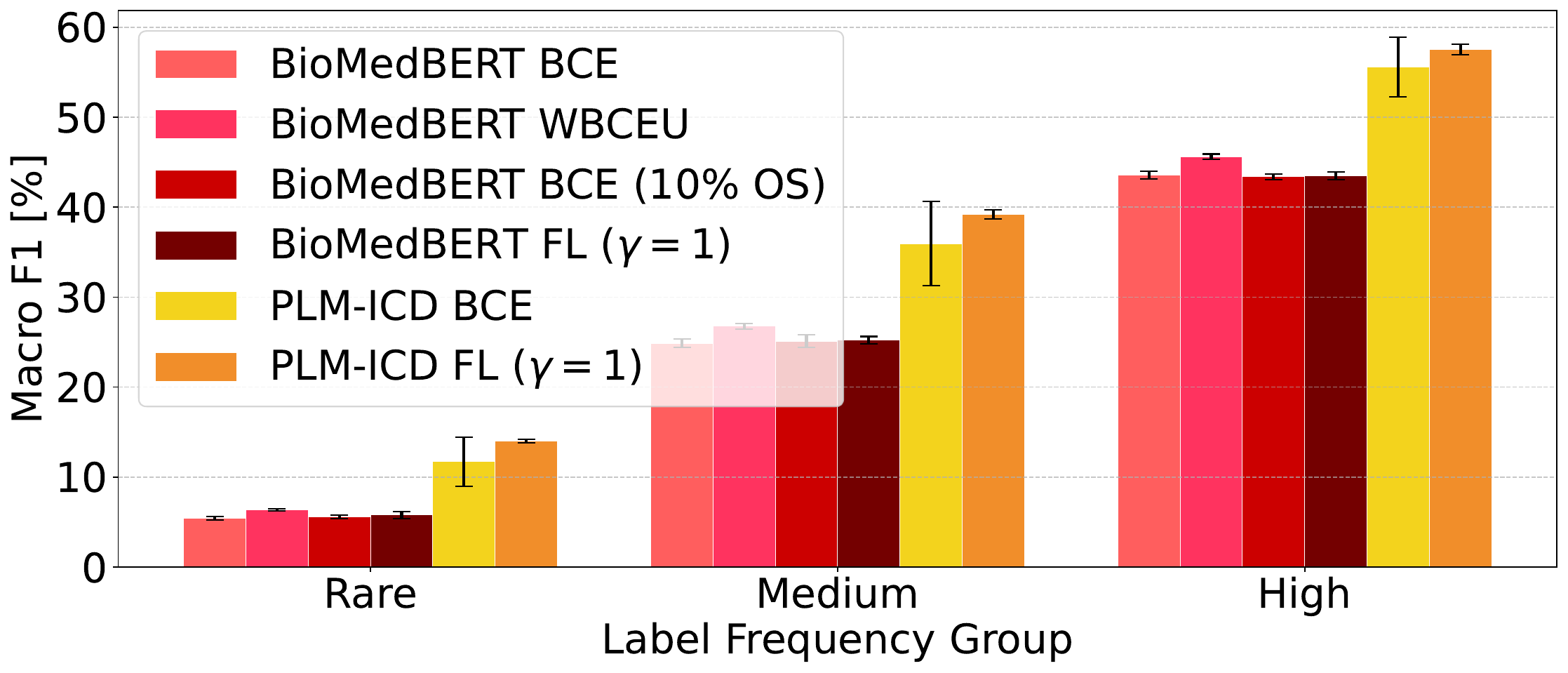}
  \caption{$\uparrow$ Macro \fscore}
  \label{fig:mimic_test_frequency_groups_performance}
  \end{subfigure}

    \hfill
    \begin{subfigure}{\linewidth}
      \centering
      \includegraphics[width=\linewidth]{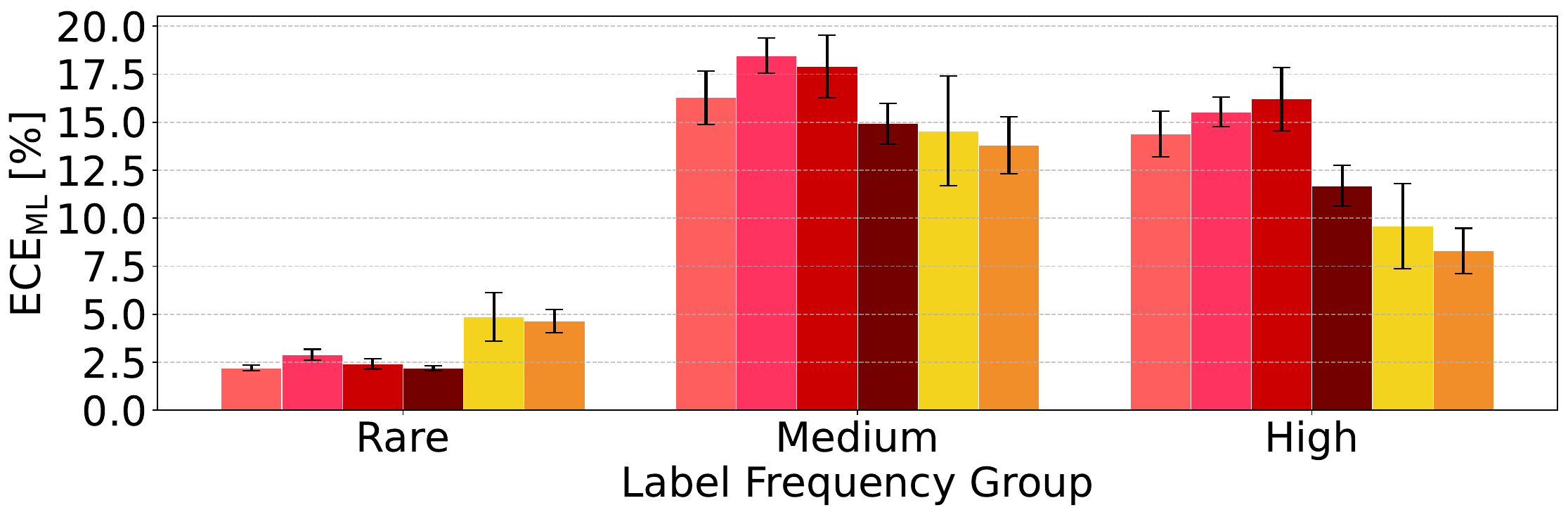}
      \caption{$\downarrow \text{ECE}_{\text{ML}}$}
      \label{fig:mimic_test_frequency_groups_ece_with_all_labels}
    \end{subfigure}

  \caption{\textbf{Performance and calibration on MIMIC-III test across frequency groups.} Compared to RCV1, the performance on MIMIC-III is even more dependent on label frequency. 
  For experimental details, see \tref{tab:mimic_test_performance}.}
  \label{fig:mimic_test_frequency_groups_calibration}
\end{figure}

\noindent \textbf{Results on MIMIC-III Test Set.}
\trefplural{tab:mimic_performance_and_calibration} and \ref{tab:mimic_test_performance} detail performance and calibration on the MIMIC-III test set. \fref{fig:mimic_test_frequency_groups_calibration} shows performance and calibration across frequency groups on that test set.

\noindent \textbf{Comparison of LLMs and Fine-Tuned Classifiers.} 
\tref{tab:rcv1_subsampled_performance_and_calibration} compares LLMs and fine-tuned classifiers in terms of their performance and calibration.
While zero-shot approaches perform notably worse and are considerably less well-calibrated than fine-tuned models, LLMs with RAG approach fine-tuned models in both performance and calibration.
However, their inference run time is substantially higher: Llama-3.3. with RAG takes almost 10,000 times as long as HiDEC (see \tref{tab:runtimes_rcv1}).

\begin{table*}[t]
    \centering
    \footnotesize
    \setlength{\tabcolsep}{3.6pt}
\begin{tabular}{lll|rr|rrrr}
\toprule
model & loss & OS & $\uparrow$ micro\fscore & $\uparrow$ macro\fscore & $ \downarrow \text{ECE}_{\text{ML}}$ & $ \downarrow \text{ECE}_{\text{ML}}^{\text{high}}$ & $\downarrow \text{ECE}_{\text{ML}}^{\text{med}}$ & $\downarrow \text{ECE}_{\text{ML}}^{\text{rare}}$ \\
    \midrule
BERT & BCE & 0 & 86.2 & 67.8 & 7.8 & 5.0 & 8.4 & 8.3\\[-0.15cm]
& & & \tiny {$\pm 0.1$} & \tiny {$\pm 0.8$} & \tiny {$\pm 0.7$} & \tiny {$\pm 0.4$} & \tiny {$\pm 0.5$} & \tiny {$\pm 1.2$}\\
BERT & WBCEU & 0 & 85.9 & 68.7 & 9.0 & 5.5 & 9.8 & 9.6\\[-0.15cm]
& & & \tiny {$\pm 0.2$} & \tiny {$\pm 0.4$} & \tiny {$\pm 0.6$} & \tiny {$\pm 0.3$} & \tiny {$\pm 0.2$} & \tiny {$\pm 1.7$}\\
BERT & WBCEP & 0 & 84.0 & 67.2 & 14.8 & 6.3 & 13.8 & 20.4\\[-0.15cm]
& & & \tiny {$\pm 0.6$} & \tiny {$\pm 0.9$} & \tiny {$\pm 0.7$} & \tiny {$\pm 0.3$} & \tiny {$\pm 0.7$} & \tiny {$\pm 0.8$}\\
BERT & BCE & 10 & 86.2 & 68.7 & 9.4 & 5.1 & 9.2 & 11.8\\[-0.15cm]
& & & \tiny {$\pm 0.3$} & \tiny {$\pm 0.4$} & \tiny {$\pm 0.9$} & \tiny {$\pm 0.6$} & \tiny {$\pm 1.1$} & \tiny {$\pm 1.2$}\\
BERT & FL ($\gamma=2$) & 0 & \textbf{86.7} & 68.7 & \textbf{5.4} & \textbf{3.4} & \textbf{5.3} & \textbf{6.3}\\[-0.15cm]
& & & \tiny {$\pm 0.0$} & \tiny {$\pm 0.6$} & \tiny {$\pm 0.6$} & \tiny {$\pm 0.7$} & \tiny {$\pm 0.7$} & \tiny {$\pm 0.8$}\\

HiDEC-s & HBM & 0 & 86.4 & \textbf{68.8} & 9.4 & 5.2 & 9.3 & 11.5\\[-0.15cm]
& & & \tiny {$\pm 0.2$} & \tiny {$\pm 0.5$} & \tiny {$\pm 0.4$} & \tiny {$\pm 0.4$} & \tiny {$\pm 0.3$} & \tiny {$\pm 0.8$}\\

$\text{HiDEC-s}^{\dagger}$ & HBM & 0 & 86.6 & 69.2 & 9.3 & 4.8 & 9.0 & 11.7\\[-0.15cm]
& & & \tiny {$\pm 0.1$} & \tiny {$\pm 0.3$} & \tiny {$\pm 0.5$} & \tiny {$\pm 0.4$} & \tiny {$\pm 0.4$} & \tiny {$\pm 1.1$}\\

$\text{HiDEC-s}^{\dagger\dagger}$  & HBM & 0 & 87.0 & 69.7 & 9.4 & 4.9 & 9.2 & 11.8\\[-0.15cm]
& & & \tiny {$\pm 0.2$} & \tiny {$\pm 0.5$} & \tiny {$\pm 0.1$} & \tiny {$\pm 0.2$} & \tiny {$\pm 0.2$} & \tiny {$\pm 0.6$}\\

HiDEC-s & FL ($\gamma=1$) & 0 & 86.5 & 68.5 & 7.6 & 3.9 & 7.8 & 9.0\\[-0.15cm]
& & & \tiny {$\pm 0.1$} & \tiny {$\pm 0.4$} & \tiny {$\pm 0.6$} & \tiny {$\pm 0.6$} & \tiny {$\pm 0.4$} & \tiny {$\pm 1.0$}\\

    \bottomrule
    \end{tabular}
    \caption{\textbf{Performance and calibration on RCV1 test set}. Frequent labels: \frequentlabels{} \rcvtestfrequentlabels{}
    Labels of medium frequency: \mediumlabels{}    \rcvtestmediumlabels{}
    Rare labels: \rarelabels{} \rcvtestrarelabels{}
    \experimentalsetting{}
    macro\fscore: hierarchical macro \fscore score.
    \hidecdaggerinfo{}
    }
    \label{tab:rcv1_performance_and_calibration_full}
\end{table*}

\begin{table*}[t]
    \centering
    \footnotesize
    \setlength{\tabcolsep}{4pt}

\begin{tabular}{lll|rrr|rrr|r}
\toprule
model  & loss & OS & micP & micR & mic\fscore & macP & macR & mac\fscore &  \exactmatch \\
    \midrule
BERT & BCE & 0 & 86.2 & 86.1 & 86.2 & 71.9 & 65.9 & 67.8 & 62.2\\[-0.15cm]
 &  &  & \tiny {$\pm 0.6$} & \tiny {$\pm 0.6$} & \tiny {$\pm 0.1$} & \tiny {$\pm 0.4$} & \tiny {$\pm 1.1$} & \tiny {$\pm 0.8$} & \tiny {$\pm 0.4$}\\
BERT & WBCEU & 0 & 84.0 & 87.8 & 85.9 & 69.9 & 69.2 & 68.7 & 61.2\\[-0.15cm]
 &  &  & \tiny {$\pm 1.1$} & \tiny {$\pm 0.9$} & \tiny {$\pm 0.2$} & \tiny {$\pm 1.4$} & \tiny {$\pm 1.1$} & \tiny {$\pm 0.4$} & \tiny {$\pm 0.8$}\\
BERT & WBCEP & 0 & 78.9 & 89.8 & 84.0 & 61.4 & 77.0 & 67.2 & 57.9\\[-0.15cm]
 &  &  & \tiny {$\pm 1.6$} & \tiny {$\pm 0.8$} & \tiny {$\pm 0.6$} & \tiny {$\pm 1.8$} & \tiny {$\pm 1.4$} & \tiny {$\pm 0.9$} & \tiny {$\pm 1.2$}\\
BERT & BCE & 10 & 86.0 & 86.4 & 86.2 & 72.0 & 67.3 & 68.7 & 62.1\\[-0.15cm]
 &  &  & \tiny {$\pm 0.8$} & \tiny {$\pm 0.3$} & \tiny {$\pm 0.3$} & \tiny {$\pm 0.6$} & \tiny {$\pm 0.6$} & \tiny {$\pm 0.4$} & \tiny {$\pm 0.7$}\\
BERT & FL ($\gamma=2$) & 0 & 87.2 & 86.2 & 86.7 & 73.4 & 66.1 & 68.7 & 63.2\\[-0.15cm]
 &  &  & \tiny {$\pm 0.5$} & \tiny {$\pm 0.5$} & \tiny {$\pm 0.0$} & \tiny {$\pm 0.9$} & \tiny {$\pm 1.4$} & \tiny {$\pm 0.6$} & \tiny {$\pm 0.1$}\\

HiDEC-s & HBM & 0 & 86.4 & 86.4 & 86.4 & 70.8 & 68.2 & 68.8 & 64.2\\[-0.15cm]
 &  &  & \tiny {$\pm 0.5$} & \tiny {$\pm 0.6$} & \tiny {$\pm 0.2$} & \tiny {$\pm 0.9$} & \tiny {$\pm 0.6$} & \tiny {$\pm 0.5$} & \tiny {$\pm 0.1$}\\
$\text{HiDEC-s}^{\dagger}$ & HBM & 0 & 87.5 & 85.8 & 86.6 & 72.9 & 67.4 & 69.2 & 64.7\\[-0.15cm]
 &  &  & \tiny {$\pm 0.6$} & \tiny {$\pm 0.6$} & \tiny {$\pm 0.1$} & \tiny {$\pm 1.1$} & \tiny {$\pm 0.9$} & \tiny {$\pm 0.3$} & \tiny {$\pm 0.4$}\\
$\text{HiDEC-s}^{\dagger\dagger}$ & HBM & 0 & 87.6 & 86.3 & 87.0 & 73.4 & 67.8 & 69.7 & 65.3\\[-0.15cm]
 &  &  & \tiny {$\pm 0.2$} & \tiny {$\pm 0.4$} & \tiny {$\pm 0.2$} & \tiny {$\pm 0.9$} & \tiny {$\pm 0.4$} & \tiny {$\pm 0.5$} & \tiny {$\pm 0.4$}\\
 HiDEC-s & FL ($\gamma=1$) & 0 & 87.4 & 85.6 & 86.5 & 72.5 & 66.2 & 68.5 & 64.3\\[-0.15cm]
 &  &  & \tiny {$\pm 0.8$} & \tiny {$\pm 0.7$} & \tiny {$\pm 0.1$} & \tiny {$\pm 0.8$} & \tiny {$\pm 0.9$} & \tiny {$\pm 0.4$} & \tiny {$\pm 0.3$}\\

    \bottomrule
    \end{tabular}
    \caption{\textbf{Performance on RCV1 test set.} BERT: bert-base-uncased, HiDEC: Hierarchical Decoder \citep{Im_Kim_Oh_Jo_Kim_2023}. \hidecdaggerinfo{} \experimentalsetting{}}
    \label{tab:rcv_test_performance}
\end{table*}

\begin{table*}[t]
    \centering
    \footnotesize
    \setlength{\tabcolsep}{3pt}
\begin{tabular}{lll|rr|rrrr}
\toprule
model & loss & OS & $\uparrow$ micro\fscore & $\uparrow$ macro\fscore & $ \downarrow \text{ECE}_{\text{ML}}$ & $ \downarrow \text{ECE}_{\text{ML}}^{\text{high}}$ & $\downarrow \text{ECE}_{\text{ML}}^{\text{med}}$ & $\downarrow \text{ECE}_{\text{ML}}^{\text{rare}}$ \\
    \midrule
BiomedBERT & BCE & 0 & 40.3 & 10.6 & 3.8 & 14.4 & 16.3 & \textbf{2.2}\\[-0.15cm]
&  &  & \tiny {$\pm 0.4$} & \tiny {$\pm 0.2$} & \tiny {$\pm 0.3$} & \tiny {$\pm 1.2$} & \tiny {$\pm 1.4$} & \tiny {$\pm 0.1$}\\
 
BiomedBERT & WBCEU & 0 & 41.6 & 11.7 & 4.6 & 15.5 & 18.5 & 2.9\\[-0.15cm]
 &  &  & \tiny {$\pm 0.3$} & \tiny {$\pm 0.1$} & \tiny {$\pm 0.4$} & \tiny {$\pm 0.8$} & \tiny {$\pm 0.9$} & \tiny {$\pm 0.3$}\\

BiomedBERT & BCE & 10 & 40.1 & 10.7 & 4.2 & 16.2 & 17.9 & 2.4\\[-0.15cm]
&  &  & \tiny {$\pm 0.1$} & \tiny {$\pm 0.3$} & \tiny {$\pm 0.4$} & \tiny {$\pm 1.6$} & \tiny {$\pm 1.6$} & \tiny {$\pm 0.3$}\\
  
BiomedBERT & FL ($\gamma=1$) & 0 & 40.4 & 10.9 & \textbf{3.6} & 11.7 & 14.9 & \textbf{2.2}\\[-0.15cm]
&  &  & \tiny {$\pm 0.2$} & \tiny {$\pm 0.4$} & \tiny {$\pm 0.2$} & \tiny {$\pm 1.1$} & \tiny {$\pm 1.1$} & \tiny {$\pm 0.1$}\\

PLM-ICD & BCE & 0 & 51.2 & 18.0 & 5.9 & 9.6 & 14.5 & 4.9\\[-0.15cm]
&  &  & \tiny {$\pm 2.9$} & \tiny {$\pm 3.1$} & \tiny {$\pm 1.4$} & \tiny {$\pm 2.2$} & \tiny {$\pm 2.9$} & \tiny {$\pm 1.3$}\\

PLM-ICD & FL ($\gamma=1$) & 0 & \textbf{53.1} & \textbf{20.4} & 5.6 & \textbf{8.3} & \textbf{13.8} & 4.6\\[-0.15cm]
&  &  & \tiny {$\pm 0.8$} & \tiny {$\pm 0.2$} & \tiny {$\pm 0.7$} & \tiny {$\pm 1.2$} & \tiny {$\pm 1.5$} & \tiny {$\pm 0.6$}\\
   \bottomrule
    \end{tabular}
    \vspace{-1em}
    \caption{\textbf{Performance and calibration on MIMIC-III test set}. Highly frequent labels: \frequentlabels{} \mimictestfrequentlabels{} 
    Medium-frequency: \mediumlabels{}    \mimictestmediumlabels{}
    Rare: \rarelabels{} \mimictestrarelabels{}
    \randomrunsdescription{}
    macro\fscore: Macro \fscore on labels with support in the test set.
}
    \label{tab:mimic_performance_and_calibration}
\end{table*}

\begin{table*}
\centering
\footnotesize
 \setlength{\tabcolsep}{4pt}
\begin{tabular}{lll|rrr|rrr|rrr|r}
\toprule
model & loss & OS & micP & micR & mic\fscore & macP & macR & mac\fscore & macPs & macRs & mac\fscore{s} & \exactmatch \\
    \midrule
BiomedBERT & BCE & 0 & 57.4 & 31.0 & 40.3 & 6.8 & 4.3 & 4.8 & 14.8 & 9.4 & 10.6 & 0.1\\[-0.15cm]
 &  &  & \tiny {$\pm 1.3$} & \tiny {$\pm 0.6$} & \tiny {$\pm 0.4$} & \tiny {$\pm 0.1$} & \tiny {$\pm 0.1$} & \tiny {$\pm 0.1$} & \tiny {$\pm 0.3$} & \tiny {$\pm 0.2$} & \tiny {$\pm 0.2$} & \tiny {$\pm 0.0$}\\
    
BiomedBERT & WBCEU & 0 & 49.8 & 35.8 & 41.6 & 6.6 & 5.1 & 5.4 & 14.5 & 11.3 & 11.7 & 0.1\\[-0.15cm]
& &  & \tiny {$\pm 0.8$} & \tiny {$\pm 0.3$} & \tiny {$\pm 0.3$} & \tiny {$\pm 0.0$} & \tiny {$\pm 0.1$} & \tiny {$\pm 0.0$} & \tiny {$\pm 0.1$} & \tiny {$\pm 0.2$} & \tiny {$\pm 0.1$} & \tiny {$\pm 0.0$}\\

BiomedBERT & BCE & 10 & 55.8 & 31.4 & 40.1 & 6.8 & 4.4 & 4.9 & 14.8 & 9.6 & 10.7 & 0.1\\[-0.15cm]
& &  & \tiny {$\pm 1.9$} & \tiny {$\pm 0.6$} & \tiny {$\pm 0.1$} & \tiny {$\pm 0.1$} & \tiny {$\pm 0.1$} & \tiny {$\pm 0.1$} & \tiny {$\pm 0.2$} & \tiny {$\pm 0.3$} & \tiny {$\pm 0.3$} & \tiny {$\pm 0.0$}\\

BiomedBERT & FL ($\gamma=1$) & 0 & 56.9 & 31.3 & 40.4 & 7.0 & 4.4 & 5.0 & 15.2 & 9.7 & 10.9 & 0.1\\[-0.15cm]
 &  &  & \tiny {$\pm 1.4$} & \tiny {$\pm 0.6$} & \tiny {$\pm 0.2$} & \tiny {$\pm 0.2$} & \tiny {$\pm 0.2$} & \tiny {$\pm 0.2$} & \tiny {$\pm 0.4$} & \tiny {$\pm 0.4$} & \tiny {$\pm 0.4$} & \tiny {$\pm 0.0$}\\

PLM-ICD & BCE & 0 & 62.1 & 43.9 & 51.2 & 10.1 & 7.9 & 8.2 & 22.1 & 17.3 & 18.0 & 0.2\\[-0.15cm]
 &  &  & \tiny {$\pm 2.6$} & \tiny {$\pm 4.5$} & \tiny {$\pm 2.9$} & \tiny {$\pm 1.2$} & \tiny {$\pm 1.6$} & \tiny {$\pm 1.4$} & \tiny {$\pm 2.5$} & \tiny {$\pm 3.5$} & \tiny {$\pm 3.1$} & \tiny {$\pm 0.1$}\\

PLM-ICD & FL ($\gamma=1$) & 0 & 61.4 & 46.8 & 53.1 & 11.1 & 9.1 & 9.3 & 24.2 & 20.0 & 20.4 & 0.1\\[-0.15cm]
& &  & \tiny {$\pm 2.3$} & \tiny {$\pm 0.5$} & \tiny {$\pm 0.8$} & \tiny {$\pm 0.3$} & \tiny {$\pm 0.1$} & \tiny {$\pm 0.1$} & \tiny {$\pm 0.6$} & \tiny {$\pm 0.3$} & \tiny {$\pm 0.2$} & \tiny {$\pm 0.0$}\\

    \bottomrule
    \end{tabular}
    \caption{\textbf{Performance on MIMIC-III test set.} BERT: bert-base-uncased, ModernBERT: ModernBERT-base, BiomedBERT: BiomedBERT-base-uncased-abstract, each using the CLS token for classification. PLM-ICD: PLM-ICD \citep{huang-etal-2022-plm} with BiomedBERT-base-uncased-abstract. \experimentalsetting{} \macsexplanation{}}
    \label{tab:mimic_test_performance}
\end{table*}

\begin{table*}[htb]
    \centering
    \footnotesize
    \setlength{\tabcolsep}{2pt}
\begin{tabular}{lll|rr|rrrr}
\toprule
model & loss & OS & $\uparrow$ micro\fscore & $\uparrow$ macro\fscore & $ \downarrow \text{ECE}_{\text{ML}}$ & $ \downarrow \text{ECE}_{\text{ML}}^{\text{high}}$ & $\downarrow \text{ECE}_{\text{ML}}^{\text{med}}$ & $\downarrow \text{ECE}_{\text{ML}}^{\text{rare}}$ \\
    \midrule
BERT & BCE & 0 & 85.1 & 66.4 & 7.9 & 5.0 & 8.9 & 7.8\\[-0.15cm]
& & &\tiny {$\pm 0.2$} & \tiny {$\pm 1.0$} & \tiny {$\pm 0.2$} & \tiny {$\pm 0.3$} & \tiny {$\pm 0.3$} & \tiny {$\pm 0.8$}\\

BERT & WBCEU & 0 & 85.2 & \textbf{67.5} & 8.9 & 5.4 & 10.1 & 8.8\\[-0.15cm]
& & &\tiny {$\pm 0.1$} & \tiny {$\pm 0.6$} & \tiny {$\pm 0.4$} & \tiny {$\pm 0.3$} & \tiny {$\pm 0.2$} & \tiny {$\pm 0.8$}\\

BERT & WBCEP & 0 & 83.7 & 65.7 & 13.9 & 6.1 & 13.4 & 18.1\\[-0.15cm]
& & &\tiny {$\pm 0.4$} & \tiny {$\pm 0.5$} & \tiny {$\pm 0.8$} & \tiny {$\pm 0.3$} & \tiny {$\pm 0.5$} & \tiny {$\pm 1.6$}\\

BERT & BCE & 10 & 85.1 & 67.1 & 8.4 & 5.1 & 9.5 & 8.4\\[-0.15cm]
& & &\tiny {$\pm 0.3$} & \tiny {$\pm 0.9$} & \tiny {$\pm 0.6$} & \tiny {$\pm 0.6$} & \tiny {$\pm 1.0$} & \tiny {$\pm 0.9$}\\

BERT & FL ($\gamma=2$) & 0 & \textbf{85.7} & 67.1 & \textbf{6.7} & \textbf{3.4} & \textbf{6.8} & 8.1\\[-0.15cm]
& & &\tiny {$\pm 0.2$} & \tiny {$\pm 0.7$} & \tiny {$\pm 0.4$} & \tiny {$\pm 0.9$} & \tiny {$\pm 0.3$} & \tiny {$\pm 0.7$}\\

HiDEC-s & HBM & 0 & 85.3 & 67.2 & 9.0 & 4.8 & 10.1 & 9.5\\[-0.15cm]
& &  & \tiny {$\pm 0.3$} & \tiny {$\pm 0.7$} & \tiny {$\pm 0.5$} & \tiny {$\pm 0.4$} & \tiny {$\pm 0.3$} & \tiny {$\pm 1.3$}\\

$\text{HiDEC-s}^{\dagger}$ & HBM & 0 & 85.3 & 67.3 & 8.7 & 4.7 & 9.7 & 9.2\\[-0.15cm]
& &  & \tiny {$\pm 0.2$} & \tiny {$\pm 0.8$} & \tiny {$\pm 0.3$} & \tiny {$\pm 0.4$} & \tiny {$\pm 0.3$} & \tiny {$\pm 0.9$}\\

$\text{HiDEC-s}^{\dagger\dagger}$ & HBM & 0 & 85.7 & 67.2 & 8.7 & 4.7 & 9.8 & 9.1\\[-0.15cm]
& &  & \tiny {$\pm 0.1$} & \tiny {$\pm 0.4$} & \tiny {$\pm 0.4$} & \tiny {$\pm 0.2$} & \tiny {$\pm 0.3$} & \tiny {$\pm 1.1$}\\

HiDEC-s & FL ($\gamma=1$) & 0 & 85.2 & 66.5 & 7.5 & 4.0 & 8.5 & \textbf{7.7}\\[-0.15cm]
& &  & \tiny {$\pm 0.1$} & \tiny {$\pm 0.6$} & \tiny {$\pm 0.4$} & \tiny {$\pm 0.6$} & \tiny {$\pm 0.4$} & \tiny {$\pm 0.8$}\\

\midrule

model & prompting & confidence & $\uparrow$ micro\fscore & $\uparrow$ macro\fscore & $ \downarrow \text{ECE}_{\text{ML}}$ & $ \downarrow \text{ECE}_{\text{ML}}^{\text{high}}$ & $\downarrow \text{ECE}_{\text{ML}}^{\text{med}}$ & $\downarrow \text{ECE}_{\text{ML}}^{\text{rare}}$ \\
\midrule

Llama-3.3 & zero-shot & Verbalized & 64.8 & 46.1 & 19.7 & 13.7 & 18.5 & 24.2\\
Llama-3.3 & zero-shot & Token Probabilities & 63.0 & 42.8 & 22.0 & 14.4 & 19.2 & 29.3\\

Llama-3.3 & RAG & Verbalized & 82.9 & 62.5 & 8.5 & 4.5 & 8.1 & 10.9\\
Llama-3.3 & RAG & Token Probabilities & 82.9 & 62.5 & 8.0 & 4.7 & 7.9 & 9.8\\

\hline

Qwen2.5 & zero-shot & Verbalized &  57.1 & 38.5 & 28.3 & 18.3 & 27.2 & 34.4\\
Qwen2.5 & zero-shot & Token Probabilities & 59.1 & 39.0 & 24.7 & 16.6 & 22.1 & 32.0\\

Qwen2.5 & RAG & Verbalized & 83.1 & 62.8 & 9.8 & 4.2 & 9.8 & 12.5\\
Qwen2.5 & RAG & Token Probabilities & 82.6 & 62.3 & 8.2 & 4.7 & 7.4 & 10.8\\

    \bottomrule
    \end{tabular}
    \caption{\textbf{Performance and calibration on subsampled RCV1 test set}. \rcvsubsampled{} Frequent labels: \frequentlabels{} \rcvtestsubsampledfrequentlabels{}
    Labels of medium frequency: \mediumlabels{}    \rcvtestsubsampledmediumlabels{}
    Rare labels: \rarelabels{} \rcvtestsubsampledrarelabels{}
    \experimentalsetting{} macro\fscore: hierarchical macro \fscore score.
    \hidecdaggerinfo{}
    }
    \label{tab:rcv1_subsampled_performance_and_calibration}
\end{table*}

\end{document}